\documentclass[journal]{IEEEtran}

\usepackage[cmex10]{amsmath}
\usepackage{amsmath}
\usepackage{bm}
\numberwithin{equation}{section}

\usepackage{amssymb}

\usepackage{tabularx}
\usepackage{booktabs}
\usepackage{multirow}
\usepackage{slashbox}

\usepackage{multirow}
\usepackage{tabularx}
\usepackage{array}
\usepackage{graphicx}
\usepackage{subfig}
\usepackage{amssymb}

\usepackage{float}
\usepackage{indentfirst}
\usepackage[numbers,sort&compress]{natbib}

\ifCLASSINFOpdf
\else
\fi

\begin{document}
%
\title{A survey of sparse representation: algorithms and applications}
%
%
%

\author{Zheng~Zhang,~\IEEEmembership{Student Member,~IEEE,}~Yong~Xu,~\IEEEmembership{Senior Member,~IEEE,}
        Jian~Yang,~\IEEEmembership{Member,~IEEE,}~Xuelong~Li,~\IEEEmembership{Fellow,~IEEE,}
        and~David~Zhang,~\IEEEmembership{~Fellow,~IEEE}%
\thanks{Zheng~Zhang and Yong Xu is with the Bio-Computing Research Center, Shenzhen Graduate School, Harbin Institute of Technology, Shenzhen 518055, Guangdong, P.R. China; Key Laboratory of Network Oriented Intelligent Computation, Shenzhen 518055, Guangdong, P.R. China e-mail: (yongxu@ymail.com).}
\thanks{Jian Yang is with the College of Computer Science and Technology, Nanjing University of Science and Technology, Nanjing 210094, P. R. China.}
\thanks{Xuelong Li is with the Center for OPTical IMagery Analysis and Learning (OPTIMAL), State Key Laboratory of Transient Optics and Photonics, Xi'an Institute of Optics and Precision Mechanics, Chinese Academy of Sciences, Xi'an 710119, Shaanxi, P. R. China.}
\thanks{David Zhang is with the Biometrics Research Center, The Hong Kong Polytechnic University, Hong Kong}
\thanks{Corresponding author: Yong Xu (email: yongxu@ymail.com).}}

\markboth{Journal}%
{A survey of sparse representation: algorithms and applications}

\maketitle

\begin{abstract}
Sparse representation has attracted much attention from researchers in fields of signal processing, image processing, computer vision and pattern recognition. Sparse representation also has a good reputation in both theoretical research and practical applications. Many different algorithms have been proposed for sparse representation. The main purpose of this article is to provide a comprehensive study and an updated review on sparse representation and to supply a guidance for researchers. The taxonomy of sparse representation methods can be studied from various viewpoints. For example, in terms of different norm minimizations used in sparsity constraints, the methods can be roughly categorized into five groups: sparse representation with  $l_0$-norm minimization, sparse representation with  $l_p$-norm (0$<$p$<$1) minimization, sparse representation with  $l_1$-norm minimization and sparse representation with  $l_{2,1}$-norm minimization. In this paper, a comprehensive overview of sparse representation is provided. The available sparse representation algorithms can also be empirically categorized into four groups: greedy strategy approximation, constrained optimization, proximity algorithm-based optimization, and homotopy algorithm-based sparse representation. The rationales of different algorithms in each category are analyzed and a wide range of sparse representation applications are summarized, which could sufficiently reveal the potential nature of the sparse representation theory. Specifically, an experimentally comparative study of these sparse representation algorithms was presented. The Matlab code used in this paper can be available at: http://www.yongxu.org/lunwen.html.
\end{abstract}


\IEEEpeerreviewmaketitle

\section{Introduction}

\IEEEPARstart{W}{ith} advancements in mathematics, linear representation methods (LRBM) have been well studied and have recently received considerable attention \cite{natarajan95, huang2014brain}. The sparse representation method is the most representative methodology of the LRBM and has also been proven to be an extraordinary powerful solution to a wide range of application fields, especially in signal processing, image processing, machine learning, and computer vision, such as image denoising, debluring, inpainting, image restoration, super-resolution, visual tracking, image classification and image segmentation \cite{lu2014group, Elad2010on, mallat2008wavelet, starck2010sparse, elad2010sparse, bruckstein2009sparse, xu2011two, wright2010sparse}. Sparse representation has shown huge potential capabilities in handling these problems.

Sparse representation, from the viewpoint of its origin, is directly related to compressed sensing (CS) \cite{donoho2006compressed, baraniuk2007compressive, candes2006robust}, which is one of the most popular topics in recent years. Donoho \cite{donoho2006compressed} first proposed the original concept of compressed sensing. CS theory suggests that if a signal is sparse or compressive, the original signal can be reconstructed by exploiting a few measured values, which are much less than the ones suggested by previously used theories such as Shannon's sampling theorem (SST). Candes et al. \cite{candes2006robust}, from the mathematical perspective, demonstrated the rationale of CS theory, i.e. the original signal could be precisely reconstructed by utilizing a small portion of Fourier transformation coefficients. Baraniuk \cite{baraniuk2007compressive} provided a concrete analysis of compressed sensing and presented a specific interpretation on some solutions of different signal reconstruction algorithms. All these literature \cite{donoho2006compressed, baraniuk2007compressive, candes2006robust, candes2008introduction, tsaig2006extensions, candes2006compressive, candes2007sparsity} laid the foundation of CS theory and provided the theoretical basis for future research. Thus, a large number of algorithms based on CS theory have been proposed to address different problems in various fields. Moreover, CS theory always includes the three basic components: sparse representation, encoding measuring, and reconstructing algorithm. As an indispensable prerequisite of CS theory, the sparse representation theory \cite{Elad2010on, elad2010sparse, bruckstein2009sparse, xu2011two, wright2010sparse, candes2007sparsity} is the most outstanding technique used to conquer difficulties that appear in many fields. For example, the methodology of sparse representation is a novel signal sampling method for the sparse or compressible signal and has been successfully applied to signal processing \cite{Elad2010on, mallat2008wavelet, starck2010sparse}.

Sparse representation has attracted much attention in recent years and many examples in different fields can be found where sparse representation is definitely beneficial and favorable \cite{lu2013manifold, yuan2009binary}. One example is image classification, where the basic goal is to classify the given test image into several predefined categories. It has been demonstrated that natural images can be sparsely represented from the perspective of the properties of visual neurons. The sparse representation based classification (SRC) method \cite{wright2009robust} first assumes that the test sample can be sufficiently represented by samples from the same subject. Specifically, SRC exploits the linear combination of training samples to represent the test sample and computes sparse representation coefficients of the linear representation system, and then calculates the reconstruction residuals of each class employing the sparse representation coefficients and training samples. The test sample will be classified as a member of the class, which leads to the minimum reconstruction residual. The literature \cite{wright2009robust} has also demonstrated that the SRC method has great superiorities when addressing the image classification issue on corrupted or disguised images. In such cases, each natural image can be sparsely represented and the sparse representation theory can be utilized to fulfill the image classification task.

For signal processing, one important task is to extract key components from a large number of clutter signals or groups of complex signals in coordination with different requirements. Before the appearance of sparse representation, SST and Nyquist sampling law (NSL) were the traditional methods for signal acquisition and the general procedures included sampling, coding compression, transmission, and decoding. Under the frameworks of SST and NSL, the greatest difficulty of signal processing lies in efficient sampling from mass data with sufficient memory-saving. In such a case, sparse representation theory can simultaneously break the bottleneck of conventional sampling rules, i.e. SST and NSL, so that it has a very wide application prospect. Sparse representation theory proposes to integrate the processes of signal sampling and coding compression. Especially, sparse representation theory employs a more efficient sampling rate to measure the original sample by abandoning the pristine measurements of SST and NSL, and then adopts an optimal reconstruction algorithm to reconstruct samples. In the context of compressed sensing, it is first assumed that all the signals are sparse or approximately sparse enough \cite{Elad2010on, starck2010sparse, elad2010sparse}. Compared to the primary signal space, the size of the set of possible signals can be largely decreased under the constraint of sparsity. Thus, massive algorithms based on the sparse representation theory have been proposed to effectively tackle signal processing issues such as signal reconstruction and recovery. To this end, the sparse representation technique can save a significant amount of sampling time and sample storage space and it is favorable and advantageous.

\subsection{Categorization of sparse representation techniques}
\noindent Sparse representation theory can be categorized from different viewpoints. Because different methods have their individual motivations, ideas, and concerns, there are varieties of strategies to separate the existing sparse representation methods into different categories from the perspective of taxonomy. For example, from the viewpoint of ``atoms", available sparse representation methods can be categorized into two general groups: naive sample based sparse representation and dictionary learning based sparse representation. However, on the basis of the availability of labels of ``atoms", sparse representation and learning methods can be coarsely divided into three groups: supervised learning, semi-supervised learning, and unsupervised learning methods. Because of the sparse constraint, sparse representation methods can be divided into two communities: structure constraint based sparse representation and sparse constraint based sparse representation. Moreover, in the field of image classification, the representation based classification methods consist of two main categories in terms of the way of exploiting the ``atoms": the holistic representation based method and local representation based method \cite{zhang2014integrating}. More specifically, holistic representation based methods exploit training samples of all classes to represent the test sample, whereas local representation based methods only employ training samples (or atoms) of each class or several classes to represent the test sample. Most of the sparse representation methods are holistic representation based methods. A typical and representative local sparse representation methods is the two-phase test sample sparse representation (TPTSR) method \cite{xu2011two}. In consideration of different methodologies, the sparse representation method can be grouped into two aspects: pure sparse representation and hybrid sparse representation, which improves the pre-existing sparse representation methods with the aid of other methods. The literature \cite{cheng2013sparse} suggests that sparse representation algorithms roughly fall into three classes: convex relaxation, greedy algorithms, and combinational methods. In the literature \cite{tropp2006a, tropp2006b}, from the perspective of sparse problem modeling and problem solving, sparse decomposition algorithms are generally divided into two sections: greedy algorithms and convex relaxation algorithms. On the other hand, if the viewpoint of optimization is taken into consideration, the problems of sparse representation can be divided into four optimization problems: the smooth convex problem, nonsmooth nonconvex problem, smooth nonconvex problem, and nonsmooth convex problem. Furthermore, Schmidt et al. \cite{schmidt2009optimization} reviewed some optimization techniques for solving $l_1$-norm regularization problems and roughly divided these approaches into three optimization strategies: sub-gradient methods, unconstrained approximation methods, and constrained optimization methods. The supplementary file attached with the paper also offers more useful information to make fully understandings of the `taxonomy' of current sparse representation techniques in this paper.

\begin{figure*}[htbp]
\centering
\includegraphics[width=7in]{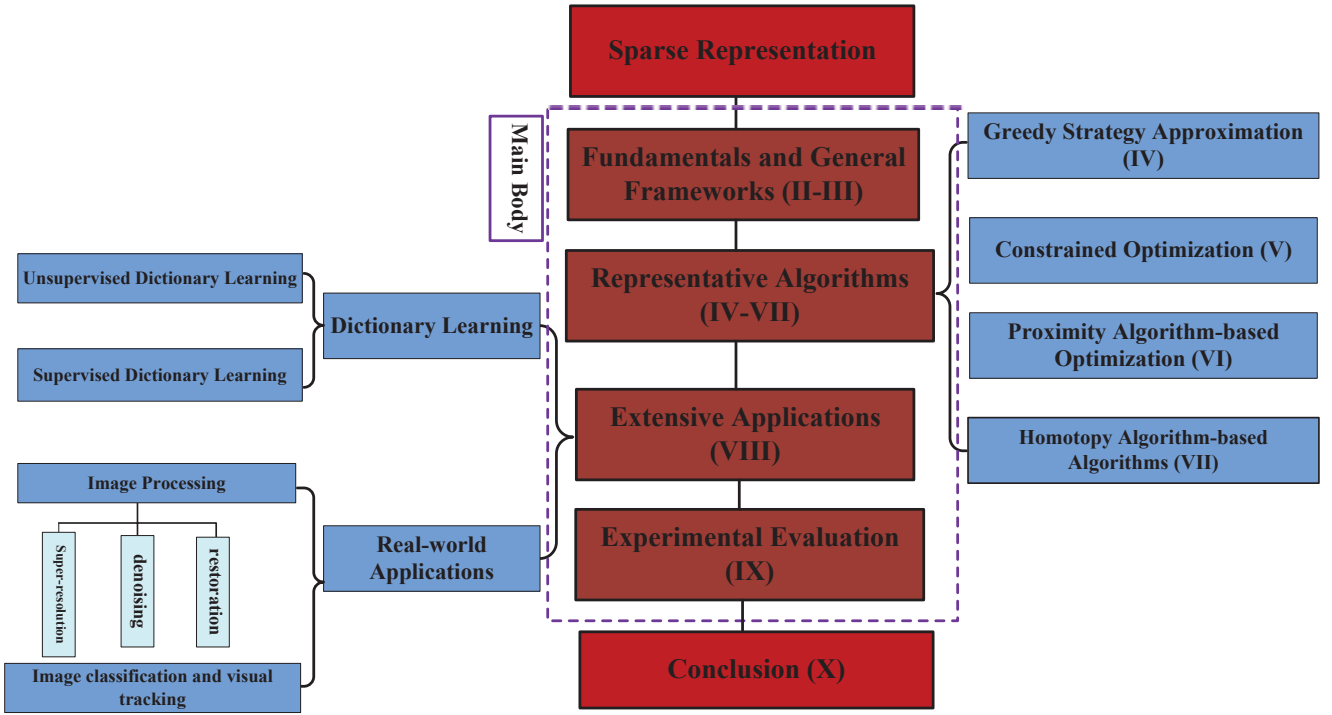}
\caption{The structure of this paper. The main body of this paper mainly consists of four parts: basic concepts and frameworks in Section II-III, representative algorithms in Section IV-VII and extensive applications in Section VIII, massive experimental evaluations in Section IX. Conclusion is summarized in Section X.}
\label{figStru}
\end{figure*}

In this paper, the available sparse representation methods are categorized into four groups, i.e. the greedy strategy approximation, constrained optimization strategy, proximity algorithm based optimization strategy, and homotopy algorithm based sparse representation, with respect to the analytical solution and optimization viewpoints.

(1) In the greedy strategy approximation for solving sparse representation problem, the target task is mainly to solve the sparse representation method with $l_0$-norm minimization. Because of the fact that this problem is an NP-hard problem \cite{amaldi1998approximability}, the greedy strategy provides an approximate solution to alleviate this difficulty. The greedy strategy searches for the best local optimal solution in each iteration with the goal of achieving the optimal holistic solution \cite{tropp2004greed}. For the sparse representation method, the greedy strategy approximation only chooses the most $k$ appropriate samples, which are called  $k$-sparsity, to approximate the measurement vector.

(2) In the constrained optimization strategy, the core idea is to explore a suitable way to transform a non-differentiable optimization problem into a differentiable optimization problem by replacing the $l_1$-norm minimization term, which is convex but nonsmooth, with a differentiable optimization term, which is convex and smooth. More specifically, the constrained optimization strategy substitutes the $l_1$-norm minimization term with an equal constraint condition on the original unconstraint problem. If the original unconstraint problem is reformulated into a differentiable problem with constraint conditions, it will become an uncomplicated problem in the consideration of the fact that $l_1$-norm minimization is global non-differentiable.

(3) Proximal algorithms can be treated as a powerful tool for solving nonsmooth, constrained, large-scale, or distributed versions of the optimization problem \cite{Parikh2013Proximal}. In the proximity algorithm based optimization strategy for sparse representation, the main task is to reformulate the original problem into the specific model of the corresponding proximal operator such as the soft thresholding operator, hard thresholding operator, and resolvent operator, and then exploits the proximity algorithms to address the original sparse optimization problem.

(4) The general framework of the homotopy algorithm is to iteratively trace the final desired solution starting from the initial point to the optimal point by successively adjusting the homotopy parameter \cite{Donoho2008Fast}. In homotopy algorithm based sparse representation, the homotopy algorithm is used to solve the $l_1$-norm minimization problem with  $k$-sparse property.

\subsection{ Motivation and objectives}
\noindent In this paper, a survey on sparse representation and overview available sparse representation algorithms from viewpoints of the mathematical and theoretical optimization is provided. This paper is designed to provide foundations of the study on sparse representation and aims to give a good start to newcomers in computer vision and pattern recognition communities, who are interested in sparse representation methodology and its related fields. Extensive state-of-art sparse representation methods are summarized and the ideas, algorithms, and wide applications of sparse representation are comprehensively presented. Specifically, there is concentration on introducing an up-to-date review of the existing literature and presenting some insights into the studies of the latest sparse representation methods. Moreover, the existing sparse representation methods are divided into different categories. Subsequently, corresponding typical algorithms in different categories are presented and their distinctness is explicitly shown. Finally, the wide applications of these sparse representation methods in different fields are introduced.

\begin{figure*}[htbp]
\centering
\subfloat[]
{
    \label{fig1-a}
    \includegraphics[width=1.15in]{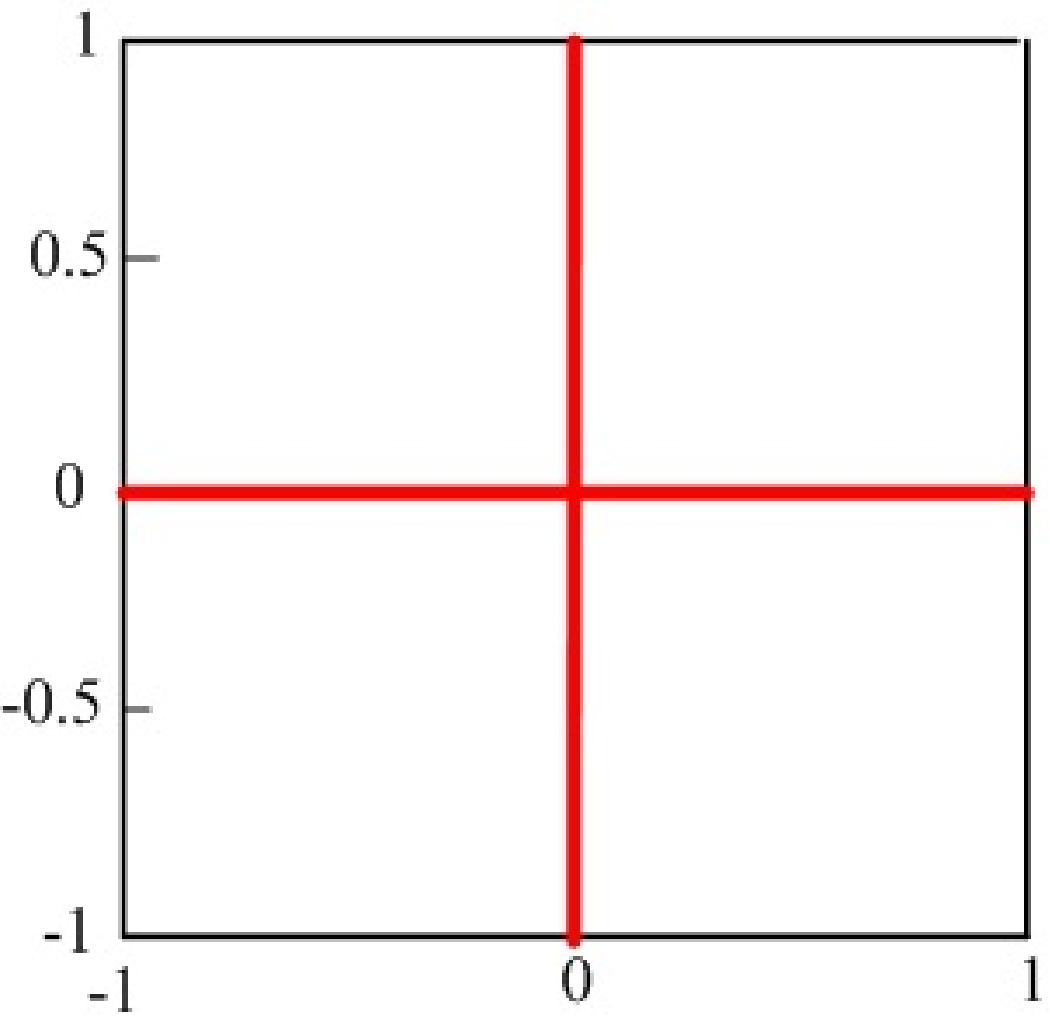}
}
\hspace{4pt}
\subfloat[]
{
    \label{fig1-b}
    \includegraphics[width=1.15in]{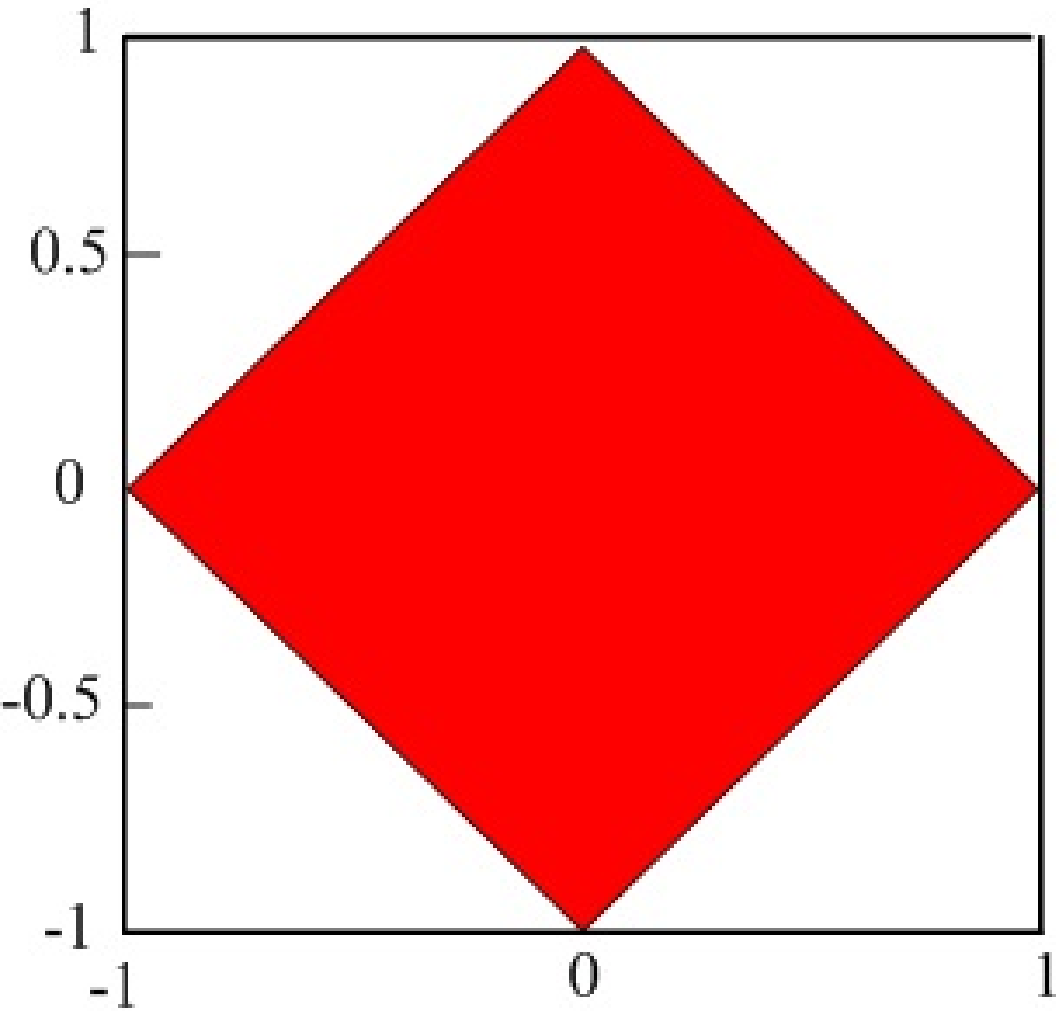}
}
\hspace{4pt}
\subfloat[]
{
    \label{fig1-c}
    \includegraphics[width=1.15in]{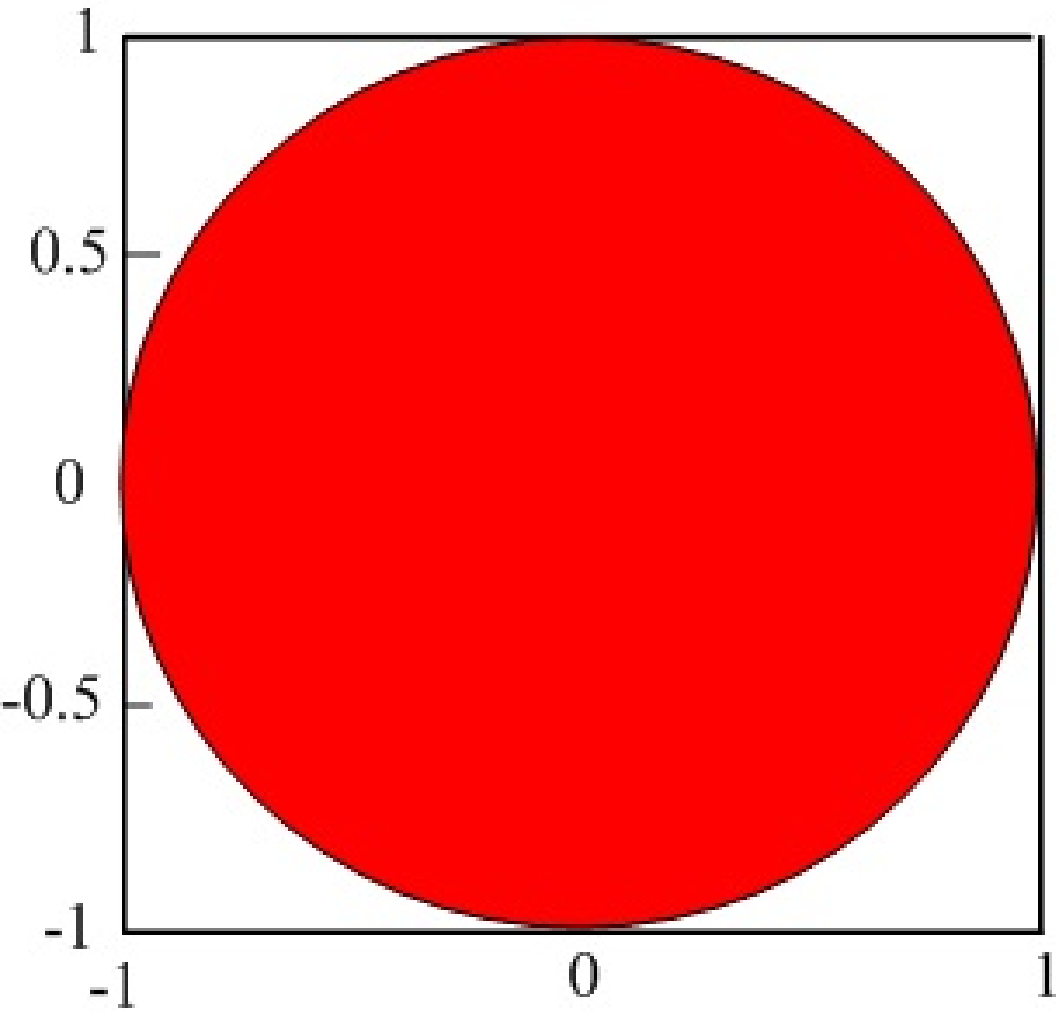}
}
\hspace{4pt}
\subfloat[]
{
    \label{fig1-d}
    \includegraphics[width=1.15in]{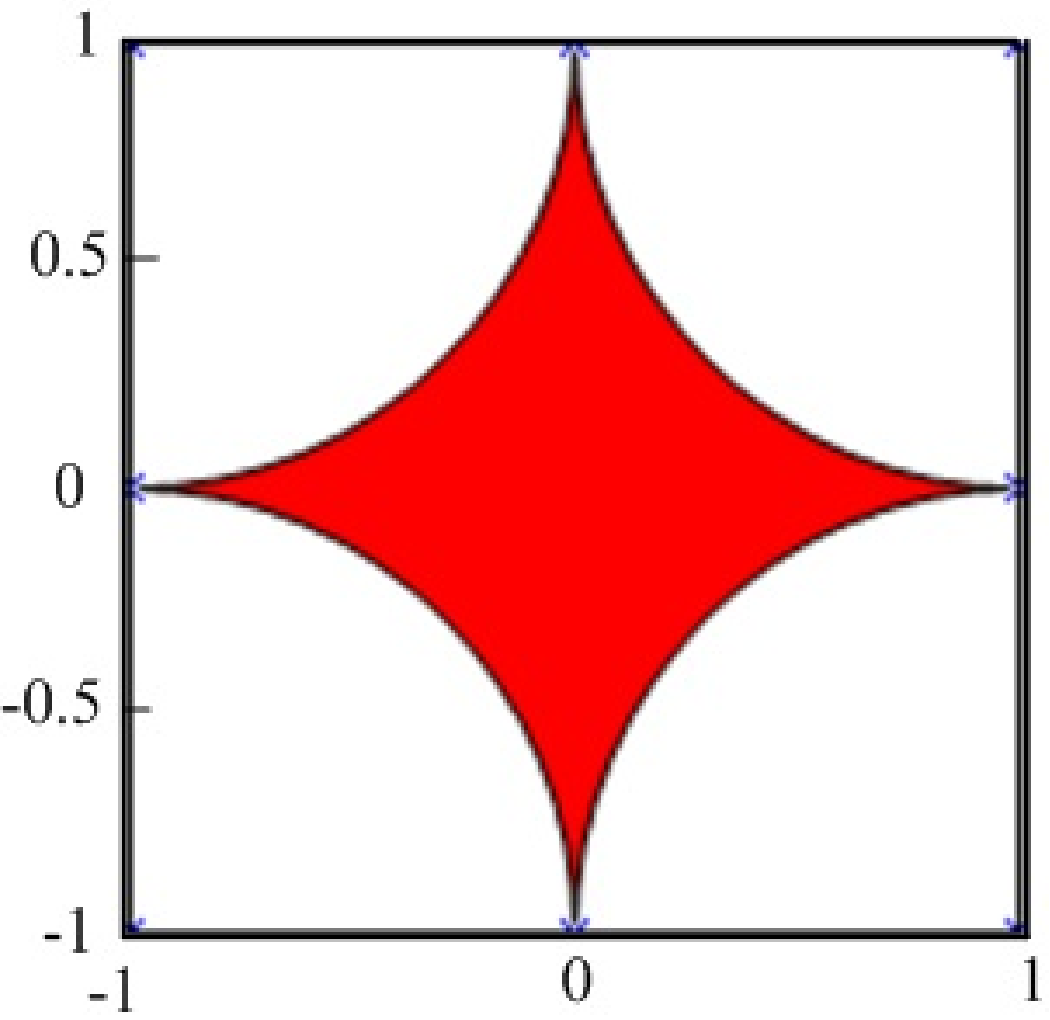}
}
\caption{Geometric interpretations of different norms in 2-D space \cite{elad2010sparse}. (a), (b), (c), (d) are the unit ball of the $l_0$-norm, $l_1$-norm, $l_2$-norm, $l_p$-norm (0$<$p$<$1) in 2-D space, respectively. The two axes of the above coordinate systems are $x_1$ and $x_2$.}
\end{figure*}

The remainder of this paper is mainly composed of four parts: basic concepts and frameworks are shown in Section II and Section III, representative algorithms are presented in Section IV-VII and extensive applications are illustrated in Section VIII, massive experimental evaluations are summarized in Section IX. More specifically, the fundamentals and preliminary mathematic concepts are presented in Section II, and then the general frameworks of the existing sparse representation with different norm regularizations are summarized in Section III. In Section IV, the greedy strategy approximation method is presented for obtaining a sparse representation solution, and in Section V, the constrained optimization strategy is introduced for solving the sparse representation issue. Furthermore, the proximity algorithm based optimization strategy and Homotopy strategy for addressing the sparse representation problem are outlined in Section VI and Section VII, respectively. Section VIII presents extensive applications of sparse representation in widespread and prevalent fields including dictionary learning methods and real-world applications. Finally, Section IX offers massive experimental evaluations and conclusions are drawn and summarized in Section X. The structure of the this paper has been summarized in Fig. \ref{figStru}.

\section{Fundamentals and preliminary concepts}
\subsection{Notations}
\noindent In this paper, vectors are denoted by lowercase letters with bold face, e.g. $\bm{x}$. Matrices are denoted by uppercase letter, e.g. $X$ and their elements are denoted with indexes such as $X_i$. In this paper, all the data are only real-valued.

Suppose that the sample is from space $\mathbb{R}^d$ and thus all the samples are concatenated to form a matrix, denoted as $D \in \mathbb{R}^{d \times n}$.  If any sample can be approximately represented by a linear combination of dictionary $D$ and the number of the samples is larger than the dimension of samples in $D$, i.e. $n>d$, dictionary $D$ is referred to as an over-complete dictionary. A signal is said to be compressible if it is a sparse signal in the original or transformed domain when there is no information or energy loss during the process of transformation.

``\textbf{sparse}'' or ``\textbf{sparsity}'' of a vector means that some elements of the vector are zero. We use a linear combination of a basis matrix $A \in R^{N\times N}$ to represent a signal $\bm{x}\in R^{N \times 1}$, i.e. $\bm{x}=A\bm{s}$ where $\bm{s}\in R^{N\times 1}$ is the column vector of weighting coefficients. If only $k$ ($k\ll N$) elements of $\bm{s}$ are nonzero and the rest elements in $\bm{s}$ are zero, we call the signal $\bm{x}$ is $k$-sparse.

\subsection{Basic background}
\noindent The standard inner product of two vectors, $\bm{x}$ and $\bm{y}$ from the set of real $n$ dimensions, is defined as
\begin{equation} \label{2-1}
\langle \bm{x},\bm{y} \rangle=\bm{x}^T\bm{y}=\bm{x}_1\bm{y}_1+\bm{x}_2\bm{y}_2+\cdots+\bm{x}_n\bm{y}_n
\end{equation}

The standard inner product of two matrixes, $X \in \mathbb{R}^{m \times n}$ and $Y \in \mathbb{R}^{m \times n}$ from the set of real $m\times n$ matrixes, is denoted as the following equation
\begin{equation} \label{2-2}
\langle X,Y \rangle=tr(X^TY)= \sum_{i=1}^m \sum_{j=1}^n X_{ij}Y_{ij}
\end{equation}
where the operator $tr(A)$ denotes the trace of the matrix $A$, i.e. the sum of its diagonal entries.

Suppose that $\bm{v}=[\bm{v}_1,\bm{v}_2,\cdots,\bm{v}_n]$ is an $n$ dimensional vector in Euclidean space, thus
\begin{equation}\label{2-3}
\|\bm{v}\|_p=(\sum_{i=1}^n |\bm{v}_i|^p)^{1/p}
\end{equation}
is denoted as the $p$-norm or the $l_p$-norm ($1\leq p \leq \infty$) of vector $v$.

When p$=$1, it is called the  $l_1$-norm. It means the sum of absolute values of the elements in vector $v$ , and its geometric interpretation is shown in Fig. \ref{fig1-b}, which is a square with a forty-five degree rotation.

When p$=$2, it is called the $l_2$-norm or Euclidean norm. It is defined as $\|\bm{v}\|_2=(\bm{v}_1^2+\bm{v}_2^2+\cdots+\bm{v}_n^2)^{1/2}$, and its geometric interpretation in 2-D space is shown in Fig. \ref{fig1-c} which is a circle.

In the literature, the sparsity of a vector $v$ is always related to the so-called $l_0$-norm, which means the number of the nonzero elements of vector $\bm{v}$. Actually, the $l_0$-norm is the limit as $p\rightarrow 0$ of the $l_p$-norms \cite{bruckstein2009sparse} and the definition of the $l_0$-norm is formulated as
\begin{equation}\label{l0norm}
\|\bm{v}\|_0= \lim_{p\rightarrow 0} \|\bm{v}\|_p^p = \lim_{p\rightarrow 0} \sum_{i=1}^n |v_i|^p
\end{equation}
We can see that the notion of the $l_0$-norm is very convenient and intuitive for defining the sparse representation problem. The property of the $l_0$-norm can also be presented from the perspective of geometric interpretation in 2-D space, which is shown in Fig. \ref{fig1-a}, and it is a crisscross.

Furthermore, the geometric meaning of the $l_p$-norm (0$<$p$<$1) is also presented, which is a form of similar recessed pentacle shown in Fig. \ref{fig1-d}.

\begin{figure}[htbp]
\centering
\includegraphics[width=3.5in]{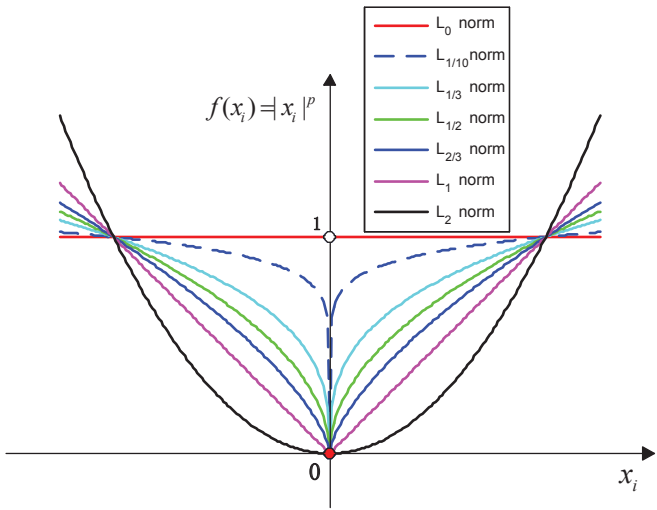}
\caption{Geometric interpretations of different norms in 1-D space \cite{elad2010sparse}.}
\label{fig2}
\end{figure}

\begin{figure*}[htbp]
\centering
\subfloat[]
{
    \label{fig3-a}
    \includegraphics[width=2in]{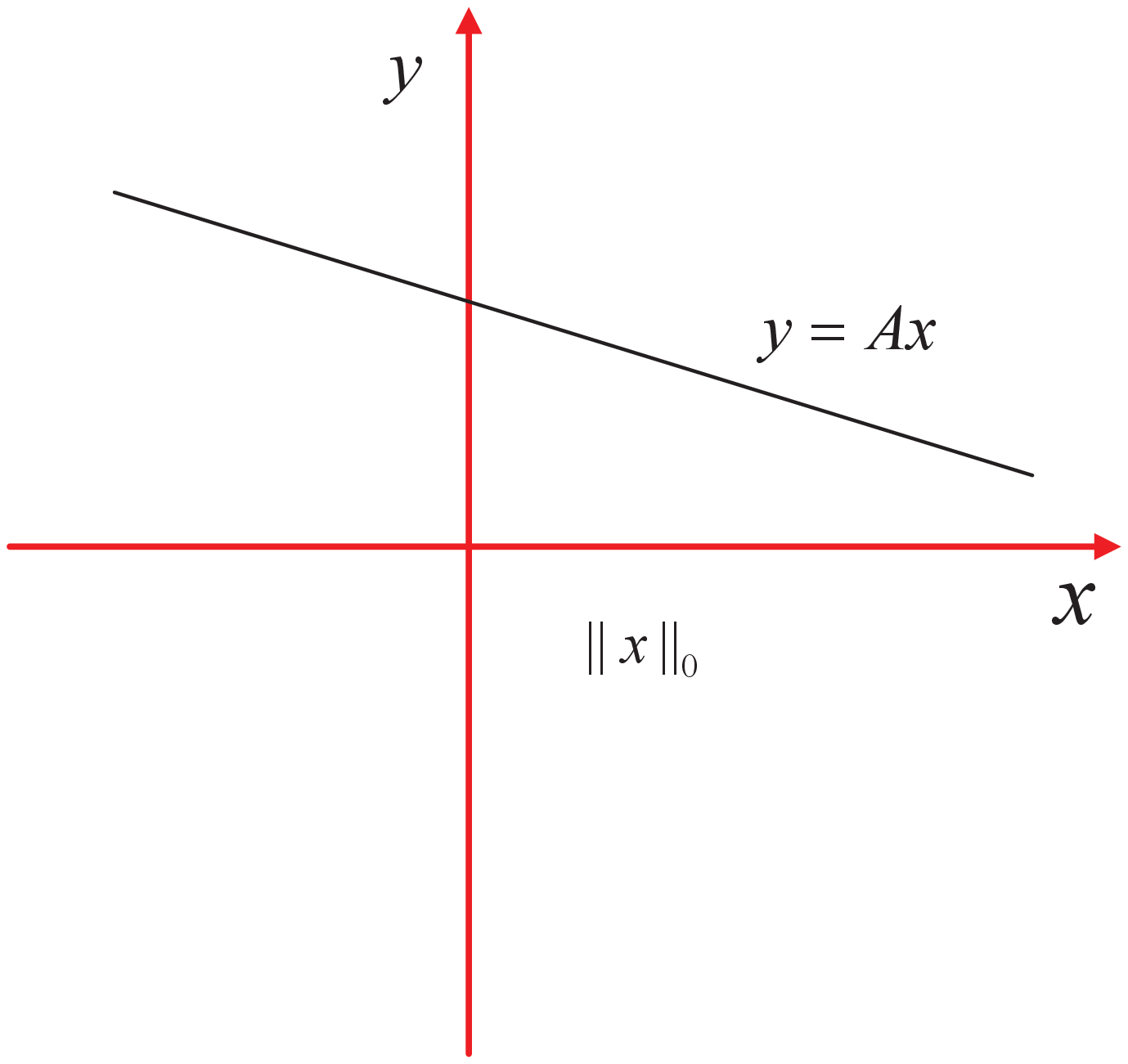}
}
\hspace{2pt}
\subfloat[]
{
    \label{fig3-b}
    \includegraphics[width=2in]{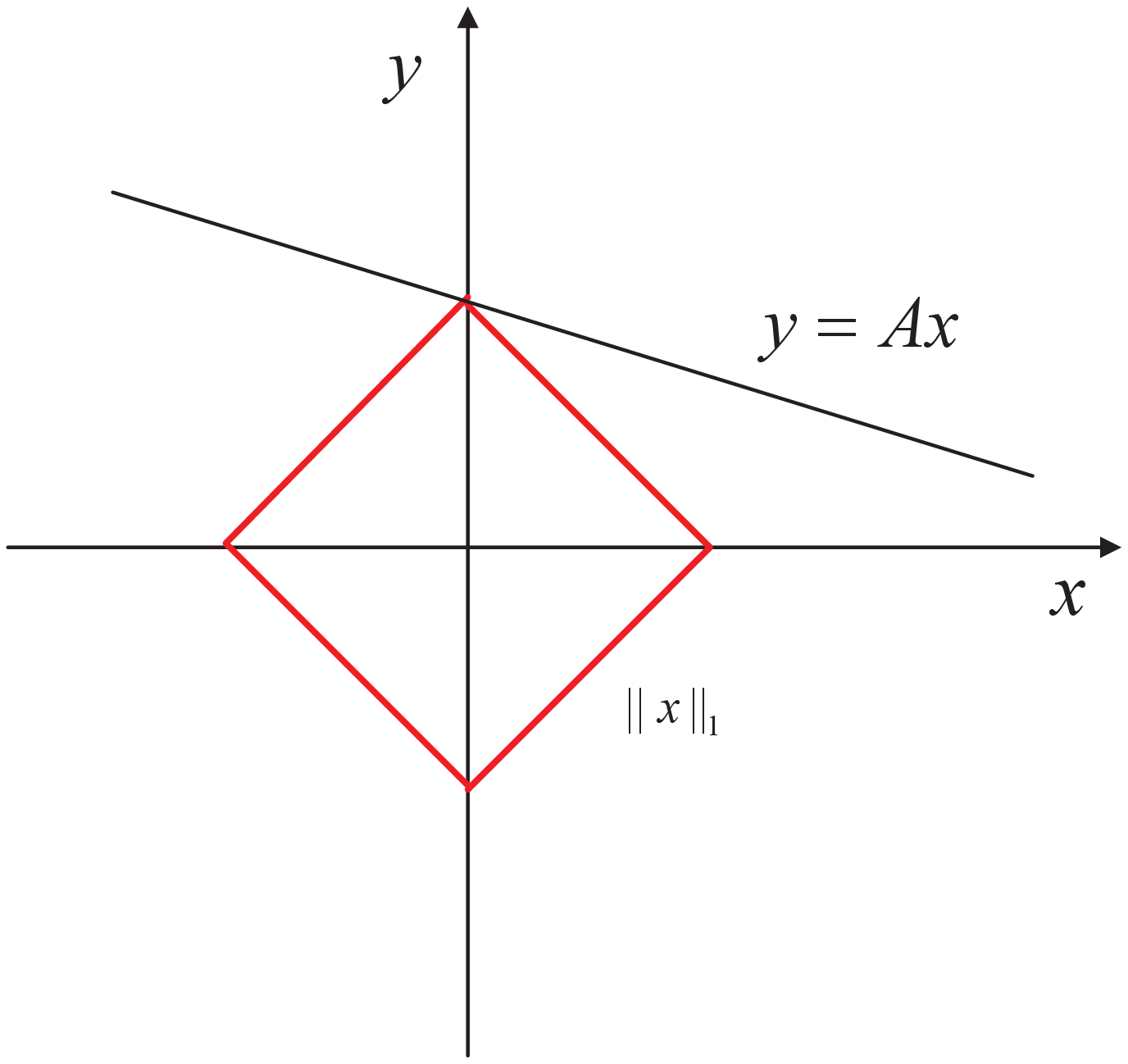}
}
\hspace{2pt}
\subfloat[]
{
    \label{fig3-c}
    \includegraphics[width=2in]{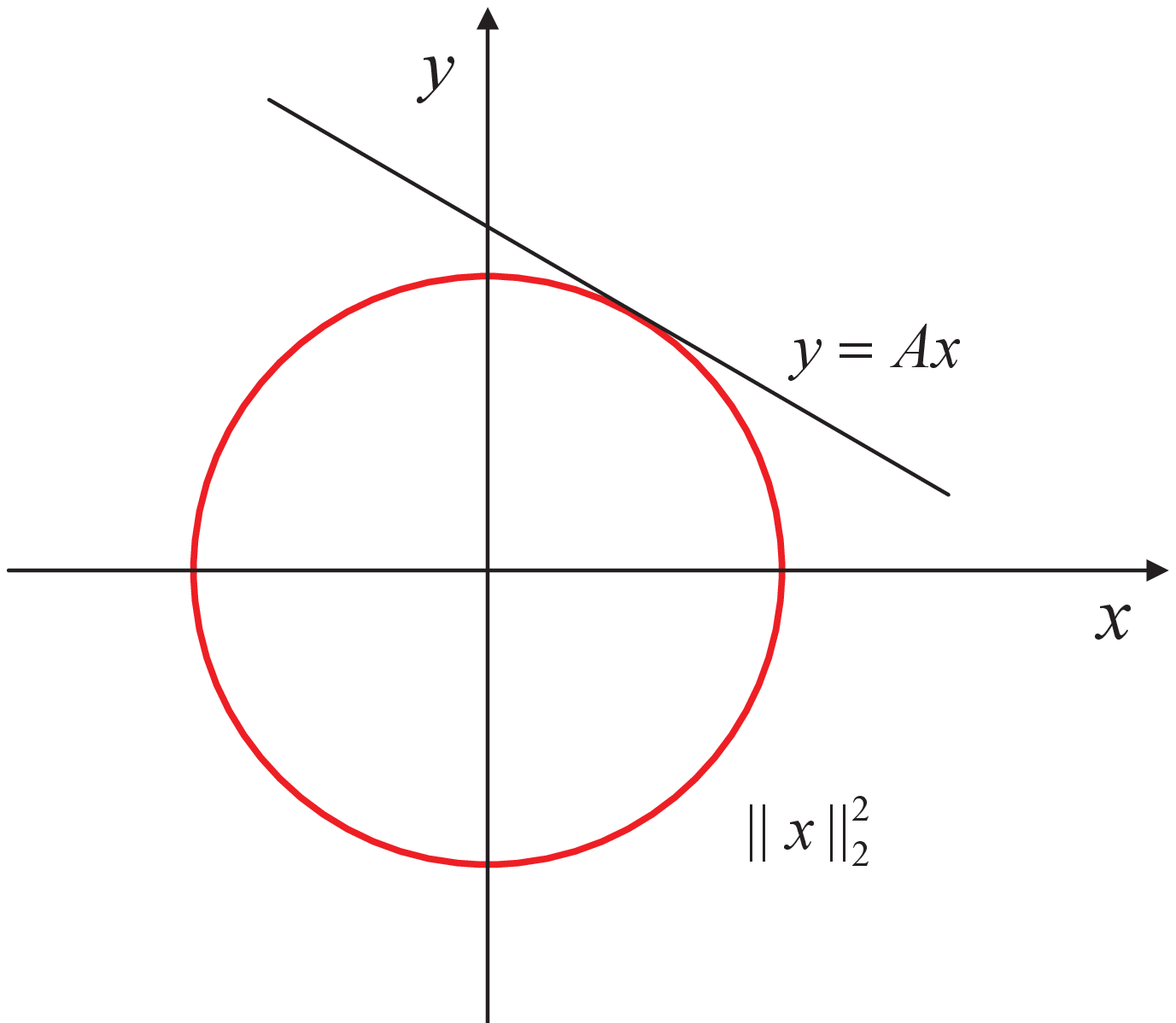}
}
\caption{The geometry of the solutions of different norm regularization in 2-D space \cite{elad2010sparse}. (a), (b) and (c) are the geometry of the solutions of the $l_0$-norm, $l_1$-norm, $l_2$-norm minimization, respectively.}
\label{fig3}
\end{figure*}

On the other hand, it is assumed that $f(x)$ is the function of the $l_p$-norm (p$>$0) on the parameter vector $\bm{x}$, and then the following function is obtained:
\begin{equation}\label{2-4}
f(x)=\|\bm{x}\|_p^p=(\sum_{i=1}^n |\bm{x}_i|^p)
\end{equation}

The relationships between different norms are summarized in Fig. \ref{fig2}. From the illustration in Fig. \ref{fig2}, the conclusions are as follows. The $l_0$-norm function is a nonconvex, nonsmooth, discontinuity, global nondifferentiable function. The $l_p$-norm (0$<$p$<$1) is a nonconvex, nonsmooth, global nondifferentiable function. The $l_1$-norm function is a convex, nonsmooth, global nondifferentiable function. The  $l_2$-norm function is a convex, smooth, global differentiable function.

In order to more specifically elucidate the meaning and solutions of different norm minimizations, the geometry in 2-D space is used to explicitly illustrate the solutions of the $l_0$-norm minimization in Fig. \ref{fig3-a}, $l_1$-norm minimization in Fig. \ref{fig3-b}, and $l_2$-norm minimization in Fig. \ref{fig3-c}. Let $S=\{x^*:Ax=y\}$ denote the line in 2-D space and a hyperplane will be formulated in higher dimensions. All possible solution $x^*$ must lie on the line of $S$. In order to visualize how to obtain the solution of different norm-based minimization problems, we take the $l_1$-norm minimization problem as an example to explicitly interpret. Suppose that we inflate the $l_1$-ball from an original status until it hits the hyperplane $S$ at some point. Thus, the solution of the $l_1$-norm minimization problem is the aforementioned touched point. If the sparse solution of the linear system is localized on the coordinate axis, it will be sparse enough. From the perspective of Fig. \ref{fig3}, it can be seen that the solutions of both the $l_0$-norm and $l_1$-norm minimization are sparse, whereas for the $l_2$-norm  minimization, it is very difficult to rigidly satisfy the condition of sparsity. However, it has been demonstrated that the representation solution of the $l_2$-norm  minimization is not strictly sparse enough but ``limitedly-sparse", which means it possesses the capability of discriminability \cite{zhang2015noise}.

The Frobenius norm, $L_1$-norm of matrix $X \in \mathbb{R}^{m \times n}$, and $l_2$-norm or spectral norm are respectively defined as
\begin{equation} \label{2-5}
\begin{split}
\|X\|_F & =( \sum_{i=1}^n \sum_{j=1}^m X_{j,i}^2 )^{1/2}, \|X\|_{L_1}= max_{j=1,\dots,n} \sum_{i=1}^m |x_{ij}|,\\
\|X\|_2 &=\delta_{max}(X)=(\lambda_{max}(X^TX))^{1/2}
\end{split}
\end{equation}
where $\delta$ is the singular value operator and the $l_2$-norm of $X$ is its maximum singular value \cite{boyd2009convex}.

The $l_{2,1}$-norm or $R_1$-norm is defined on matrix term, that is
\begin{equation} \label{2-6}
\|X\|_{2,1}=\sum_{i=1}^n (\sum_{j=1}^m X_{j,i}^2 )^{1/2}
\end{equation}

As shown above, a norm can be viewed as a measure of the length of a vector $\bm{v}$. The distance between two vectors $\bm{x}$ and $\bm{y}$, or matrices $X$ and $Y$, can be measured by the length of their differences, i.e.
\begin{equation} \label{2-7}
dist(\bm{x},\bm{y})=\|\bm{x}-\bm{y}\|_2^2,~~dist(X,Y)=\|X-Y\|_F
\end{equation}
which are denoted as the distance between $\bm{x}$ and $\bm{y}$ in the context of the $l_2$-norm and the distance between $X$ and $Y$ in the context of the Frobenius norm, respectively.

Assume that $X \in \mathbb{R}^{m \times n}$ and the rank of $X$, i.e. $rank(X)=r$. The SVD of $X$ is computed as
\begin{equation} \label{2-8}
X=U \Lambda V^T
\end{equation}
where $U \in \mathbb{R}^{m \times r}$ with $U^TU=I$ and $V \in \mathbb{R}^{n \times r}$ with $V^TV=I$. The columns of $U$ and $V$ are called left and right singular vectors of $X$, respectively. Additionally, $\Lambda$ is a diagonal matrix and its elements are composed of the singular values of $X$, i.e. $\Lambda = diag(\lambda_1, \lambda_2, \cdots, \lambda_r)$ with $\lambda_1 \geq \lambda_2 \geq \cdots \geq \lambda_r>0$. Furthermore, the singular value decomposition can be rewritten as
\begin{equation} \label{2-9}
X=\sum_{i=1}^r \lambda_i\bm{u}_i\bm{v}_i
\end{equation}
where $\lambda_i$, $\bm{u}_i$ and $\bm{v}_i$ are the $i$-th singular value, the $i$-th column of $U$, and the $i$-th column of $V$, respectively \cite{boyd2009convex}.

\section{Sparse representation problem with different norm regularizations}

\noindent In this section, sparse representation is summarized and grouped into different categories in terms of the norm regularizations used. The general framework of sparse representation is to exploit the linear combination of some samples or ``atoms" to represent the probe sample, to calculate the representation solution, i.e. the representation coefficients of these samples or ``atoms", and then to utilize the representation solution to reconstruct the desired results. The representation results in sparse representation, however, can be greatly dominated by the regularizer (or optimizer) imposed on the representation solution \cite{chen2006theoretical, pant2011unconstrained, chartrand2008iteratively, yang2012beyond}. Thus, in terms of the different norms used in optimizers, the sparse representation methods can be roughly grouped into five general categories: sparse representation with the $l_0$-norm minimization \cite{tropp2007signal, needell2009uniform}, sparse representation with the $l_p$-norm (0$<$p$<$1) minimization \cite{saab2008stable, chartrand2007exact, xu2010data}, sparse representation with the $l_1$-norm minimization \cite{tibshirani1996regre, efron2004least, yang2011alternating, schmidt2007fast}, sparse representation with the $l_{2,1}$-norm minimization \cite{nie2010efficient, yang2011l2, shi2014face, liu2009multi, hou2013joint}, sparse representation with the $l_2$-norm minimization \cite{xu2011two, naseem2010linear, zhang2011sparse}.

\subsection{Sparse representation with $l_0$-norm minimization}
\noindent Let $\bm{x}_1,\bm{x}_2,\cdots,\bm{x}_n \in \mathbb{R}^d$ be all the $n$ known samples and matrix $X \in \mathbb{R}^{d \times n}$ (d$<$n), which is constructed by known samples, is the measurement matrix or the basis dictionary and should also be an over-completed dictionary. Each column of $X$ is one sample and the probe sample is $\bm{y} \in \mathbb{R}^d$ , which is a column vector. Thus, if all the known samples are used to approximately represent the probe sample, it should be expressed as:
\begin{equation} \label{3-1}
\bm{y}=\bm{x}_1{\bm\alpha}_1+\bm{x}_2{\bm\alpha}_2+\cdots+\bm{x}_n{\bm\alpha}_n
\end{equation}
where ${\bm\alpha}_i$ ($i$=1,2,$\cdots$,$n$) is the coefficient of $\bm{x}_i$ and Eq. \ref{3-1} can be rewritten into the following equation for convenient description:
\begin{equation} \label{3-2}
\bm{y}=X{\bm\alpha}
\end{equation}
where matrix $X$=[$\bm{x}_1,\bm{x}_2,\cdots,\bm{x}_n$] and ${\bm\alpha}$=$[{\bm\alpha}_1,{\bm\alpha}_2,\cdots,{\bm\alpha}_n]^T$.

However, problem \ref{3-2} is an underdetermined linear system of equations and the main problem is how to solve it. From the viewpoint of linear algebra, if there is not any prior knowledge or any constraint imposed on the representation solution ${\bm\alpha}$, problem \ref{3-2} is an ill-posed problem and will never have a unique solution. That is, it is impossible to utilize equation \ref{3-2} to uniquely represent the probe sample $y$ using the measurement matrix $X$. To alleviate this difficulty, it is feasible to impose an appropriate regularizer constraint or regularizer function on representation solution ${\bm\alpha}$. The sparse representation method demands that the obtained representation solution should be sparse. Hereafter, the meaning of `sparse' or `sparsity' refers to the condition that when the linear combination of measurement matrix is exploited to represent the probe sample, many of the coefficients should be zero or very close to zero and few of the entries in the representation solution are differentially large.

The sparsest representation solution can be acquired by solving the linear representation system \ref{3-2} with the $l_0$-norm minimization constraint \cite{donoho2003optimally}. Thus problem \ref{3-2} can be converted to the following optimization problem:
\begin{equation} \label{3-3}
\hat{{\bm\alpha}}=\arg\min\|{\bm\alpha}\|_0 ~~~~ s.t. ~~~~  \bm{y}=X{\bm\alpha}
\end{equation}
where $\|\cdot\|_0$ refers to the number of nonzero elements in the vector and is also viewed as the measure of sparsity. Moreover, if just $k$ ($k<n$) atoms from the measurement matrix $X$ are utilized to represent the probe sample, problem \ref{3-3} will be equivalent to the following optimization problem:
\begin{equation} \label{3-4}
\bm{y}=X{\bm\alpha}~~~~s.t.~~~~\|{\bm\alpha}\|_0 \leq k
\end{equation}
Problem \ref{3-4} is called the $k$-sparse approximation problem. Because real data always contains noise, representation noise is unavoidable in most cases. Thus the original model \ref{3-2} can be revised to a modified model with respect to small possible noise by denoting
\begin{equation} \label{3-5}
\bm{y}=X{\bm\alpha}+\bm{s}
\end{equation}
where $\bm{s} \in \mathbb{R}^d$ refers to representation noise and is bounded as $\|\bm{s}\|_2\leq\varepsilon$. With the presence of noise, the sparse solutions of problems \ref{3-3} and \ref{3-4} can be approximately obtained by resolving the following optimization problems:
\begin{equation} \label{3-6}
\hat{{\bm\alpha}}=\arg\min\|{\bm\alpha}\|_0~~~~s.t.~~~~\|\bm{y}-X{\bm\alpha}\|_2^2 \leq \varepsilon
\end{equation}
or
\begin{equation} \label{3-7}
\hat{{\bm\alpha}}=\arg\min\|\bm{y}-X{\bm\alpha}\|_2^2~~~~s.t.~~~~ \|{\bm\alpha}\|_0 \leq \varepsilon
\end{equation}
Furthermore, according to the Lagrange multiplier theorem, a proper constant $\lambda$ exists such that problems \ref{3-6} and \ref{3-7} are equivalent to the following unconstrained minimization problem with a proper value of $\lambda$.
\begin{equation} \label{3-8}
\hat{{\bm\alpha}}=L({\bm\alpha},\lambda)=\arg\min\|\bm{y}-X{\bm\alpha}\|_2^2 + \lambda \|{\bm\alpha}\|_0
\end{equation}
where $\lambda$ refers to the Lagrange multiplier associated with $\|{\bm\alpha}\|_0$.

\subsection{Sparse representation with $l_1$-norm minimization}\label{subsec3-2}
The $l_1$-norm originates from the Lasso problem \cite{tibshirani1996regre, efron2004least} and it has been extensively used to address issues in machine learning, pattern recognition, and statistics \cite{liu2014realistic, patel2011sparse, yuan2014learning}. Although the sparse representation method with $l_0$-norm minimization can obtain the fundamental sparse solution of ${\bm\alpha}$ over the matrix $X$, the problem is still a non-deterministic polynomial-time hard (NP-hard) problem and the solution is difficult to approximate \cite{amaldi1998approximability}. Recent literature \cite{donoho2006most, candes2006stable, candes2006near, wright2009robust} has demonstrated that when the representation solution obtained by using the $l_1$-norm minimization constraint is also content with the condition of sparsity and the solution using $l_1$-norm minimization with sufficient sparsity can be equivalent to the solution obtained by $l_0$-norm minimization with full probability. Moreover, the $l_1$-norm optimization problem has an analytical solution and can be solved in polynomial time. Thus, extensive sparse representation methods with the $l_1$-norm minimization have been proposed to enrich the sparse representation theory. The applications of sparse representation with the $l_1$-norm minimization are extraordinarily and remarkably widespread. Correspondingly, the main popular structures of sparse representation with the $l_1$-norm minimization , similar to sparse representation with $l_0$-norm minimization, are generally used to solve the following problems:

\begin{equation} \label{3-9}
\hat{{\bm\alpha}}=\arg\min_{\bm\alpha}\|{\bm\alpha}\|_1 ~~~~ s.t. ~~~~  \bm{y}=X{\bm\alpha}
\end{equation}

\begin{equation} \label{3-10}
\hat{{\bm\alpha}}=\arg\min_{\bm\alpha}\|{\bm\alpha}\|_1~~~~s.t.~~~~\|\bm{y}-X{\bm\alpha}\|_2^2 \leq \varepsilon
\end{equation}
or
\begin{equation} \label{3-11}
\hat{{\bm\alpha}}=\arg\min_{\bm\alpha}\|\bm{y}-X{\bm\alpha}\|_2^2~~~~s.t.~~~~ \|{\bm\alpha}\|_1 \leq \tau
\end{equation}

\begin{equation} \label{3-12}
\hat{{\bm\alpha}}=L({\bm\alpha},\lambda)=\arg\min_{\bm\alpha} \frac{1}{2} \|y-X{\bm\alpha}\|_2^2 + \lambda \|{\bm\alpha}\|_1
\end{equation}
where $\lambda$ and $\tau$ are both small positive constants.

\subsection{Sparse representation with $l_p$-norm (0$<$p$<$1) minimization}
\noindent The general sparse representation method is to solve a linear representation system with the $l_p$-norm minimization problem. In addition to the $l_0$-norm minimization and $l_1$-norm minimization, some researchers are trying to solve the sparse representation problem with the $l_p$-norm (0$<$p$<$1) minimization, especially $p$ = $0.1$, $\frac{1}{2}$, $\frac{1}{3}$, or $0.9$ \cite{qin2013comparison, xu2012regularization, guo2013enhancing}. That is, the sparse representation problem with the $l_p$-norm (0$<$p$<$1) minimization is to solve the following problem:
\begin{equation} \label{3-13}
\hat{{\bm\alpha}}=\arg\min_{\bm\alpha}\|{\bm\alpha}\|_p^p~~~~s.t.~~~~\|\bm{y}-X{\bm\alpha}\|_2^2 \leq \varepsilon
\end{equation}
or
\begin{equation} \label{3-14}
\hat{{\bm\alpha}}=L({\bm\alpha},\lambda)=\arg\min_{\bm\alpha} \|\bm{y}-X{\bm\alpha}\|_2^2 + \lambda \|{\bm\alpha}\|_p^p
\end{equation}
In spite of the fact that sparse representation methods with the $l_p$-norm (0$<$p$<$1) minimization are not the mainstream methods to obtain the sparse representation solution, it tremendously influences the improvements of the sparse representation theory.

\subsection{Sparse representation with $l_{2,1}$-norm minimization}

\noindent The representation solution obtained by the $l_2$-norm minimization is not rigorously sparse. It can only obtain a `limitedly-sparse' representation solution, i.e. the solution has the property that it is discriminative and distinguishable but is not really sparse enough \cite{zhang2015noise}. The objective function of the sparse representation method with the $l_2$-norm minimization is to solve the following problem:

\begin{equation} \label{3-15}
\hat{{\bm\alpha}}=\arg\min_{\bm\alpha}\|{\bm\alpha}\|_2^2~~~~s.t.~~~~\|\bm{y}-X{\bm\alpha}\|_2^2 \leq \varepsilon
\end{equation}
or
\begin{equation} \label{3-16}
\hat{{\bm\alpha}}=L({\bm\alpha},\lambda)=\arg\min_{\bm\alpha} \|\bm{y}-X{\bm\alpha}\|_2^2 + \lambda \|{\bm\alpha}\|_2^2
\end{equation}

On the other hand, the $l_{2,1}$-norm is also called the rotation invariant $l_1$-norm, which is proposed to overcome the difficulty of robustness to outliers \cite{ding2006r}. The objective function of the sparse representation problem with the $l_{2,1}$-norm minimization is to solve the following problem:
\begin{equation} \label{3-17}
\arg\min_{A} \|Y-XA\|_{2,1} + \mu \|A\|_{2,1}
\end{equation}
where $Y=[\bm{y}_1,\bm{y}_2,\cdots,\bm{y}_N]$ refers to the matrix composed of samples, $A=[\bm{a}_1,\bm{a}_2,\cdots,\bm{a}_N]$ is the corresponding coefficient matrix of $X$, and $\mu$ is a small positive constant. Sparse representation with the $l_{2,1}$-norm minimization can be implemented by exploiting the proposed algorithms in literature \cite{nie2010efficient, yang2011l2, shi2014face}.

\section{Greedy strategy approximation}
\noindent Greedy algorithms date back to the 1950s. The core idea of the greedy strategy \cite{elad2010sparse, tropp2006a} is to determine the position based on the relationship between the atom and probe sample, and then to use the least square to evaluate the amplitude value. Greedy algorithms can obtain the local optimized solution in each step in order to address the problem. However, the greedy algorithm can always produce the global optimal solution or an approximate overall solution \cite{elad2010sparse, tropp2006a}. Addressing sparse representation with $l_0$-norm regularization, i.e. problem \ref{3-3}, is an NP hard problem \cite{wright2009robust, donoho2006most}. The greedy strategy provides a special way to obtain an approximate sparse representation solution. The greedy strategy actually can not directly solve the optimization problem and it only seeks an approximate solution for problem \ref{3-3}.

\subsection{Matching pursuit algorithm}
\noindent
The matching pursuit (MP) algorithm \cite{mallat1993matching} is the earliest and representative method of using the greedy strategy to approximate problem \ref{3-3} or \ref{3-4}. The main idea of the MP is to iteratively choose the best atom from the dictionary based on a certain similarity measurement to approximately obtain the sparse solution. Taking as an example of the sparse decomposition with a vector sample $y$  over the over-complete dictionary $D$, the detailed algorithm description is presented as follows:

Suppose that the initialized representation residual is $\bm{R}_0=\bm{y}$, $D=[\bm{d}_1, \bm{d}_2, \cdots, \bm{d}_N] \in \mathbb{R}^{d \times N}$ and each sample in dictionary $D$ is an $l_2$-norm unity vector, i.e. $\|\bm{d}_i\|=1$. To approximate $\bm{y}$, MP first chooses the best matching atom from $D$ and the selected atom should satisfy the following condition:

\begin{equation} \label{4-1}
|\langle \bm{R}_0 , \bm{d}_{l_0} \rangle| = sup |\langle \bm{R}_0 , \bm{d}_i \rangle|
\end{equation}
where $l_0$ is a label index from dictionary $D$. Thus $\bm{y}$ can be decomposed into the following equation:
\begin{equation} \label{4-2}
\bm{y}= \langle \bm{y} , \bm{d}_{l_0} \rangle  \bm{d}_{l_0} + \bm{R}_1
\end{equation}
So $\bm{y}= \langle \bm{R}_0 , \bm{d}_{l_0} \rangle  \bm{d}_{l_0} + \bm{R}_1$ where $\langle \bm{R}_0 , \bm{d}_{l_0} \rangle  \bm{d}_{l_0}$ represents the orthogonal projection of $\bm{y}$ onto $\bm{d}_{l_0}$, and $\bm{R}_1$ is the representation residual by using $\bm{d}_{l_0}$ to represent $\bm{y}$. Considering the fact that $\bm{d}_{l_0}$ is orthogonal to $\bm{R_1}$, Eq. \ref{4-2} can be rewritten as
\begin{equation} \label{4-3}
\|\bm{y}\|^2= | \langle \bm{y} , \bm{d}_{l_0} \rangle |^2  + \| \bm{R}_1\| ^2
\end{equation}

To obtain the minimum representation residual, the MP algorithm iteratively figures out the best matching atom from the over-completed dictionary, and then utilizes the representation residual as the next approximation target until the termination condition of iteration is satisfied. For the $t$-th iteration, the best matching atom is $\bm{d}_{l_t}$ and the approximation result is found from the following equation:
\begin{equation} \label{4-4}
\bm{R}_t = \langle \bm{R}_t , \bm{d}_{l_t} \rangle \bm{d}_{l_t}  + \bm{R}_{t+1}
\end{equation}
where the $\bm{d}_{l_t}$ satisfies the equation:
\begin{equation} \label{4-5}
|\langle \bm{R}_t , \bm{d}_{l_t} \rangle| = sup |\langle \bm{R}_t , \bm{d}_i \rangle|
\end{equation}

Clearly, $\bm{d}_{l_t}$ is orthogonal to $\bm{R_{k+1}}$, and then
\begin{equation} \label{4-6}
\|\bm{R}_{k}\|^2= | \langle \bm{R}_t , \bm{d}_{l_t} \rangle |^2  + \| \bm{R}_{t+1}\| ^2
\end{equation}

For the $n$-th iteration, the representation residual $\|\bm{R}_n\|^2 \leq \tau$ where $\tau$ is a very small constant and the probe sample $\bm{y}$ can be formulated as:
\begin{equation} \label{4-7}
\bm{y}= \sum_{j=1}^{n-1} \langle \bm{R}_j , \bm{d}_{l_j} \rangle \bm{d}_{l_j} + \bm{R}_n
\end{equation}

If the representation residual is small enough, the probe sample $\bm{y}$ can approximately satisfy the following equation: $ \bm{y} \approx \sum_{j=1}^{n-1} \langle \bm{R}_j, \bm{d}_{l_j} \rangle \bm{d}_{l_j} $ where $n\ll N$. Thus, the probe sample can be represented by a small number of elements from a large dictionary. In the context of the specific representation error, the termination condition of sparse representation is that the representation residual is smaller than the presupposed value. More detailed analysis on matching pursuit algorithms can be found in the literature \cite{mallat1993matching}.

\subsection{Orthogonal matching pursuit algorithm}
\noindent The orthogonal matching pursuit (OMP) algorithm \cite{pati1993orthogonal, tropp2007signal} is an improvement of the MP algorithm. The OMP employs the process of orthogonalization to guarantee the orthogonal direction of projection in each iteration. It has been verified that the OMP algorithm can be converged in limited iterations \cite{tropp2007signal}. The main steps of OMP algorithm have been summarized in Algorithm 1.

\begin{table}[htbp]
\centering
\begin{tabular}{m{85mm}}
\toprule
\textbf{Algorithm 1.} Orthogonal matching pursuit algorithm\\
\textbf{Task:} Approximate the constraint problem:\\~~~~~~~ $\hat{{\bm\alpha}}=\arg\min_{{\bm\alpha}}\|{\bm\alpha}\|_0 ~~ s.t. ~~  \bm{y}=X{\bm\alpha}$\\
\midrule
\textbf{Input:} Probe sample $\bm{y}$, measurement matrix $X$, sparse coefficients vector ${\bm\alpha}$ \\
\textbf{Initialization:} $t=1$, $\bm{r}_0=y$, ${\bm\alpha}=0$, $D_0=  \phi$, index set $\Lambda_0= \phi$ where $\phi$ denotes empty set, $\tau$ is a small constant.\\

While  $\|\bm{r}_t\|> \tau$  do\\
~~~~Step 1:~Find the best matching sample, i.e. the biggest inner product \\~~~~~~~~~~ between $\bm{r}_{t-1}$ and $\bm{x}_j$ ($j\not\in \Lambda_{t-1}$) by exploiting \\~~~~~~~~~~~$\lambda_t=\arg\max_{j\not\in \Lambda_{t-1}} |\langle \bm{r}_{t-1}, \bm{x}_{j}\rangle|$.\\
~~~Step 2:~Update the index set $\Lambda_t=\Lambda_{t-1}\bigcup \lambda_t$ and reconstruct data set\\~~~~~~~~~~ $D_t=[D_{t-1},\bm{x}_{\lambda_t}]$.\\
~~~Step 3:~Compute the sparse coefficient by using the least square algorithm\\ ~~~~~~~~~~ $\tilde{{\bm\alpha}}=\arg\min\|\bm{y}-D_t\tilde{{\bm\alpha}}\|_2^2$.\\
~~~Step 4:~Update the representation residual using $\bm{r}_t=\bm{y}-D_t\tilde{{\bm\alpha}}$.\\
~~~Step 5:~$t=t+1$.\\
End \\
\textbf{Output:} $D$, ${\bm\alpha}$\\
\bottomrule
\end{tabular}
\end{table}

\subsection{Series of matching pursuit algorithms}
\noindent It is an excellent choice to employ the greedy strategy to approximate the solution of sparse representation with the $l_0$-norm minimization. These algorithms are typical greedy iterative algorithms. The earliest algorithms were the matching pursuit (MP) and orthogonal matching pursuit (OMP). The basic idea of the MP algorithm is to select the best matching atom from the overcomplete dictionary to construct sparse approximation during each iteration, to compute the signal representation residual, and then to choose the best matching atom till the stopping criterion of iteration is satisfied. Many more greedy algorithms based on the MP and OMP algorithm such as the efficient orthogonal matching pursuit algorithm \cite{vitaladevuni2011efficient} subsequently have been proposed to improve the pursuit algorithm. Needell et al. proposed an regularized version of orthogonal matching pursuit (ROMP) algorithm \cite{needell2009uniform}, which recovered all $k$ sparse signals based on the Restricted Isometry Property of random frequency measurements, and then proposed another variant of OMP algorithm called compressive sampling matching pursuit (CoSaMP) algorithm \cite{needell2009cosamp}, which incorporated several existing ideas such as restricted isometry property (RIP) and pruning technique into a greedy iterative structure of OMP. Some other algorithms also had an impressive influence on future research on CS. For example, Donoho et al. proposed an extension of OMP, called stage-wise orthogonal matching pursuit (StOMP) algorithm \cite{donoho2012sparse}, which depicted an iterative algorithm with three main steps, i.e.  threholding, selecting and projecting. Dai and Milenkovic proposed a new method for sparse signal reconstruction named subspace pursuit (SP) algorithm \cite{dai2009subspace}, which sampled signals satisfying the constraints of the RIP with a constant parameter. Do et al. presented a sparsity adaptive matching pursuit (SAMP) algorithm \cite{do2008sparsity}, which borrowed the idea of the EM algorithm to alternatively estimate the sparsity and support set. Jost et al. proposed a tree-based matching pursuit (TMP) algorithm \cite{jost2006tree}, which constructed a tree structure and employed a structuring strategy to cluster similar signal atoms from a highly redundant dictionary as a new dictionary. Subsequently, La and Do proposed a new tree-based orthogonal matching pursuit (TBOMP) algorithm \cite{la2006tree}, which treated the sparse tree representation as an additional prior knowledge for linear inverse systems by using a small number of samples. Recently, Karahanoglu and Erdogan conceived a forward-backward pursuit (FBP) method \cite{karahanoglu2013com} with two greedy stages, in which the forward stage enlarged the support estimation and the backward stage removed some unsatisfied atoms. More detailed treatments of the greedy pursuit for sparse representation can be found in the literature \cite{tropp2006a}.

\section{Constrained optimization strategy}
\noindent Constrained optimization strategy is always utilized to obtain the solution of sparse representation with the $l_1$-norm regularization. The methods that address the non-differentiable unconstrained problem will be presented by reformulating it as a smooth differentiable constrained optimization problem. These methods exploit the constrained optimization method with efficient convergence to obtain the sparse solution. What is more, the constrained optimization strategy emphasizes the equivalent transformation of $\|{\bm\alpha}\|_1$ in problem \ref{3-12} and employs the new reformulated constrained problem to obtain a sparse representation solution. Some typical methods that employ the constrained optimization strategy to solve the original unconstrained non-smooth problem are introduced in this section.

\subsection{Gradient Projection Sparse Reconstruction}
\noindent The core idea of the gradient projection sparse representation method is to find the sparse representation solution along with the gradient descent direction. The first key procedure of gradient projection sparse reconstruction (GPSR) \cite{figueiredo2007gra} provides a constrained formulation where each value of ${\bm\alpha}$ can be split into its positive and negative parts. Vectors ${\bm\alpha}_+$ and ${\bm\alpha}_-$ are introduced to denote the positive and negative coefficients of ${\bm\alpha}$, respectively. The sparse representation solution ${\bm\alpha}$  can be formulated as:

\begin{equation} \label{5-1}
{\bm\alpha} = {\bm\alpha}_+-{\bm\alpha}_-,~ {\bm\alpha}_+\geq 0,~ {\bm\alpha}_-\geq 0
\end{equation}
where the operator $(\cdot)_+$ denotes the positive-part operator, which is defined as $(x)_+$=$\max\{0,x\}$. Thus, $\|{\bm\alpha}\|_1 = \mathbf{1}_d^T {\bm\alpha}_+ + \mathbf{1}_d^T {\bm\alpha}_-$, where $\mathbf{1}_d=[\underbrace{1,1,\cdots,1}_{d}]^T$ is a $d$--dimensional vector with $d$ ones. Accordingly, problem \ref{3-12} can be reformulated as a constrained quadratic problem:
\begin{equation} \label{5-2}
\begin{split}
\arg\min L({\bm\alpha})=\arg\min \frac{1}{2} \|\bm{y}-X[{\bm\alpha}_+-{\bm\alpha}_-]\|_2^2 +\\  \lambda (\mathbf{1}_d^T {\bm\alpha}_+ + \mathbf{1}_d^T {\bm\alpha}_-)~~~~s.t.~~~~{\bm\alpha}_+\geq 0,~ {\bm\alpha}_-\geq 0
\end{split}
\end{equation}
or
\begin{equation} \label{5-3}
\begin{split}
\arg\min L({\bm\alpha})=\arg\min \frac{1}{2} \|\bm{y}-[X_+,X_-][{\bm\alpha}_+-{\bm\alpha}_-]\|_2^2 \\ + \lambda (\mathbf{1}_d^T {\bm\alpha}_+ + \mathbf{1}_d^T {\bm\alpha}_-)~~~~s.t.~~~~{\bm\alpha}_+\geq 0,~ {\bm\alpha}_-\geq 0
\end{split}
\end{equation}
Furthermore, problem \ref{5-3} can be rewritten as:
\begin{equation} \label{5-4}
\arg\min~G(\bm{z})=\bm{c}^T \bm{z} + \frac{1}{2} {\bm z}^T A \bm{z} ~~~~s.t.~~~~ \bm{z}\geq \bm{0}
\end{equation}
where $\bm{z}=[{\bm\alpha}_+;{\bm\alpha}_-]$, $\bm{c} = \lambda \mathbf{1}_{2d} + [-X^T\bm{y}; X^T\bm{y}]$,\\ $\mathbf{1}_{2d}=[\underbrace{1,\cdots,1}_{2d}]^T$, $A=\left( \begin{array}{ccc} X^TX&-X^TX \\ -X^TX&X^TX \\ \end{array} \right)$.

The GPSR algorithm employs the gradient descent and standard line-search method \cite{boyd2009convex} to address problem \ref{5-4}. The value of $z$  can be iteratively obtained by utilizing
\begin{equation} \label{5-5}
\arg\min~{\bm z}^{t+1} = {\bm z}^t - \sigma \nabla G({\bm z}^t)
\end{equation}
where the gradient of $\nabla G({\bm z}^t)= \bm{c} + A{\bm z}^t$ and $\sigma$ is the step size of the iteration. For step size $\sigma$, GPSR updates the step size by using
\begin{equation} \label{5-6}
\sigma^t=\arg\min_{\sigma}~{G({\bm z}^t - \sigma g^t)}
\end{equation}
where the function $g^t$ is pre-defined as

\begin{eqnarray} \label{5-7}
g_i^t =
\left\{
\begin{array}{ll}
(\nabla G({\bm z}^t))_i,& if~~ \bm{z}_i^t > 0~~ or ~~(\nabla G({\bm z}^t))_i < 0 \\
0, & otherwise.\\
\end{array}
\right.
\end{eqnarray}
Problem \ref{5-6} can be addressed with the close-form solution
\begin{equation} \label{5-8}
\sigma^t=\frac{(g^t)^T(g^t)}{(g^t)^TA(g^t)}
\end{equation}
Furthermore, the basic GPSR algorithm employs the backtracking linear search method \cite{boyd2009convex} to ensure that the step size of gradient descent, in each iteration, is a more proper value. The stop condition of the backtracking linear search should satisfy
\begin{equation} \label{5-9}
\begin{split}
G(({\bm z}^t - \sigma^t \nabla G({\bm z}^t))_+) > G({\bm z}^t)- \beta \nabla G({\bm z}^t)^T \\ ({\bm z}^t-({\bm z}^t - \sigma^t \nabla G({\bm z}^t))_+)
\end{split}
\end{equation}
where $\beta$ is a small constant. The main steps of GPSR are summarized in Algorithm 2. For more detailed information, one can refer to the literature \cite{figueiredo2007gra}.

\begin{table}[htbp]
\centering
\begin{tabular}{m{86.5mm}}
\toprule
\textbf{Algorithm 2.} Gradient Projection Sparse Reconstruction (GPSR)\\
\textbf{Task:} To address the unconstraint problem:\\~~~~~~~~
 $\hat{{\bm\alpha}}=\arg\min_{{\bm\alpha}} \frac{1}{2} \|\bm{y}-X{\bm\alpha}\|_2^2 + \lambda \|{\bm\alpha}\|_1$\\
\midrule
\textbf{Input:} Probe sample $y$, the measurement matrix $X$, small constant $\lambda$ \\
\textbf{Initialization:} $t=0$, $\beta \in (0,0.5)$, $\gamma \in (0,1)$, given ${\bm\alpha}$ so that $\bm{z}=[{\bm\alpha}_+,{\bm\alpha}_-]$.\\

While not converged do\\
~~~Step 1:~Compute $\sigma^t$ exploiting Eq. \ref{5-8} and $\sigma^t \leftarrow mid(\sigma_{min}, \sigma^t, \sigma_{max})$,\\~~~~~~~~~~~ where $mid(\cdot,\cdot,\cdot)$ denotes the middle value of the three parameters.\\
~~~Step 2:~While Eq. \ref{5-9} not satisfied \\~~~~~~~~~~~~~~do~~~$\sigma^t \leftarrow \gamma \sigma^t$~~~end\\
~~~Step 3:~${\bm z}^{t+1}= ({\bm z}^t - \sigma^t \nabla G({\bm z}^t))_+$ ~and~$t=t+1$.\\
End \\
\textbf{Output:} ${\bm z}^{t+1}$, ${\bm\alpha}$\\
\bottomrule
\end{tabular}
\end{table}

\subsection{Interior-point method based sparse representation strategy}
\noindent The Interior-point method \cite{boyd2009convex} is not an iterative algorithm but a smooth mathematic model and it always incorporates the Newton method to efficiently solve unconstrained smooth problems of modest size \cite{Parikh2013Proximal}. When the Newton method is used to address the optimization issue, a complex Newton equation should be solved iteratively which is very time-consuming. A method named the truncated Newton method can effectively and efficiently obtain the solution of the Newton equation. A prominent algorithm called the truncated Newton based interior-point method (TNIPM) exists, which can be utilized to solve the large-scale $l_1$-regularized least squares (i.e. $l_1\_l_s$) problem \cite{kim2007an}.

The original problem of $l_1\_l_s$ is to solve problem \ref{3-12} and the core procedures of $l_1\_l_s$ are shown below:\\
(1) Transform the original unconstrained non-smooth problem to a constrained smooth optimization problem.\\
(2) Apply the interior-point method to reformulate the constrained smooth optimization problem as a new unconstrained smooth optimization problem.\\
(3) Employ the truncated Newton method to solve this unconstrained smooth problem.

The main idea of the $l_1\_l_s$ will be briefly described. For simplicity of presentation, the following one-dimensional problem is used as an example.
\begin{equation} \label{5-10}
|\alpha|= \arg\min_{-\sigma \leq \alpha \leq \sigma} \sigma
\end{equation}
where $\sigma$ is a proper positive constant.

Thus, problem \ref{3-12} can be rewritten as
\begin{equation} \label{5-11}
\begin{array}{lll}
\hat {\bm\alpha} &=& \arg \min \frac{1}{2}\|\bm{y}-X {\bm\alpha} \|_2^2 + \lambda \|{\bm\alpha}\|_1 \\ &=& \arg\min  \frac{1}{2}\|\bm{y}-X {\bm\alpha} \|_2^2 + \lambda \sum_{i=1}^{N} \min_{-\bm{\sigma}_i \leq {\bm\alpha}_i \leq \bm{\sigma}_i} \bm{\sigma}_i\\
&=& \arg\min  \frac{1}{2}\|\bm{y}-X {\bm\alpha} \|_2^2 + \lambda \min_{-\bm{\sigma}_i \leq {\bm\alpha}_i \leq \bm{\sigma}_i} \sum_{i=1}^N\bm{\sigma}_i\\
&=& \arg\min_{-\bm{\sigma}_i \leq {\bm\alpha}_i \leq \bm{\sigma}_i}  \frac{1}{2}\|\bm{y}-X {\bm\alpha} \|_2^2 + \lambda \sum_{i=1}^{N} \bm{\sigma}_i\\
\end{array}
\end{equation}

Thus problem \ref{3-12} is also equivalent to solve the following problem:
\begin{equation} \label{5-12}
\hat {\bm\alpha} = \arg\min_{{\bm\alpha},\bm{\sigma} \in \mathbb{R}^N}  \frac{1}{2}\|\bm{y}-X {\bm\alpha} \|_2^2 + \lambda \sum_{i=1}^N\bm{\sigma}_i~s.t.~-\bm{\sigma}_i \leq {\bm\alpha}_i \leq \bm{\sigma}_i
\end{equation}
or
\begin{equation} \label{5-13}
\begin{split}
\hat {\bm\alpha} = \arg\min_{{\bm\alpha},\bm{\sigma} \in \mathbb{R}^N}  \frac{1}{2}\|\bm{y}-X {\bm\alpha} \|_2^2 + \lambda \sum_{i=1}^N\bm{\sigma}_i~~~~\\ s.t.~~~\bm{\sigma}_i + {\bm\alpha}_i \geq 0, ~ \bm{\sigma}_i - {\bm\alpha}_i \geq 0
\end{split}
\end{equation}
The interior-point strategy can be used to transform problem \ref{5-13} into an unconstrained smooth problem
\begin{equation} \label{5-14}
\hat {\bm\alpha} = \arg\min_{{\bm\alpha},\bm{\sigma} \in \mathbb{R}^N} G({\bm\alpha},\bm{\sigma}) =  \frac{v}{2}\|\bm{y}-X {\bm\alpha} \|_2^2 + \lambda v \sum_{i=1}^N\bm{\sigma}_i - B({\bm\alpha},\bm{\sigma})
\end{equation}
where $B({\bm\alpha},\bm{\sigma})=\sum_{i=1}^N log(\bm{\sigma}_i + {\bm\alpha}_i) + \sum_{i=1}^N log(\bm{\sigma}_i - {\bm\alpha}_i)$ is a barrier function, which forces the algorithm to be performed within the feasible region in the context of unconstrained condition.

Subsequently, $l_1\_l_s$ utilizes the truncated Newton method to solve problem \ref{5-14}. The main procedures of addressing problem \ref{5-14} are presented as follows:\\
First, the Newton system is constructed
\begin{equation} \label{5-15}
H \left[ \begin{array}{l} \triangle {\bm\alpha}\\ \triangle \bm{\sigma} \end{array} \right] = -\nabla G({\bm\alpha},\bm{\sigma}) \in \mathbb{R}^{2N}
\end{equation}
where $H = -\nabla^2 G({\bm\alpha},\bm{\sigma}) \in \mathbb{R}^{2N \times 2N}$ is the Hessian matrix, which is computed using the preconditioned conjugate gradient algorithm, and then the direction of linear search $[\triangle {\bm\alpha}, \triangle \bm{\sigma}]$ is obtained.\\
Second, the Lagrange dual of problem \ref{3-12} is used to construct the dual feasible point and duality gap:\\
a) The Lagrangian function and Lagrange dual of problem \ref{3-12} are constructed. The Lagrangian function is reformulated as
\begin{equation} \label{5-16}
L({\bm\alpha},\bm{z},\bm{u}) = {\bm z}^T\bm{z} + \lambda \|{\bm\alpha}\|_1 + u(X{\bm\alpha} - \bm{y} -\bm{z})
\end{equation}
where its corresponding Lagrange dual function is
\begin{equation} \label{5-17}
\begin{split}
\hat{{\bm\alpha}}= \arg\max F(\bm{u}) = - \frac{1}{4} \bm{u}^T\bm{u} - \bm{u}^T\bm{y} ~~~~s.t.\\ |(X^T\bm{u})_i| \leq \lambda_i~(i=1,2,\cdots,N)
\end{split}
\end{equation}
b) A dual feasible point is constructed
\begin{equation} \label{5-18}
\bm{u}=2s(\bm{y}-X{\bm\alpha}),~s = \min \{\lambda / |2\bm{y}_i - 2(X^TX{\bm\alpha})_i|\}\forall i
\end{equation}
where $u$ is a dual feasible point and $s$ is the step size of the linear search.\\
c) The duality gap is constructed, which is the gap between the primary problem and the dual problem:
\begin{equation} \label{5-19}
g=\|\bm{y}-X{\bm\alpha}\| + \lambda \|{\bm\alpha}\|_1 - F(\bm{u})
\end{equation}
Third, the method of backtracking linear search is used to determine an optimal step size of the Newton linear search. The stopping condition of the backtracking linear search is
\begin{equation} \label{5-20}
G({\bm\alpha} + \eta^t \triangle {\bm\alpha}, \bm{\sigma} + \eta^t \triangle \bm{\sigma}) > G({\bm\alpha}, \bm{\sigma}) + \rho \eta^t \nabla G({\bm\alpha}, \bm{\sigma}) [\triangle {\bm\alpha}, \triangle \bm\sigma]
\end{equation}
where $\rho \in (0,0.5)$ and $\eta^t \in (0,1)$  is the step size of the Newton linear search.\\
Finally, the termination condition of the Newton linear search is set to
\begin{equation} \label{5-21}
\zeta = \min \{0.1, \beta g / \|h\|_2\}
\end{equation}
where the function $h = \nabla G({\bm\alpha}, \bm{\sigma})$, $\beta$ is a small constant, and $g$ is the duality gap. The main steps of algorithm $l_1\_l_s$ are summarized in Algorithm 3. For further description and analyses, please refer to the literature \cite{kim2007an}.
\begin{table}[htbp]
\centering
\begin{tabular}{m{85mm}}
\toprule
\textbf{Algorithm 3.} Truncated Newton based interior-point method (TNIPM) for $l_1\_l_s$\\
\textbf{Task:} To address the unconstraint problem:\\~~~~~~~$\hat{{\bm\alpha}}=\arg\min_{{\bm\alpha}} \frac{1}{2} \|\bm{y}-X{\bm\alpha}\|_2^2 + \lambda \|{\bm\alpha}\|_1$\\
\midrule
\textbf{Input:} Probe sample $\bm{y}$, the measurement matrix $X$, small constant $\lambda$ \\
\textbf{Initialization:} $t=1$, $v= \frac{1}{\lambda}$, $\rho \in (0,0.5)$, $\sigma = \mathbf{1}_N$\\
~~~Step 1:~ Employ preconditioned conjugate gradient algorithm to obtain the approximation of $H$ in Eq. \ref{5-15}, and then obtain the descent direction of linear search $[\triangle {\bm\alpha}^t, \triangle \bm{\sigma}^t]$.\\
~~~Step 2:~Exploit the algorithm of backtracking linear search to find the optimal step size of Newton linear search $\eta^t$, which satisfies the Eq. \ref{5-20}.\\
~~~Step 3:~Update the iteration point utilizing $({\bm\alpha}^{t+1},\bm{\sigma}^{t+1}) = ({\bm\alpha}^t, \bm{\sigma}^t) + (\triangle {\bm\alpha}^t + \triangle \bm{\sigma}^t)$.\\
~~~Step 4:~Construct feasible point using eq. \ref{5-18} and duality gap in Eq. \ref{5-19}, and compute the termination tolerance $\zeta$ in Eq. \ref{5-21}.\\
~~~Step 5:~If the condition $g/F(\bm{u})>\zeta$ is satisfied, stop; Otherwise, return to step 1, update $v$ in Eq. \ref{5-14} and $t=t+1$.\\
\textbf{Output:} ${\bm\alpha}$\\
\bottomrule
\end{tabular}
\end{table}

The truncated Newton based interior-point method (TNIPM) \cite{portugal2000trun} is a very effective method to solve the $l_1$-norm regularization problems. Koh et al. \cite{koh2007interior} also utilized the TNIPM to solve large scale logistic regression problems, which employed a preconditioned conjugate gradient method to compute the search step size with warm-start techniques. Mehrotra proposed to exploit the interior-point method to address the primal-dual problem \cite{mehrotra1992imp} and introduced the second-order derivation of Taylor polynomial to approximate a primal-dual trajectory. More analyses of interior-point method for sparse representation can be found in the literature \cite{wright1997primal}.

\subsection{Alternating direction method (ADM) based sparse representation strategy}
\noindent This section shows how the ADM \cite{yang2011alternating} is used to solve primal and dual problems in \ref{3-12}. First, an auxiliary variable is introduced to convert problem in \ref{3-12} into a constrained problem with the form of problem \ref{5-22}. Subsequently, the alternative direction method is used to efficiently address the sub-problems of problem \ref{5-22}. By introducing the auxiliary term $\bm{s} \in \mathbb{R}^d$, problem \ref{3-12} is equivalent to a constrained problem
\begin{equation} \label{5-22}
\arg\min_{{\bm\alpha},s} \frac{1}{2\tau} \|\bm{s}\|_2 + \|{\bm\alpha}\|_1~~~~s.t.~~~~\bm{s}=\bm{y}-X{\bm\alpha}
\end{equation}
The optimization problem of the augmented Lagrangian function of problem \ref{5-22} is considered
\begin{equation} \label{5-23}
\begin{split}
\arg\min_{{\bm\alpha},\bm{s},\lambda} L({\bm\alpha},\bm{s},\bm{\lambda}) = \frac{1}{2\tau} \|\bm{s}\|_2 + \|{\bm\alpha}\|_1 -\lambda^T (\bm{s}+ \\ X{\bm\alpha}-y) + \frac{\mu}{2}\|\bm{s}+X{\bm\alpha}-\bm{y}\|_2^2
\end{split}
\end{equation}
where $\bm{\lambda} \in \mathbb{R}^d$ is a Lagrange multiplier vector and $\mu$ is a penalty parameter. The general framework of ADM is used to solve problem \ref{5-23} as follows:

\begin{eqnarray} \label{5-24}
\left\{
\begin{array}{llll}
\bm{s}^{t+1}&=&\arg\min L(\bm{s},{\bm\alpha}^t,\bm{\lambda}^t)~~~~~~~~&(a)\\
{\bm\alpha}^{t+1}&=&\arg\min L(\bm{s}^{t+1},{\bm\alpha},\bm{\lambda}^t)&(b)\\
\bm{\lambda}^{t+1}&=&\bm{\lambda}^t - \mu(\bm{s}^{t+1}+X{\bm\alpha}^{t+1}-\bm{y})~~~~&(c)\\
\end{array}
\right.
\end{eqnarray}
First, the first optimization problem \ref{5-24}(a) is considered
\begin{equation} \label{5-25}
\begin{split}
\arg\min L(\bm{s}, {\bm\alpha}^t,\bm{\lambda}^t) =& \frac{1}{2\tau} \|\bm{s}\|_2 + \| {\bm\alpha}^t \|_1 - (\bm{\lambda}^t)^T (\bm{s} + X {\bm\alpha}^t\\&-\bm{y}) + \frac{\mu}{2}\|\bm{s}+X{\bm\alpha}^t-\bm{y}\|_2^2 \\
= \frac{1}{2\tau} \|\bm{s}\|_2 &-(\bm{\lambda}^t)^T\bm{s}+ \frac{\mu}{2}\|\bm{s}+X{\bm\alpha}^t-\bm{y}\|_2^2 + \\ & ~~~~~\| {\bm\alpha}^t \|_1 - (\bm{\lambda}^t)^T (X{\bm\alpha}^t-\bm{y})\\
\end{split}
\end{equation}
Then, it is known that the solution of problem \ref{5-25} with respect to $s$ is given by
\begin{equation} \label{5-26}
\bm{s}^{t+1}=\frac{\tau}{1+\mu\tau} (\bm{\lambda}^t - \mu (\bm{y}-X{\bm\alpha}^t))
\end{equation}
Second, the optimization problem \ref{5-24}(b) is considered
\[ \begin{split} \arg\min L(\bm{s}^{t+1},{\bm\alpha},\lambda^t) &= \frac{1}{2\tau} \|\bm{s}^{t+1}\|_2 + \| {\bm\alpha}\|_1 -(\bm{\lambda})^T (\bm{s}^{t+1}\\ & +X{\bm\alpha}-\bm{y}) + \frac{\mu}{2}\|\bm{s}^{t+1}+X{\bm\alpha}-\bm{y}\|_2^2 \end{split} \]
which is equivalent to

\begin{equation} \label{5-27}
\begin{split}
\arg\min\{& \|{\bm\alpha} \|_1-(\bm{\lambda}^t)^T (\bm{s}^{t+1}+ X{\bm\alpha}-\bm{y}) + \frac{\mu}{2} \| \bm{s}^{t+1}+ \\ &~~~~~~~~~~~~~~~~~~~~~~~~~~~~~~~~~~~~ X{\bm\alpha}-\bm{y}\|_2^2\}
\\&= {\| {\bm\alpha}\|_1 + \frac{\mu}{2}\|\bm{s}^{t+1}+ X{\bm\alpha}-\bm{y}-\bm{\lambda}^t / \mu \|_2^2}\\ &= \|{\bm\alpha}\|_1 + f({\bm\alpha})
\end{split}
\end{equation}
where $f({\bm\alpha})=\frac{\mu}{2}\|\bm{s}^{t+1}+ X{\bm\alpha}-\bm{y}-\bm{\lambda}^t / \mu \|_2^2$.
If the second order Taylor expansion is used to approximate $f({\bm\alpha})$, the problem \ref{5-27} can be approximately reformulated as
\begin{equation} \label{5-28}
\begin{split}
\arg\min \{\| {\bm\alpha}\|_1 + ({\bm\alpha}-{\bm\alpha}^t)^T X^T (\bm{s}^{t+1}+ X{\bm\alpha}^t-\bm{y}-\bm{\lambda}^t / \mu )\\ + \frac{1}{2\tau}\| {\bm\alpha}-{\bm\alpha}^t \|_2^2\}
\end{split}
\end{equation}
where $\tau$ is a proximal parameter. The solution of problem \ref{5-28} can be obtained by the soft thresholding operator
\begin{equation} \label{5-29}
{\bm\alpha}^{t+1} = soft\{ {\bm\alpha}^t - \tau X^T (\bm{s}^{t+1}+ X{\bm\alpha}^t-\bm{y}-\bm{\lambda}^t / \mu ),  \frac{\tau}{\mu}\}
\end{equation}
where $soft(\sigma, \eta) = sign(\sigma) \max\{ |\sigma|-\eta,0 \}$.\\
Finally, the Lagrange multiplier vector $\bm{\lambda}$  is updated by using Eq. \ref{5-24}(c).

The algorithm presented above utilizes the second order Taylor expansion to approximately solve the sub-problem \ref{5-27} and thus the algorithm is denoted as an inexact ADM or approximate ADM. The main procedures of the inexact ADM based sparse representation method are summarized in Algorithm 4. More specifically, the inexact ADM described above is to reformulate the unconstrained problem as a constrained problem, and then utilizes the alternative strategy to effectively address the corresponding sub-optimization problem. Moreover, ADM can also efficiently solve the dual problems of the primal problems \ref{3-9}-\ref{3-12}. For more information, please refer to the literature \cite{yang2011alternating, boyd2011distributed}.

\begin{table}[htbp]
\centering
\begin{tabular}{m{85mm}}
\toprule
\textbf{Algorithm 4.} Alternating direction method (ADM) based sparse representation strategy\\
\textbf{Task:} To address the unconstraint problem:\\~~~~~~~$\hat{{\bm\alpha}}=\arg\min_{{\bm\alpha}} \frac{1}{2}\|\bm{y} - X{\bm\alpha}\|_2^2 + \tau \|{\bm\alpha}\|_1$\\
\midrule
\textbf{Input:} Probe sample $y$, the measurement matrix $X$, small constant $\lambda$ \\
\textbf{Initialization:} $t=0$, $\bm{s}^0 = \textbf{0}$, ${\bm\alpha}^0 = \textbf{0}$, $\bm{\lambda}^0 = \textbf{0}$, $\tau = 1.01$, $\mu$ is a small constant.\\
~Step 1: Construct the constraint optimization problem of problem \ref{3-12} by introducing the auxiliary parameter and its augmented Lagrangian function, i.e. problem (\ref{5-22}) and (\ref{5-23}).\\
While not converged do \\
~~~Step 2:~Update the value of the $\bm{s}^{t+1}$ by using Eq. (\ref{5-25}).\\
~~~Step 2:~Update the value of the ${\bm\alpha}^{t+1}$ by using Eq. (\ref{5-29}). \\
~~~Step 3:~Update the value of the $\bm{\lambda}^{t+1}$ by using Eq. (\ref{5-24}(c)).\\
~~~Step 4:~$\mu^{t+1} = \tau \mu^t$ and $t=t+1$.\\
End While \\
\textbf{Output:}~${\bm\alpha}^{t+1}$\\
\bottomrule
\end{tabular}
\end{table}

\section{Proximity algorithm based optimization strategy }
\noindent In this section, the methods that exploit the proximity algorithm to solve constrained convex optimization problems are discussed. The core idea of the proximity algorithm is to utilize the proximal operator to iteratively solve the sub-problem, which is much more computationally efficient than the original problem. The proximity algorithm is frequently employed to solve nonsmooth, constrained convex optimization problems \cite{Parikh2013Proximal}. Furthermore, the general problem of sparse representation with $l_1$-norm regularization is a nonsmooth convex optimization problem, which can be effectively addressed by using the proximal algorithm.

Suppose a simple constrained optimization problem is
\begin{equation} \label{6-1}
\min\{ h(\bm{x}) | \bm{x} \in \chi\}
\end{equation}
where $\chi \subset \mathbb{R}^n$. The general framework of addressing the constrained convex optimization problem \ref{6-1} using the proximal algorithm can be reformulated as
\begin{equation} \label{6-2}
\tilde{\bm{x}}^t=\arg\min\{ h(\bm{x}) + \frac{\tau}{2} \|\bm{x}-\bm{x}^t\|^2 | \bm{x} \in \chi\}
\end{equation}
where $\tau$ and $\bm{x}^t$ are given. For definiteness and without loss of generality, it is assumed that there is the following linear constrained convex optimization problem
\begin{equation} \label{6-3}
\arg\min\{ F(\bm{x}) + G(\bm{x}) | \bm{x} \in \chi\}
\end{equation}
The solution of problem \ref{6-3} obtained by employing the proximity algorithm is:
\begin{equation} \label{6-4}
\begin{split}
\bm{x}^{t+1} =& \arg\min \{ F(\bm{x}) + \langle \nabla G(\bm{x}^t), \bm{x}-\bm{x}^t \rangle + \frac {1}{2\tau} \|\bm{x}-\bm{x}^t\|^2 \}
\\=& \arg\min \{ F(\bm{x}) + \frac {1}{2\tau} \|\bm{x}-\bm{\theta}^t\|^2 \}
\end{split}
\end{equation}
where $\bm{\theta} = \bm{x}^t - \tau \nabla G(\bm{x}^t)$. More specifically, for the sparse representation problem with $l_1$-norm regularization, the main problem can be reformulated as:
\begin{equation} \label{6-5}
\begin{split}
&\min P({\bm\alpha}) = \{\lambda \|{\bm\alpha}\|_1 ~|~ A{\bm\alpha} =\bm{y} \}~~\\or~~&\min P({\bm\alpha}) = \{\lambda \|{\bm\alpha}\|_1 + \| A{\bm\alpha} - \bm{y} \|_2^2 ~|~ {\bm\alpha} \in \mathbb{R}^n \}
\end{split}
\end{equation}
which are considered as the constrained sparse representation of problem \ref{3-12}.

\subsection{Soft thresholding or shrinkage operator}
\noindent First, a simple form of problem \ref{3-12} is introduced, which has a closed-form solution, and it is formulated as:
\begin{equation} \label{6-6}
\begin{split}
{\bm\alpha}^* = \min_{\bm\alpha} h({\bm\alpha}) = & \lambda \|{\bm\alpha}\|_1 + \frac{1}{2}  \|{\bm\alpha} - s\|^2 \\ = & \sum_{j=1}^N \lambda|{\bm\alpha}_j| + \sum_{j=1}^N \frac{1}{2} ({\bm\alpha}_j-s_j)^2
\end{split}
\end{equation}
where ${\bm\alpha}^*$ is the optimal solution of problem \ref{6-6}, and then there are the following conclusions:\\
(1) if ${\bm\alpha}_j>0$, then $h({\bm\alpha})=\lambda {\bm\alpha} + \frac{1}{2} \|{\bm\alpha} - s\|^2$ and its derivative is $h' ({\bm\alpha}_j) = \lambda + {\bm\alpha}_j^* - s_j$. \\
Let $h' ({\bm\alpha}_j)=0$ $\Rightarrow$ ${\bm\alpha}_j^*=s_j - \lambda$, where it indicates $s_j > \lambda$;\\
(2) if ${\bm\alpha}_j<0$, then $h({\bm\alpha})=-\lambda {\bm\alpha} + \frac{1}{2} \|{\bm\alpha} - s\|^2$ and its derivative is $h' ({\bm\alpha}_j) = -\lambda + {\bm\alpha}_j^* - s_j$. \\
Let $h' ({\bm\alpha}_j)=0$ $\Rightarrow$ ${\bm\alpha}_j^*=s_j + \lambda$, where it indicates $s_j <- \lambda$;\\
(3) if $ - \lambda \leq s_j \leq \lambda$, and then ${\bm\alpha}_j^* =0$.\\
So the solution of problem \ref{6-6} is summarized as

\begin{eqnarray} \label{6-7}
{\bm\alpha}_j^* =
\left\{
\begin{array}{ll}
s_j - \lambda, ~~& if~~ s_j > \lambda \\
s_j + \lambda, ~~& if~~ s_j <- \lambda \\
0, & otherwise\\
\end{array}
\right.
\end{eqnarray}
The equivalent expression of the solution is ${\bm\alpha}^* = shrink (s,\lambda)$, where the $j$-th component of $shrink(s,\lambda)$ is $shrink(s,\lambda)_j = sign(s_j) \max \{ |s_j| - \lambda, 0 \}$. The operator $shrink(\bullet)$ can be regarded as a proximal operator.

\subsection{Iterative shrinkage thresholding algorithm (ISTA)}
\noindent The objective function of ISTA \cite{figueiredo2005bound} has the form of

\begin{equation} \label{6-8}
\arg \min F({\bm\alpha}) = \frac{1}{2} \|X{\bm\alpha} - \bm{y}\|_2^2 +\lambda \|{\bm\alpha}\|_1 =f({\bm\alpha}) + \lambda g({\bm\alpha})
\end{equation}
and is usually difficult to solve. Problem \ref{6-8} can be converted to the form of an easy problem \ref{6-6} and the explicit procedures are presented as follows.

First, Taylor expansion is used to approximate $f({\bm\alpha}) = \frac{1}{2} \|X{\bm\alpha} - \bm{y}\|_2^2$ at a point of ${\bm\alpha}^t$. The second order Taylor expansion is
\begin{equation} \label{6-9}
\begin{split}
f({\bm\alpha}) =  f({\bm\alpha}^t) + ({\bm\alpha}-{\bm\alpha}^t)^T  \nabla f({\bm\alpha}^t) + \frac{1}{2}({\bm\alpha}-{\bm\alpha}^t)^T \\ H_f({\bm\alpha}^t) ({\bm\alpha}-{\bm\alpha}^t) + \cdots
\end{split}
\end{equation}
where $H_f({\bm\alpha}^t)$ is the Hessian matrix of $f({\bm\alpha})$ at ${\bm\alpha}^t$. For the function $f({\bm\alpha})$, $\nabla f({\bm\alpha})=X^T(X{\bm\alpha}-\bm{y})$ and $H_f({\bm\alpha})=X^TX$ can be obtained.

\begin{equation} \label{6-10}
\begin{split}
f({\bm\alpha}) =  \frac{1}{2} \|X{\bm\alpha}^t - \bm{y}\|_2^2 + ({\bm\alpha}-{\bm\alpha}^t)^T  X^T (X{\bm\alpha}^t-\bm{y}) + \\ \frac{1}{2}({\bm\alpha}-{\bm\alpha}^t)^T X^TX ({\bm\alpha}-{\bm\alpha}^t)
\end{split}
\end{equation}
If the Hessian matrix $H_f({\bm\alpha})$ is replaced or approximated in the third term above by using a scalar $\frac{1}{\tau}I$, and then
\begin{equation} \label{6-11}
\begin{split}
f({\bm\alpha}) \approx  \frac{1}{2} \|X{\bm\alpha}^t - \bm{y}\|_2^2 + ({\bm\alpha}-{\bm\alpha}^t)^T  X^T (X{\bm\alpha}^t-\bm{y})\\ +\frac{1}{2\tau}({\bm\alpha}-{\bm\alpha}^t)^T ({\bm\alpha}-{\bm\alpha}^t)= Q_t({\bm\alpha}, {\bm\alpha}^t)
\end{split}
\end{equation}
Thus problem \ref{6-8} using the proximal algorithm can be successively addressed by
\begin{equation} \label{6-12}
{\bm\alpha}^{t+1} = \arg \min Q_t({\bm\alpha}, {\bm\alpha}^t) +\lambda \|{\bm\alpha}\|_1
\end{equation}
Problem \ref{6-12} is reformulated to a simple form of problem \ref{6-6} by
\begin{equation} \label{6-13}
\begin{split}
Q_t({\bm\alpha}, {\bm\alpha}^t) = \frac{1}{2} \|X{\bm\alpha}^t - \bm{y}\|_2^2 + &({\bm\alpha}-{\bm\alpha}^t)^T  X^T (X{\bm\alpha}^t-\bm{y})+ \\&~~~~~~~~~~~~~~~~~ \frac{1}{2\tau} \|{\bm\alpha}-{\bm\alpha}^t\|_2^2\\
= \frac{1}{2} \|X{\bm\alpha}^t - \bm{y}\|_2^2 + \frac{1}{2\tau}\| {\bm\alpha}-&{\bm\alpha}^t+\tau X^T (X{\bm\alpha}^t-\bm{y}) \|_2^2 \\&~~~~~~- \frac{\tau}{2}\| X^T (X{\bm\alpha}^t-\bm{y}) \|_2^2\\
=  \frac{1}{2\tau}\| {\bm\alpha}-({\bm\alpha}^t- \tau X^T(X {\bm\alpha}^t& -\bm{y})) \|_2^2 +B({\bm\alpha}^t)
\end{split}
\end{equation}
where the term $B({\bm\alpha}^t) =\frac{1}{2} \|X{\bm\alpha}^t - \bm{y}\|_2^2 - \frac{\tau}{2}\| X^T (X{\bm\alpha}^t-\bm{y}) \|^2$ in problem \ref{6-12} is a constant with respect to variable ${\bm\alpha}$, and it can be omitted. As a result, problem \ref{6-12} is equivalent to the following problem:
\begin{equation} \label{6-14}
{\bm\alpha}^{t+1} = \arg \min \frac{1}{2\tau}\| {\bm\alpha}- \theta({\bm\alpha}^t)\|_2^2 +\lambda \|{\bm\alpha}\|_1
\end{equation}
where $\theta({\bm\alpha}^t) = {\bm\alpha}^t-\tau X^T (X{\bm\alpha}^t-\bm{y}) $.

The solution of the simple problem \ref{6-6} is applied to solve problem \ref{6-14} where the parameter $t$ is replaced by the equation $\bm{\theta}({\bm\alpha}^t)$, and the solution of problem \ref{6-14} is ${\bm\alpha}^{t+1} = shrink (\bm{\theta}({\bm\alpha}^t), \lambda \tau)$. Thus, the solution of ISTA is reached. The techniques used here are called linearization or preconditioning and more detailed information can be found in the literature \cite{figueiredo2005bound, combettes2011proximal}.

\subsection{Fast Iterative shrinkage thresholding algorithm (FISTA)}
\noindent The fast iterative shrinkage thresholding algorithm (FISTA) is an improvement of ISTA. FISTA \cite{beck2009fast} not only preserves the efficiency of the original ISTA but also promotes the effectiveness of ISTA so that FISTA can obtain global convergence.

Considering that the Hessian matrix $H_f({\bm\alpha})$ is approximated by using a scalar $\frac{1}{\tau}I$ for ISTA in Eq. \ref{6-9}, FISTA utilizes the minimum Lipschitz constant of the gradient $\nabla f({\bm\alpha})$ to approximate the Hessian matrix of $f({\bm\alpha})$, i.e. $L(f)=2\lambda_{max}(X^TX)$. Thus, the problem \ref{6-8} can be converted to the problem below:
\begin{equation} \label{6-15}
\begin{split}
f({\bm\alpha}) \approx  \frac{1}{2} \|X{\bm\alpha}^t - \bm{y}\|_2^2 + ({\bm\alpha}-{\bm\alpha}^t)^T  X^T (X{\bm\alpha}^t-\bm{y}) \\ + \frac{L}{2}({\bm\alpha}-{\bm\alpha}^t)^T({\bm\alpha}-{\bm\alpha}^t)= P_t({\bm\alpha}, {\bm\alpha}^t)
\end{split}
\end{equation}
where the solution can be reformulated as
\begin{equation} \label{6-16}
{\bm\alpha}^{t+1} = \arg \min \frac{L}{2}\| {\bm\alpha}- \theta({\bm\alpha}^t)\|_2^2 +\lambda \|{\bm\alpha}\|_1
\end{equation}
where $\theta({\bm\alpha}^t) = {\bm\alpha}^t-\frac{1}{L} X^T (X{\bm\alpha}^t-\bm{y}) $.

Moreover, to accelerate the convergence of the algorithm, FISTA also improves the sequence of iteration points, instead of employing the previous point it utilizes a specific linear combinations of the previous two points $\{{\bm\alpha}^t, {\bm\alpha}^{t-1}\}$, i.e.
\begin{equation} \label{6-17}
{\bm\alpha}^t = {\bm\alpha}^t + \frac{\mu ^t -1}{\mu ^{t+1}} ({\bm\alpha}^t-{\bm\alpha}^{t-1})
\end{equation}
where $\mu^t$ is a positive sequence, which satisfies $\mu^t \geq (t+1)/2$, and the main steps of FISTA are summarized in Algorithm 5. The backtracking linear research strategy can also be utilized to explore a more feasible value of $L$ and more detailed analyses on FISTA can be found in the literature \cite{beck2009fast, yang2010fast}.

\begin{table}[htbp]
\centering
\begin{tabular}{m{85mm}}
\toprule
\textbf{Algorithm 5.} Fast Iterative shrinkage thresholding algorithm (FISTA)\\
\textbf{Task:} To address the problem $\hat{{\bm\alpha}}=\arg\min F({\bm\alpha}) = \frac{1}{2} \|X{\bm\alpha} - \bm{y}\|_2^2 + \lambda \|{\bm\alpha}\|_1$\\
\midrule
\textbf{Input:} Probe sample $\bm{y}$, the measurement matrix $X$, small constant $\lambda$ \\
\textbf{Initialization:} $t=0$, $\mu^0 = 1$, $L= 2 \Lambda_{max} (X^TX)$, i.e. Lipschitz constant of $\nabla f$.\\
While not converged do\\
~~~Step 1:~Exploit the shrinkage operator in equation \ref{6-7} to solve problem\\~~~~~~~ \ref{6-16}.\\
~~~Step 2:~Update the value of $\mu$ using $\mu^{t+1} = \frac{1+\sqrt{1+4(\mu^t)^2}}{2}$.\\
~~~Step 3:~Update iteration sequence ${\bm\alpha}^t$ using equation \ref{6-17}.\\
End\\
\textbf{Output:}~${\bm\alpha}$\\
\bottomrule
\end{tabular}
\end{table}

\subsection{Sparse reconstruction by separable approximation (SpaRSA)}
\noindent
Sparse reconstruction by separable approximation (SpaRSA) \cite{sjwright2009sparse} is another typical proximity algorithm based on sparse representation, which can be viewed as an accelerated version of ISTA. SpaRSA provides a general algorithmic framework for solving the sparse representation problem and here a simple specific SpaRSA with adaptive continuation on ISTA is introduced. The main contributions of SpaRSA are trying to optimize the parameter $\lambda$ in problem \ref{6-8} by using the worm-starting technique, i.e. continuation, and choosing a more reliable approximation of $H_f({\bm\alpha})$ in problem \ref{6-9} using the Barzilai-Borwein (BB) spectral method \cite{dai2006cyclic}. The worm-starting technique and BB spectral approach are introduced as follows.

\noindent(1) Utilizing the worm-starting technique to optimize $\lambda$

The values of $\lambda$ in the sparse representation methods discussed above are always set to be a specific small constant. However, Hale et al. \cite{hale2007fixed} concluded that the technique that exploits a decreasing value of $\lambda$ from a warm-starting point can more efficiently solve the sub-problem \ref{6-14} than ISTA that is a fixed point iteration scheme. SpaRSA uses an adaptive continuation technique to update the value of $\lambda$ so that it can lead to the fastest convergence. The procedure regenerates the value of $\lambda$ using
\begin{equation} \label{6-18}
\lambda = \max \{ \gamma \|X^T \bm{y}\|_\infty, \lambda \}
\end{equation}
where $\gamma$ is a small constant.

\noindent(2) Utilizing the BB spectral method to approximate $H_f({\bm\alpha})$

ISTA employs $\frac{1}{\tau}I$ to approximate the matrix $H_f({\bm\alpha})$, which is the Hessian matrix of $f({\bm\alpha})$ in problem \ref{6-9} and FISTA exploits the Lipschitz constant of $\nabla f({\bm\alpha})$ to replace $H_f({\bm\alpha})$. However, SpaRSA utilizes the BB spectral method to choose the value of $\tau$ to mimic the Hessian matrix. The value of $\tau$ is required to satisfy the condition:
\begin{equation} \label{6-19}
\frac{1}{\tau^{t+1}}({\bm\alpha}^{t+1} - {\bm\alpha}^t) \approx \nabla f({\bm\alpha}^{t+1}) - \nabla f({\bm\alpha}^{t})
\end{equation}
which satisfies the minimization problem
\begin{equation} \label{6-20}
\begin{split}
\frac{1}{\tau^{t+1}} = &\arg \min \| \frac{1}{\tau}({\bm\alpha}^{t+1} - {\bm\alpha}^t) -  (\nabla f({\bm\alpha}^{t+1}) - \nabla f({\bm\alpha}^{t})) \|_2^2\\
&= \frac{({\bm\alpha}^{t+1} - {\bm\alpha}^t)^T(\nabla f({\bm\alpha}^{t+1}) - \nabla f({\bm\alpha}^{t}))}{({\bm\alpha}^{t+1} - {\bm\alpha}^t)^T({\bm\alpha}^{t+1} - {\bm\alpha}^t)}
\end{split}
\end{equation}

For problem \ref{6-14}, SpaRSA requires that the value of $\lambda$ is a decreasing sequence using the Eq. \ref{6-18} and the value of $\tau$ should meet the condition of Eq. \ref{6-20}. The sparse reconstruction by separable approximation (SpaRSA) is summarized in Algorithm 6 and more information can be found in the literature \cite{sjwright2009sparse}.

\begin{table}[htbp]
\centering
\begin{tabular}{m{85mm}}
\toprule
\textbf{Algorithm 6.} Sparse reconstruction by separable approximation (SpaRSA)\\
\textbf{Task:} To address the problem\\~~~~~~~~~~~~~~~$\hat{{\bm\alpha}}=\arg\min F({\bm\alpha}) = \frac{1}{2} \|X{\bm\alpha} - \bm{y}\|_2^2 + \lambda \|{\bm\alpha}\|_1$\\
\midrule
\textbf{Input:} Probe sample $y$, the measurement matrix $X$, small constant $\lambda$ \\
\textbf{Initialization:} $t=0$, $i=0$, $\bm{y}^0=\bm{y}$, $\frac{1}{\tau^0}I \approx H_f({\bm\alpha}) = X^TX$, tolerance $\varepsilon = 10^{-5}$.\\
~~~Step 1:~$\lambda_t = \max \{ \gamma \|X^T \bm{y}^t\|_{\infty}, \lambda\}.$\\
~~~Step 2:~Exploit shrinkage operator to solve problem \ref{6-14}, i.e.\\~~~~~~~~ ${\bm\alpha}^{i+1} = shrink ({\bm\alpha}^i - \tau^i X^T(X^T{\bm\alpha}^t - \bm{y}), \lambda_t \tau^i)$. \\
~~~Step 3:~Update the value of $\frac{1}{\tau^{i+1}}$ using the Eq. \ref{6-20}.\\
~~~Step 4:~If $\frac{\|{\bm\alpha}^{i+1}-{\bm\alpha}^i\|}{{\bm\alpha}^i} \leq \varepsilon$, go to step 5; Otherwise, return to step 2 \\~~~~~~~~ and $i=i+1$.\\
~~~Step 5:~$\bm{y}^{t+1} = \bm{y} - X{\bm\alpha}^{t+1}$.\\
~~~Step 6:~If $\lambda_t=\lambda$, stop; Otherwise, return to step 1 and $t=t+1$.\\
\textbf{Output:}~${\bm\alpha}^i$\\
\bottomrule
\end{tabular}
\end{table}

\subsection{$l_{1/2}$-norm regularization based sparse representation}
\noindent
Sparse representation with the $l_p$-norm (0$<$p$<$1) regularization leads to a nonconvex, nonsmooth, and non-Lipschitz optimization problem and its general forms are described as problems \ref{3-13} and \ref{3-14}. The $l_p$-norm (0$<$p$<$1) regularization problem is always difficult to be efficiently addressed and it has also attracted wide interests from large numbers of research groups. However, the research group led by Zongben Xu summarizes the conclusion that the most impressive and representative algorithm of the $l_p$-norm (0$<$p$<$1) regularization is sparse representation with the $l_{1/2}$-norm regularization \cite{zeng2013accelerated}. Moreover, they have proposed some effective methods to solve the $l_{1/2}$-norm regularization problem \cite{xu2012regularization, Zeng2014ell}.

In this section, a half proximal algorithm is introduced to solve the $l_{1/2}$-norm regularization problem \cite{xu2012regularization}, which matches the iterative shrinkage thresholding algorithm for the $l_1$-norm regularization discussed above and the iterative hard thresholding algorithm for the $l_0$-norm regularization. Sparse representation with the $l_{1/2}$-norm regularization is explicitly to solve the problem as follows:
\begin{equation} \label{6-21}
\hat {\bm\alpha} = \arg\min \{ F({\bm\alpha}) = \| X{\bm\alpha} - \bm{y} \|_2^2 + \lambda \|{\bm\alpha}\|_{1/2}^{1/2} \}
\end{equation}
where the first-order optimality condition of $F({\bm\alpha})$ on ${\bm\alpha}$ can be formulated as
\begin{equation} \label{6-22}
\nabla F({\bm\alpha}) = X^T (X{\bm\alpha} - \bm{y}) + \frac{\lambda}{2} \nabla (\|{\bm\alpha}\|_{1/2}^{1/2}) = 0
\end{equation}
which admits the following equation:
\begin{equation} \label{6-23}
X^T (\bm{y} - X{\bm\alpha}) = \frac{\lambda}{2} \nabla (\|{\bm\alpha}\|_{1/2}^{1/2})
\end{equation}
where $\nabla (\|{\bm\alpha}\|_{1/2}^{1/2})$ denotes the gradient of the regularization term $\|{\bm\alpha}\|_{1/2}^{1/2}$. Subsequently, an equivalent transformation of Eq. \ref{6-23} is made by multiplying a positive constant $\tau$ and adding a parameter ${\bm\alpha}$ to both sides. That is,

\begin{equation} \label{6-24}
{\bm\alpha} + \tau X^T (\bm{y} - X{\bm\alpha}) = {\bm\alpha} + \tau \frac{\lambda}{2} \nabla (\|{\bm\alpha}\|_{1/2}^{1/2})
\end{equation}

To this end, the resolvent operator \cite{xu2012regularization} is introduced to compute the resolvent solution of the right part of Eq. \ref{6-24}, and the resolvent operator is defined as

\begin{equation} \label{6-25}
R_{\lambda, \frac{1}{2}} (\bullet) = \left( I + \frac{\lambda \tau}{2} \nabla (\| \bullet \|_{1/2}^{1/2}) \right) ^{-1}
\end{equation}
which is very similar to the inverse function of the right part of Eq. \ref{6-24}. The resolvent operator is always satisfied no matter whether the resolvent solution of $\nabla (\| \bullet \|_{1/2}^{1/2})$ exists or not \cite{xu2012regularization}. Applying the resolvent operator to solve problem \ref{6-24}
\begin{equation} \label{6-26}
\begin{split}
{\bm\alpha} &= (I + \frac{\lambda \tau}{2} \nabla (\| \bullet \|_{1/2}^{1/2}))^{-1} ({\bm\alpha} + \tau X^t (\bm{y}-X{\bm\alpha})) \\&= R_{\lambda, 1/2} ({\bm\alpha} + \tau X^T(\bm{y}-X{\bm\alpha}))
\end{split}
\end{equation}
can be obtained which is well-defined. $\theta ({\bm\alpha}) = {\bm\alpha} + \tau X^T (\bm{y}- X{\bm\alpha})$ is denoted and the resolvent operator can be explicitly expressed as:
\begin{equation} \label{6-27}
R_{\lambda, \frac{1}{2}} (\bm{x}) = (f_{\lambda, \frac{1}{2}} (\bm{x}_1), f_{\lambda, \frac{1}{2}} (\bm{x}_2), \cdots, f_{\lambda, \frac{1}{2}} (\bm{x}_N))^T
\end{equation}
where
\begin{equation} \label{6-28}
\begin{split}
f_{\lambda, \frac{1}{2}} (\bm{x}_i)&= \frac{2}{3} \bm{x}_i(1+\cos({\frac{2\pi}{3} -\frac{2}{3} g_\lambda(\bm{x}_i)}),\\g_{\lambda}(\bm{x}_i) &= \arg \cos (\frac{\lambda}{8}(\frac{|\bm{x}_i|}{3})^{-\frac{3}{2}})
\end{split}
\end{equation}
which have been demonstrated in the literature \cite{xu2012regularization}.

Thus the half proximal thresholding function for the $l_{1/2}$-norm regularization is defined as below:
\begin{eqnarray} \label{6-29}
h_{\lambda\tau, \frac{1}{2}} (\bm{x}_i) =
\left\{
\begin{array}{ll}
f_{\lambda\tau, \frac{1}{2}}(\bm{x}_i), ~~&if~~ |\bm{x}_i|> \frac{\sqrt[3]{54}}{4}(\lambda \tau)^{\frac{2}{3}} \\
0, & otherwise\\
\end{array}
\right.
\end{eqnarray}
where the threshold $\frac{\sqrt[3]{54}}{4}(\lambda \tau)^{\frac{2}{3}}$ has been conceived and demonstrated in the literature \cite{xu2012regularization}.

Therefore, if Eq. \ref{6-29} is applied to Eq. \ref{6-27}, the half proximal thresholding function, instead of the resolvent operator, for the $l_{1/2}$-norm regularization problem \ref{6-25} can be explicitly reformulated as:
\begin{equation} \label{6-30}
{\bm\alpha} = H_{\lambda \tau, \frac{1}{2}}(\theta ({\bm\alpha}))
\end{equation}
where the half proximal thresholding operator $H$ \cite{xu2012regularization} is deductively constituted by Eq. \ref{6-29}.

Up to now, the half proximal thresholding algorithm has been completely structured by Eq. \ref{6-30}. However, the options of the regularization parameter $\lambda$ in Eq. \ref{6-24} can seriously dominate the quality of the representation solution in problem \ref{6-21}, and the values of $\lambda$ and $\tau$ can be specifically fixed by
\begin{equation} \label{6-31}
\tau = \frac{1-\varepsilon}{\|X\|^2} ~~and~~\lambda = \frac{\sqrt{96}}{9\tau} |[\theta ({\bm\alpha})] _{k+1}|^{\frac{3}{2}}
\end{equation}
where $\varepsilon$ is a very small constant, which is very close to zero, the $k$ denotes the limit of sparsity (i.e. $k$-sparsity), and $[\bullet]_k$ refers to the $k$-th largest component of $[\bullet]$. The half proximal thresholding algorithm for $l_{1/2}$-norm regularization based sparse representation is summarized in Algorithm 7 and more detailed inferences and analyses can be found in the literature \cite{xu2012regularization, Zeng2014ell}.

\begin{table}[htbp]
\centering
\begin{tabular}{m{85mm}}
\toprule
\textbf{Algorithm 7.} The half proximal thresholding algorithm for $l_{1/2}$-norm regularization\\
\textbf{Task:} To address the problem \\
~~~~~~~~$\hat{{\bm\alpha}}=\arg\min F({\bm\alpha}) = \|X{\bm\alpha} - \bm{y}\|_2^2 + \lambda \|{\bm\alpha}\|_{1/2}^{1/2}$\\
\midrule
\textbf{Input:} Probe sample $y$, the measurement matrix $X$ \\
\textbf{Initialization:} $t=0$, $\varepsilon = 0.01$, $\tau = \frac{1-\varepsilon}{\|X\|^2}$.\\
While not converged do\\
~~~Step 1:~Compute $\theta({\bm\alpha}^t) = {\bm\alpha}^t + \tau X^T (\bm{y}-X{\bm\alpha}^t)$.\\
~~~Step 2:~Compute $\lambda_t = \frac{\sqrt{96}}{9\tau} |[\theta ({\bm\alpha}^t)] _{k+1}|^{\frac{3}{2}}$ in Eq. \ref{6-31}. \\
~~~Step 3:~Apply the half proximal thresholding operator to obtain \\~~~~~~~~~~the representation solution ${\bm\alpha}_{t+1} = H_{\lambda_t \tau, \frac{1}{2}}(\theta ({\bm\alpha}^t))$.\\
~~~Step 4:~$t=t+1$.\\
End\\
\textbf{Output:}~${\bm\alpha}$\\
\bottomrule
\end{tabular}
\end{table}

\subsection{Augmented Lagrange Multiplier based optimization strategy}
\noindent
The Lagrange multiplier is a widely used tool to eliminate the equality constrained problem and convert it to address the unconstrained problem with an appropriate penalty function. Specifically, the sparse representation problem \ref{3-9} can be viewed as an equality constrained problem and the equivalent problem \ref{3-12} is an unconstrained problem, which augments the objective function of problem \ref{3-9} with a weighted constraint function. In this section, the augmented Lagrangian method (ALM) is introduced to solve the sparse representation problem \ref{3-9}.

First, the augmented Lagrangian function of problem \ref{3-9} is conceived by introducing an additional equality constrained function, which is enforced on the Lagrange function in problem \ref{3-12}. That is,

\begin{equation} \label{6-32}
L({\bm\alpha}, \lambda)= \|{\bm\alpha}\|_1 + \frac{\lambda}{2}\| \bm{y}-X{\bm\alpha}\|_2^2~~~~ s.t. ~~~~  \bm{y}-X{\bm\alpha}=0
\end{equation}
Then, a new optimization problem \ref{6-32} with the form of the Lagrangain function is reformulated as
\begin{equation} \label{6-33}
\arg \min L_{\lambda}({\bm\alpha}, \bm{z}) = \|{\bm\alpha}\|_1 + \frac{\lambda}{2}\| \bm{y}-X{\bm\alpha}\|_2^2 +{\bm z}^T (\bm{y}-X{\bm\alpha})
\end{equation}
where $z \in \mathbb{R}^d$ is called the Lagrange multiplier vector or dual variable and $L_{\lambda}({\bm\alpha},\bm{z})$ is denoted as the augmented Lagrangian function of problem \ref{3-9}. The optimization problem \ref{6-33} is a joint optimization problem of the sparse representation coefficient ${\bm\alpha}$ and the Lagrange multiplier vector $\bm{z}$.  Problem \ref{6-33} is solved by optimizing ${\bm\alpha}$ and $\bm{z}$ alternatively as follows:
\begin{equation} \label{6-34}
\begin{split}
{\bm\alpha}^{t+1}=\arg \min & L_{\lambda}({\bm\alpha},{\bm z}^t) \\=\arg \min (& \|{\bm\alpha}\|_1 + \frac{\lambda}{2}\| \bm{y}- X {\bm\alpha} \|_2^2 +({\bm z}^t)^T X {\bm\alpha})
\end{split}
\end{equation}

\begin{equation} \label{6-35}
{\bm z}^{t+1} = {\bm z}^t + \lambda (\bm{y}-X {\bm\alpha}^{t+1})
\end{equation}
where problem \ref{6-34} can be solved by exploiting the FISTA algorithm. Problem \ref{6-34} is iteratively solved and the parameter $z$ is updated using Eq. \ref{6-35} until the termination condition is satisfied. Furthermore, if the method of employing ALM to solve problem \ref{6-33} is denoted as the primal augmented Lagrangian method (PALM) \cite{yang2013fast}, the dual function of problem \ref{3-9} can also be addressed by the ALM algorithm, which is denoted as the dual augmented Lagrangian method (DALM) \cite{yang2013fast}. Subsequently, the dual optimization problem \ref{3-9} is discussed and the ALM algorithm is utilized to solve it.

First, consider the following equation:
\begin{equation} \label{6-36}
\|{\bm\alpha}\|_1 = \max_{\|\bm{\theta}\|_\infty \leq 1} \langle \bm{\theta}, {\bm\alpha} \rangle
\end{equation}
which can be rewritten as
\begin{equation} \label{6-37}
\begin{split}
\|{\bm\alpha}\|_1 &= \max \{\langle \bm{\theta}, {\bm\alpha} \rangle - I_{B_{\infty}^1}\} \\ or ~~~~\|{\bm\alpha}\|_1 &= \sup \{\langle \bm{\theta}, {\bm\alpha} \rangle - I_{B_{\infty}^1}\}
\end{split}
\end{equation}
where $B_p^{\lambda} = \{\bm{x} \in R^N ~|~ \|\bm{x}\|_p \leq \lambda\}$ and $I_{\Omega} (\bm{x})$ is a indicator function, which is defined as $I_{\Omega} (\bm{x}) = \left\{ \begin{array}{ll} 0&,\bm{x} \in \Omega\\ \infty&,\bm{x} \not \in \Omega\\ \end{array} \right.$.

Hence,
\begin{equation} \label{6-38}
\|{\bm\alpha}\|_1 = \max \{ \langle \bm{\theta}, {\bm\alpha} \rangle\ : \bm{\theta} \in B_{\infty}^1 \}
\end{equation}

Second, consider the Lagrange dual problem of problem \ref{3-9} and its dual function is
\begin{equation} \label{6-39}
g(\lambda) = \inf_{{\bm\alpha}} \{ \|{\bm\alpha}\|_1 + \bm{\lambda}^T(\bm{y}-X{\bm\alpha}) \} =\bm{\lambda}^T\bm{y} - \sup_{{\bm\alpha}} \{ \bm{\lambda}^TX{\bm\alpha} - \|{\bm\alpha}\|_1 \}
\end{equation}
where $\bm{\lambda} \in \mathbb{R}^d$ is a Lagrangian multiplier. If the definition of conjugate function is applied to Eq. \ref{6-37}, it can be verified that the conjugate function of $I_{B_{\infty}^1}(\theta)$ is $\|{\bm\alpha}\|_1$. Thus Eq. \ref{6-39} can be equivalently reformulated as
\begin{equation} \label{6-40}
g(\bm{\lambda}) = \bm{\lambda}^T\bm{y} - I_{B_{\infty}^1} (X^T \bm{\lambda})
\end{equation}
The Lagrange dual problem, which is associated with the primal problem \ref{3-9}, is an optimization problem:

\begin{equation} \label{6-41}
\max_\lambda \bm{\lambda}^T \bm{y}~~~~s.t.~~~~(X^T \bm{\lambda}) \in B_{\infty}^1
\end{equation}
Accordingly,
\begin{equation} \label{6-42}
\min_{\lambda,z} - \bm{\lambda}^T \bm{y}~~~~s.t.~~~~\bm{z}-X^T \bm{\lambda}=0,~ \bm{z} \in B_{\infty}^1
\end{equation}
Then, the optimization problem \ref{6-42} can be reconstructed as
\begin{equation} \label{6-43}
\begin{split}
\arg \min_{\bm{\lambda},\bm{z}, \bm{\mu}} L(\bm{\lambda},\bm{z},\bm{\mu}) = -\bm{\lambda}^T \bm{y} - \bm{\mu}^T(\bm{z}-X^T \bm{\lambda}) \\ + \frac{\tau}{2} \|\bm{z}-X^T \bm{\lambda}\|_2^2~~~~s.t.~~~~\bm{z} \in B_{\infty}^1
\end{split}
\end{equation}
where $\bm{\mu} \in \mathbb{R}^d$ is a Lagrangian multiplier and $\tau$ is a penalty parameter.

Finally, the dual optimization problem \ref{6-43} is solved and a similar alternating minimization idea of PALM can also be applied to problem \ref{6-43}, that is,

\begin{equation} \label{6-44}
\begin{split}
{\bm z}^{t+1} &=\arg \min_{z \in B_\infty^1} L_{\tau}(\bm{\lambda}^t,\bm{z},\bm{\mu}^t)
\\&= \arg \min_{\bm{z} \in B_\infty^1} \{- \bm{\mu}^T(\bm{z}-X^T \bm{\lambda}^t) + \frac{\tau}{2} \|\bm{z}-X^T\bm{\lambda}^t \|_2^2 \}\\
&= \arg \min_{\bm{z} \in B_\infty^1} \{ \frac{\tau}{2} \|\bm{z}-(X^T\bm{\lambda}^t + \frac{2}{\tau} \bm{\mu}^T)\|_2^2 \}
\\&= P_{B_\infty^1} (X^T\bm{\lambda}^t + \frac{1}{\tau} \bm{\mu}^T)\\
\end{split}
\end{equation}
where $P_{B_\infty^1}(\bm{u})$ is a projection, or called a proximal operator, onto $B_\infty ^1$ and it is also called group-wise soft-thresholding. For example, let $x=P_{B_\infty^1}(\bm{u})$, then the $i$-th component of solution $\bm{x}$ satisfies $\bm{x}_i = sign (\bm{u}_i) \min\{|\bm{u}_i|,1\}$
\begin{equation} \label{6-45}
\begin{split}
\bm{\lambda}^{t+1} &= \arg \min_{\lambda} L_{\tau}(\bm{\lambda},{\bm z}^{t+1},\bm{\mu}^t)
\\&= \arg \min_{\lambda} \{ -\bm{\lambda}^T \bm{y} + (\bm{\mu}^t)^T X^T \bm{\lambda} + \frac{\tau}{2} \|{\bm z}^{t+1}-X^T \bm{\lambda}\|_2^2 \}
\\&= Q(\bm{\lambda})
\end{split}
\end{equation}

Take the derivative of $Q(\bm{\lambda})$ with respect to $\bm{\lambda}$ and obtain
\begin{equation} \label{6-46}
\bm{\lambda}^{t+1} = (\tau X X^T)^{-1} (\tau X{\bm z}^{t+1} + \bm{y} -X\bm{\mu}^t)
\end{equation}
\begin{equation} \label{6-47}
\bm{\mu}^{t+1} = \bm{\mu}^t - \tau ({\bm z}^{t+1} -X^T \bm{\lambda}^{t+1})
\end{equation}

The DALM for sparse representation with $l_1$-norm regularization mainly exploits the augmented Lagrange method to address the dual optimization problem of problem \ref{3-9} and a proximal operator, the projection operator, is utilized to efficiently solve the subproblem. The algorithm of DALM is summarized in Algorithm 8. For more detailed description, please refer to the literature \cite{yang2013fast}.
\begin{table}[htbp]
\centering
\begin{tabular}{m{85mm}}
\toprule
\textbf{Algorithm 8.} Dual augmented Lagrangian method for  $l_1$-norm regularization\\
\textbf{Task:} To address the dual problem of~~$\hat{{\bm\alpha}}=\arg\min_{{\bm\alpha}} \|{\bm\alpha}\|_1~~s.t.~~ \bm{y} = X{\bm\alpha}$\\
\midrule
\textbf{Input:} Probe sample $y$, the measurement matrix $X$, a small constant $\lambda^0$. \\
\textbf{Initialization:} $t=0$, $\varepsilon = 0.01$, $\tau = \frac{1-\varepsilon}{\|X\|^2}$, $\mu^0=0$.\\
While not converged do \\
~~~Step 1:~Apply the projection operator to compute \\~~~~~~~~~~${\bm z}^{t+1}=P_{B_\infty^1} (X^T\bm{\lambda}^t + \frac{1}{\tau} \bm{\mu}^T)$.\\
~~~Step 2:~Update the value of $\bm{\lambda}^{t+1} = (\tau X X^T)^{-1} (\tau X{\bm z}^{t+1} + \bm{y} -X\bm{\mu}^t)$. \\
~~~Step 3:~Update the value of $\bm{\mu}^{t+1} = \bm{\mu}^t - \tau ({\bm z}^{t+1} -X^T \bm{\lambda}^{t+1})$.\\
~~~Step 4:~$t=t+1$.\\
End While \\
\textbf{Output:}~${\bm\alpha}=\bm{\mu}[1:N]$\\
\bottomrule
\end{tabular}
\end{table}

\subsection{Other proximity algorithm based optimization methods}
\noindent
The theoretical basis of the proximity algorithm is to first construct a proximal operator, and then utilize the proximal operator to solve the convex optimization problem. Massive proximity algorithms have followed up with improved techniques to improve the effectiveness and efficiency of proximity algorithm based optimization methods. For example, Elad et al. proposed an iterative method named parallel coordinate descent algorithm (PCDA) \cite{elad2007coordinate} by introducing the element-wise optimization algorithm to solve the regularized linear least squares with non-quadratic regularization problem.

Inspired by belief propagation in graphical models, Donoho et al. developed a modified version of the iterative thresholding method, called approximate message passing (AMP) method \cite{donoho2009message}, to satisfy the requirement that the sparsity undersampling tradeoff of the new algorithm is equivalent to the corresponding convex optimization approach. Based on the development of the first-order method called Nesterov's smoothing framework in convex optimization, Becker et al. proposed a generalized Nesterov's algorithm (NESTA) \cite{becker2011nesta} by employing the continuation-like scheme to accelerate the efficiency and flexibility.
Subsequently, Becker et al. \cite{becker2011templates} further constructed a general framework, i.e. templates for convex cone solvers (TFOCS), for solving massive certain types of compressed sensing reconstruction problems by employing the optimal first-order method to solve the smoothed dual problem of the equivalent conic formulation of the original optimization problem. Further detailed analyses and inference information related to proximity algorithms can be found in the literature \cite{Parikh2013Proximal, yang2010fast}.

\section{Homotopy algorithm based sparse representation}
\noindent
The concept of homotopy derives from topology and the homotopy technique is mainly applied to address a nonlinear system of equations problem. The homotopy method was originally proposed to solve the least square problem with the $l_1$-penalty \cite{osborne2000new}. The main idea of homotopy is to solve the original optimization problems by tracing a continuous parameterized path of solutions along with varying parameters. Having a highly intimate relationship with the conventional sparse representation method such as least angle regression (LAR) \cite{efron2004least}, OMP \cite{pati1993orthogonal} and polytope faces pursuit (PFP) \cite{plumbley2006recovery}, the homotopy algorithm has been successfully employed to solve the $l_1$-norm minimization problems. In contrast to LAR and OMP, the homotopy method is more favorable for sequentially updating the sparse solution by adding or removing elements from the active set. Some representative methods that exploit the homotopy-based strategy to solve the sparse representation problem with the $l_1$-norm regularization are explicitly presented in the following parts of this section.

\subsection{LASSO homotopy}
\noindent
Because of the significance of parameters in $l_1$-norm minimization, the well-known LASSO homotopy algorithm is proposed to solve the LASSO problem in \ref{3-9} by tracing the whole homotopy solution path in a range of decreasing values of parameter $\lambda$. It is demonstrated that problem \ref{3-12} with an appropriate parameter value is equivalent to problem \ref{3-9} \cite{Donoho2008Fast}. Moreover, it is apparent that as we change $\lambda$ from a very large value to zero, the solution of problem \ref{3-12} is converging to the solution of problem \ref{3-9} \cite{Donoho2008Fast}. The set of varying value $\lambda$ conceives the solution path and any point on the solution path is the optimality condition of problem \ref{3-12}. More specifically, the LASSO homotopy algorithm starts at an large initial value of parameter $\lambda$ and terminates at a point of $\lambda$, which approximates zero, along the homotopy solution path so that the optimal solution converges to the solution of problem \ref{3-9}. The fundamental of the homotopy algorithm is that the homotopy solution path is a piecewise linear path with a discrete number of operations while the value of the homotopy parameter changes, and the direction of each segment and the step size are absolutely determined by the sign sequence and the support of the solution on the corresponding segment, respectively \cite{asif2012fast}.

Based on the basic ideas in a convex optimization problem, it is a necessary condition that the zero vector should be a solution of the subgradient of the objective function of problem \ref{3-12}. Thus, we can obtain the subgradiential of the objective function with respect to ${\bm\alpha}$ for any given value of $\lambda$, that is,

\begin{equation} \label{7-1}
\frac{\partial L}{\partial {\bm\alpha}} = -X^T(y-X{\bm\alpha})+\lambda \partial \|{\bm\alpha}\|_1
\end{equation}
where the first term $r=X^T(y-X{\bm\alpha})$ is called the vector of residual correlations, and $\partial \|{\bm\alpha}\|_1$ is the subgradient obtained by

$\partial \|{\bm\alpha}\|_1 = \left \{ \bm{\theta} \in R^N \bigg| \begin{array}{ll} \bm{\theta}_i = sgn({\bm\alpha}_i), & {\bm\alpha}_i \neq 0 \\ \bm{\theta}_i \in [-1,1], & {\bm\alpha}_i=0 \\ \end{array} \right\}$

Let $\Lambda$ and $\bm{u}$ denote the support of ${\bm\alpha}$ and the sign sequence of ${\bm\alpha}$ on its support $\Lambda$, respectively. $X_{\Lambda}$ denotes that the indices of all the samples in  $X_{\Lambda}$ are all included in the support set $\Lambda$. If we analyze the KKT optimality condition for problem \ref{3-12}, we can obtain the following two equivalent conditions of problem \ref{7-1}, i.e.
\begin{equation} \label{7-2}
X_{\Lambda}(\bm{y}-X {\bm\alpha}) = \lambda \bm{u} ;~~~ \|X_{\Lambda^c}^T (\bm{y}-X{\bm\alpha})\|_{\infty} \leq \lambda
\end{equation}
where $\Lambda^c$ denotes the complementary set of the set $\Lambda$.  Thus, the optimality conditions in \ref{7-2} can be divided into $N$ constraints and the homotopy algorithm maintains both of the conditions along the optimal homotopy solution path for any $\lambda \geq 0$. As we decrease the value of $\lambda$ to $\lambda - \tau$, for a small value of $\tau$, the following conditions should be satisfied
\begin{equation} \label{7-3}
\begin{array}{ll}
X_{\Lambda}^T(\bm{y}-X{\bm\alpha})+\tau X_{\Lambda}^TX\bm{\delta} = (\lambda-\tau)\bm{u} & (a) \\
\|p+\tau q\|_{\infty} \leq \lambda - \tau& (b) \\
\end{array}
\end{equation}

where $\bm{p}=X^T(\bm{y}-X{\bm\alpha})$, $\bm{q}=X^TX\bm{\delta}$ and $\bm{\delta}$ is the update direction.

Generally, the homotopy algorithm is implemented iteratively and it follows the homotopy solution path by updating the support set by decreasing parameter $\lambda$ from a large value to the desired value. The support set of the solution will be updated and changed only at a critical point of $\lambda$, where either an existing nonzero element shrinks to zero or a new nonzero element will be added into the support set. The two most important parameters are the step size $\tau$ and the update direction $\bm{\delta}$. At the $l$-th stage (if $(X_{\Lambda}^TX_{\Lambda})^{-1}$ exists), the homotopy algorithm first calculates the update direction, which can be obtained by solving
\begin{equation} \label{7-4}
X_{\Lambda}^TX_{\Lambda}\bm{\delta}_l = \bm{u}
\end{equation}

Thus, the solution of problem \ref{7-4} can be written as

\begin{equation} \label{7-5}
\bm{\delta}_l = \left \{
\begin{array}{lc}
(X_{\Lambda}^TX_{\Lambda})^{-1}\bm{u}, & on~\Lambda \\
0, & otherwise\\
\end{array} \right.
\end{equation}

Subsequently, the homotopy algorithm computes the step size $\tau$ to the next critical point by tracing the homotopy solution path. i.e. the homotopy algorithm moves along the update direction until one of constraints in \ref{7-3} is not satisfied. At this critical point, a new nonzero element must enter the support $\Lambda$, or one of the nonzero elements in ${\bm\alpha}$ will be shrink to zero, i.e. this element must be removed from the support set $\Lambda$. Two typical cases may lead to a new critical point, where either condition of \ref{7-3} is violated. The minimum step size which leads to a critical point can be easily obtained by computing $\tau_l^*=min({\tau_l^+,\tau_l^-})$, and $\tau_l^+$ and $\tau_l^-$ are computed by
\begin{equation} \label{7-6}
\tau_l^+ = min_{i\in \Lambda^c} \left( \frac{\lambda-\bm{p}_i}{1-\bm{x}_i^TX_{\Lambda}\bm{\delta}_l}, \frac{\lambda+\bm{p}_i}{1+\bm{x}_i^TX_{\Lambda}\bm{\delta}_l} \right)_+
\end{equation}
\begin{equation} \label{7-7}
\tau_l^- = min_{i\in \Lambda} \left( \frac{-{\bm\alpha}_l^i}{\bm{\delta}_l^i} \right)_+
\end{equation}
where $p_i=\bm{x}_i^T(\bm{y}-\bm{x}_i{\bm\alpha}_l^i)$ and $min(\cdot)_+$ denotes that the minimum is operated over only positive arguments. $\tau_l^+$ is the minimum step size that turns an inactive element at the index $i^+$ in to an active element, i.e. the index $i^+$ should be added into the support set. $\tau_l^-$ is the minimum step size that shrinks the value of a nonzero active element to zero at the index $i^-$ and the index $i^-$ should be removed from the support set. The solution is updated by ${\bm\alpha}_{l+1}={\bm\alpha}_l+\tau_l^*\bm{\delta}$, and its support and sign sequence are renewed correspondingly.

The homotopy algorithm iteratively computes the step size and the update direction, and updates the homotopy solution and its corresponding support and sign sequence till the condition $\|\bm{p}\|_{\infty}=0$ is satisfied so that the solution of problem \ref{3-9} is reached. The principal steps of the LASSO homotopy algorithm have been summarized in Algorithm 9. For further description and analyses, please refer to the literature \cite{asif2012fast, Donoho2008Fast}.

\begin{table}[htbp]
\centering
\begin{tabular}{m{85mm}}
\toprule
\textbf{Algorithm 9.} Lasso homotopy algorithm\\
\textbf{Task:} To addrss the Lasso problem:\\~~~~~~~ $\hat{{\bm\alpha}}=\arg\min_{{\bm\alpha}}\|\bm{y}-X{\bm\alpha}\|_2^2 ~~ s.t. ~~\|{\bm\alpha}\|_1\leq \varepsilon$\\
\midrule
\textbf{Input:} Probe sample $y$, measurement matrix $X$.\\
\textbf{Initialization:} $l=1$, initial solution ${\bm\alpha}_l$ and its support set $\Lambda_l$.\\

Repeat:\\
~~~~Step 1:~Compute update direction $\delta_l$ by using Eq. (\ref{7-5}).\\
~~~~Step 2:~Compute $\tau_l^+$ and $\tau_l^-$ by using Eq. (\ref{7-6}) and Eq. (\ref{7-7}).\\
~~~~Step 3:~Compute the optimal minimum step size $\tau_l^*$ by using\\~~~~~~~~~~~~~ $\tau_l^*=min\{\tau_l^+,\tau_l^-\}$.\\
~~~~Step 4:~Update the solution ${\bm\alpha}_{l+1}$ by using ${\bm\alpha}_{l+1} = {\bm\alpha}_l+\tau_l^*\bm{\delta}_l$.\\
~~~~Step 5:~Update the support set:\\
~~~~~~~~~~~~~If~$\tau_l^+==\tau_l^-$~then\\
~~~~~~~~~~~~~~~~Remove the $i^-$ from the support set, i.e. $\Lambda_{l+1}=\Lambda_l \backslash i^-$.\\
~~~~~~~~~~~~~else\\
~~~~~~~~~~~~~~~~Add the $i^+$ into the support set, i.e. $\Lambda_{l+1}=\Lambda_l \bigcup i^+$\\
~~~~~~~~~~~~~End if\\
~~~~Step 6: $l=l+1$.\\
Until $\|X^T(\bm{y}-X{\bm\alpha})\|_{\infty}=0$\\
\textbf{Output:} ${\bm\alpha}_{l+1}$\\
\bottomrule
\end{tabular}
\end{table}

\subsection{BPDN homotopy}
\noindent
Problem \ref{3-11}, which is called basis pursuit denoising (BPDN) in signal processing, is the unconstrained Lagrangian function of the LASSO problem \ref{3-9}, which is an unconstrained problem. The BPDN homotopy algorithm is very similar to the LASSO homotopy algorithm. If we consider the KKT optimality condition for problem \ref{3-12}, the following condition should be satisfied for the solution ${\bm\alpha}$
\begin{equation}\label{7-8}
\|X^T(\bm{y}-X{\bm\alpha})\|_{\infty} \leq \lambda
\end{equation}
As for any given value of $\lambda$ and the support set $\Lambda$, the following two conditions also need to be satisfied

\begin{equation}\label{7-9}
X_{\Lambda}^T(\bm{y}-X{\bm\alpha})=\lambda \bm{u};~~~
\|X_{\Lambda^c}^T(\bm{y}-X{\bm\alpha})\|_{\infty} \leq \lambda
\end{equation}

The BPDN homotopy algorithm directly computes the homotopy solution by

\begin{equation} \label{7-10}
{\bm\alpha} = \left \{
\begin{array}{lc}
(X_{\Lambda}^TX_{\Lambda})^{-1}(X_{\Lambda}^T\bm{y}-\lambda \bm{u}), & on~\Lambda \\
0, & otherwise\\
\end{array} \right.
\end{equation}
which is somewhat similar to the soft-thresholding operator. The value of the homotopy parameter $\lambda$ is initialized with a large value, which satisfies $\lambda_0>\|X^Ty\|_{\infty}$. As the value of the homotopy parameter $\lambda$ decreases, the BPDN homotopy algorithm traces the solution in the direction of $(X_{\Lambda}^TX_{\Lambda})^{-1}u$ till the critical point is obtained. Each critical point is reached when either an inactive element is transferred into an active element, i.e. its corresponding index should be added into the support set, or an nonzero active element value in ${\bm\alpha}$ shrinks to zero, i.e. its corresponding index should be removed from the support set. Thus, at each critical point, only one element is updated, i.e. one element being either removed from or added into the active set, and each operation is very computationally efficient. The algorithm is terminated when the value of the homotopy parameter is lower than its desired value. The BPDN homotopy algorithm has been summarized in Algorithm 10. For further detail description and analyses, please refer to the literature \cite{efron2004least}.

\begin{table}[htbp]
\centering
\begin{tabular}{m{85mm}}
\toprule
\textbf{Algorithm 10.}  BPDN homotopy algorithm\\
\textbf{Task:} To address the Lasso problem:\\~~~~~~~ $\hat{{\bm\alpha}}=\arg\min_{{\bm\alpha}}\|\bm{y}-X{\bm\alpha}\|_2^2+\lambda \|{\bm\alpha}\|_1$\\
\midrule
\textbf{Input:} Probe sample $\bm{y}$, measurement matrix $X$.\\
\textbf{Initialization:} $l=0$, initial solution ${\bm\alpha}_0$ and its support set $\Lambda_0$, a large value $\lambda_0$, step size $\tau$, tolerance $\varepsilon$.\\

Repeat:\\
~~~~Step 1:~Compute update direction $\bm{\delta}_{l+1}$ by using\\~~~~~~~~~~~~ $\bm{\delta}_{l+1}=(X_{\Lambda}^TX_{\Lambda})^{-1}\bm{u}_l$ .\\
~~~~Step 2:~Update the solution ${\bm\alpha}_{l+1}$ by using Eq. (\ref{7-10}).\\
~~~~Step 3:~Update the support set and the sign sequence set.\\
~~~~Step 6:~$\lambda_{l+1}=\lambda_l-\tau$, $l=l+1$.\\
Until $\lambda\leq \varepsilon$\\
\textbf{Output:} ${\bm\alpha}_{l+1}$\\
\bottomrule
\end{tabular}
\end{table}

\subsection{ Iterative Reweighting $l_1$-norm minimization via homotopy }
\noindent
Based on the homotopy algorithm, Asif and Romberg \cite{asif2012fast} presented a enhanced sparse representation objective function, a weighted $l_1$-norm minimization, and then provided two fast and accurate solutions, i.e. the iterative reweighting algorithm, which updated the weights with a new ones, and the adaptive reweighting algorithm, which adaptively selected the weights in each iteration. Here the iterative reweighting algorithm via homotopy is introduced. The objective function of the weighted $l_1$-norm minimization is formulated as

\begin{equation}\label{7-11}
argmin \frac{1}{2}\|X{\bm\alpha}-\bm{y}\|_2^2+\|W{\bm\alpha}\|_1
\end{equation}
where $W=diag[w_1,w_2,\cdots,w_N]$ is the weight of the $l_1$-norm and also is a diagonal matrix. For more explicit description, problem \ref{7-11} can be rewritten as

\begin{equation}\label{7-12}
argmin \frac{1}{2}\|X{\bm\alpha}-\bm{y}\|_2^2+ \sum_{i=1}^{N} w_i|{\bm\alpha}_i|
\end{equation}

A common method \cite{figueiredo2007gra,efron2004least} to update the weight $W$ is achieved by exploiting the solution of problem \ref{7-12}, i.e. ${\bm\alpha}$, at the previous iteration, and for the $i$-th element of the weight $w_i$ is updated by
\begin{equation}\label{7-13}
w_i = \frac{\lambda}{|{\bm\alpha}_i|+\sigma}
\end{equation}
where parameters $\lambda$ and $\sigma$ are both small constants. In order to efficiently update the solution of problem (7-9), the homotopy algorithm introduces a new weight of the $l_1$-norm and a new homotopy based reweighting minimization problem is reformulated as
\begin{equation}\label{7-14}
argmin~\frac{1}{2}\|X{\bm\alpha} - \bm{y}\|_2^2 + \sum_{i=1}^N ((1-\sigma)\hat{w}_i + \sigma \hat{w}_i) |{\bm\alpha}_i|
\end{equation}

where $\hat{w}_i$ denotes the new obtained weight by the homotopy algorithm, parameter $\tau$ is denoted as the homotopy parameter varying from 0 to 1. Apparently, problem \ref{7-14} can be evolved to problem \ref{7-12} with the increasing value of the homotopy parameter by tracing the homotopy solution path. Similar to the LASSO homotopy algorithm, problem \ref{7-14} is also piecewise linear along the homotopy path, and for any value of $\sigma$, the following conditions should be satisfied
\begin{equation} \label{7-15}
\begin{array}{lr}
x_i^T(X{\bm\alpha}-y) = -((1-\sigma)w_i + \sigma \hat{w}_i)u_i &for~i\in \Lambda~(a) \\
|x_i^T(X{\bm\alpha}-y)|<(1-\sigma)w_i + \sigma \hat{w}_i &~for~i\in \Lambda^c~(b) \\
\end{array}
\end{equation}

where $x_i$ is the $i$-th column of the measurement $X$, $w_i$ and $\hat{w}_i$ are the given weight and new obtained weight, respectively. Moreover, for the optimal step size $\sigma$, when the homotopy parameter changes from $\sigma$ to $\sigma+\tau$ in the update direction $\bm{\delta}$, the following optimality conditions also should be satisfied

\begin{equation} \label{7-16}
\begin{array}{l}
X_{\Lambda}^T(X{\bm\alpha}-\bm{y})+\tau X_{\Lambda}^TX\bm{\delta} = \\-((1-\sigma)W+ \sigma \hat{W})\bm{u} +\tau(W-\hat{W})\bm{u}~~(a) \\
|\bm{p}-\tau \bm{q}|\leq r+\tau s  ~~~~~~~~~~~~~~~~~~~~~~~~~~(b) \\
\end{array}
\end{equation}
where $\bm{u}$ is the sign sequence of ${\bm\alpha}$ on its support $\Lambda$, $p_i=x_i^T(X{\bm\alpha}-\bm{y})$, $q_i=x_i^TX\bm{\delta}$, $r_i=(1-\sigma)w_i + \sigma \hat{w}_i$ and $s_i=\hat{w}_i-w_i$. Thus, at the $l$-th stage (if $(X_i^TX_i)^{-1}$ exists), the update direction of the homotopy algorithm can be computed by
\begin{equation} \label{7-17}
\bm{\delta}_l = \left \{
\begin{array}{lc}
(X_{\Lambda}^TX_{\Lambda})^{-1}(W-\hat{W})\bm{u}, & on~\Lambda \\
0, & otherwise\\
\end{array} \right.
\end{equation}

The step size which can lead to a critical point can be computed by $\tau_l^*=min({\tau_l^+,\tau_l^-})$, and $\tau_l^+$ and $\tau_l^-$ are computed by
\begin{equation} \label{7-18}
\tau_l^+ = min_{i\in \Lambda^c} \left( \frac{r_i-p_i}{q_i-s_i}, \frac{-r_i-p_i}{q_i+s_i} \right)_+
\end{equation}
\begin{equation} \label{7-19}
\tau_l^- = min_{i\in \Lambda} \left( \frac{-{\bm\alpha}_l^i}{\bm{\delta}_l^i} \right)_+
\end{equation}

where $\tau_l^+$ is the minimum step size so that the index $i^+$ should be added into the support set and $\tau_l^-$ is the minimum step size that shrinks the value of a nonzero active element to zero at the index $i^-$. The solution and homotopy parameter are updated by ${\bm\alpha}_{l+1}={\bm\alpha}_l+\tau_l^*\bm{\delta}$, and $\sigma_{l+1}=\sigma_l+\tau_l^*$, respectively. The homotopy algorithm updates its support set and sign sequence accordingly until the new critical point of the homotopy parameter $\sigma_{l+1}=1$. The main steps of this algorithm are summarized in Algorithm 11 and more information can be found in literature \cite{asif2012fast}.

\begin{table}[htbp]
\centering
\begin{tabular}{m{85mm}}
\toprule
\textbf{Algorithm 11.}  Iterative reweighting homotopy algorithm for weighted $l_1$-norm minimization\\
\textbf{Task:} To addrss the  weighted $l_1$-norm minimization:\\~~~~~~~ $\hat{{\bm\alpha}}=arg\min \frac{1}{2}\|X{\bm\alpha}-\bm{y}\|_2^2+W\|{\bm\alpha}\|_1$\\
\midrule
\textbf{Input:} Probe sample $y$, measurement matrix $X$.\\
\textbf{Initialization:} $l=1$, initial solution ${\bm\alpha}_l$ and its support set $\Lambda_l$, $\sigma_1=0$.\\
Repeat:\\
~~~~Step 1:~Compute update direction $\bm{\delta}_l$ by using Eq. (\ref{7-17}).\\
~~~~Step 2:~Compute $p$, $q$, $r$ and $s$ by using Eq. (\ref{7-16}).\\
~~~~Step 2:~Compute $\tau_l^+$ and $\tau_l^-$ by using Eq. (\ref{7-18}) and Eq. (\ref{7-19}).\\
~~~~Step 3:~Compute the step size $\tau_l^*$ by using\\~~~~~~~~~~~~~ $\tau_l^*=min\{\tau_l^+,\tau_l^-\}$.\\
~~~~Step 4:~Update the solution ${\bm\alpha}_{l+1}$ by using ${\bm\alpha}_{l+1} = {\bm\alpha}_l+\tau_l^*\bm{\delta}_l$.\\
~~~~Step 5:~Update the support set:\\
~~~~~~~~~~~~~If~$\tau_l^+==\tau_l^-$~then\\
~~~~~~~~~~~~~~~~Shrink the value to zero at the index $i^-$ and remove $i^-$,\\~~~~~~~~~~~~~~~~i.e. $\Lambda_{l+1}=\Lambda_l \backslash i^-$.\\
~~~~~~~~~~~~~else\\
~~~~~~~~~~~~~~~~Add the $i^+$ into the support set, i.e. $\Lambda_{l+1}=\Lambda_l \bigcup i^+$\\
~~~~~~~~~~~~~End if\\
~~~~Step 6: $\sigma_{l+1}=\sigma_l+\tau_l$ and $l=l+1$.\\
Until $\sigma_{l+1}=1$\\
\textbf{Output:} ${\bm\alpha}_{l+1}$\\
\bottomrule
\end{tabular}
\end{table}

\subsection{Other homotopy algorithms for sparse representation }
\noindent
The general principle of the homotopy method is to reach the optimal solution along with the homotopy solution path by evolving the homotopy parameter from a known initial value to the final expected value. There are extensive hotomopy algorithms, which are related to the sparse representation with the $l_1$-norm regularization.
Malioutov et al. first exploited the homotopy method to choose a suitable parameter for $l_1$-norm regularization with a noisy term in an underdetermined system and employed the homotopy continuation-based method to solve BPDN for sparse signal processing \cite{malioutov2005homotopy}. Garrigues and Ghaoui \cite{garrigues2009homotopy} proposed a modified homotopy algorithm to solve the Lasso problem with online observations by optimizing the homotopy parameter from the current solution to the solution after obtaining the next new data point. Efron et al. \cite{efron2004least} proposed a basic pursuit denoising (BPDN) homotopy algorithm, which shrinked the parameter to a final value with series of efficient optimization steps. Similar to BPDN homotopy, Asif \cite{asif2008primal} presented a homotopy algorithm for the Dantzing selector (DS) under the consideration of primal and dual solution. Asif and Romberg \cite{asif2009dynamic} proposed a framework of dynamic updating solutions for solving $l_1$-norm minimization programs based on homotopy algorithm and demonstrated its effectiveness in addressing the decoding issue. More recent literature related to homotopy algorithms can be found in the streaming recovery framework \cite{asif2013sparse} and a summary \cite{asif2013dynamic}.

\section{The applications of the sparse representation method}
\noindent
Sparse representation technique has been successfully implemented to numerous applications, especially in the fields of computer vision, image processing, pattern recognition and machine learning. More specifically, sparse representation has also been successfully applied to extensive real-world applications, such as image denoising, deblurring, inpainting, super-resolution, restoration, quality assessment, classification, segmentation, signal processing, object tracking, texture classification, image retrieval, bioinformatics, biometrics and other artificial intelligence systems. Moreover, dictionary learning is one of the most typical representative examples of sparse representation for realizing the sparse representation of a signal. In this paper, we only concentrate on the three applications of sparse representation, i.e. sparse representation in dictionary learning, image processing, image classification and visual tracking.

\subsection{Sparse representation in dictionary learning}
\noindent The history of modeling dictionary could be traced back to 1960s, such as the fast Fourier transform (FFT) \cite{cooley1965algorithm}. An over-complete dictionary that can lead to sparse representation is usually achieved by exploiting pre-specified set of transformation functions, i.e. transform domain method \cite{mallat2008wavelet}, or is devised based on learning, i.e. dictionary learning methods \cite{rubinstein2010dic}. Both of the transform domain and dictionary learning based methods transform image samples into other domains and the similarity of transformation coefficients are exploited \cite{shao2014heuristic}. The difference between them is that the transform domain methods usually utilize a group of fixed transformation functions to represent the image samples, whereas the dictionary learning methods apply sparse representations on a over-complete dictionary with redundant information. Moreover, exploiting the pre-specified transform matrix in transform domain methods is attractive because of its fast and simplicity. Specifically, the transform domain methods usually represent the image patches by using the orthonormal basis such as over-complete wavelets transform \cite{simoncelli1996noise}, super-wavelet transform \cite{he2015automatic}, bandelets \cite{le2005sparse}, curvelets transform \cite{starck2002curvelet}, contourlets transform \cite{do2005contourlet} and steerable wavelet filters \cite{simoncelli1992shiftable}. However, the dictionary learning methods exploiting sparse representation have the potential capabilities of outperforming the pre-determined dictionaries based on transformation functions. Thus, in this subsection we only focus on the modern over-complete dictionary learning methods.

An effective dictionary can lead to excellent reconstruction results and satisfactory applications, and the choice of dictionary is also significant to the success of sparse representation technique. Different tasks have different dictionary learning rules. For example, image classification requires that the dictionary contains discriminative information such that the solution of sparse representation possesses the capability of distinctiveness. The purpose of dictionary learning is motivated from sparse representation and aims to learn a faithful and effective dictionary to largely approximate or simulate the specific data. In this section, some parameters are defined as matrix $Y=[\bm{y}_1, \bm{y}_2, \cdots, \bm{y}_N]$, matrix $X=[\bm{x}_1, \bm{x}_2, \cdots, \bm{x}_t]^T$, and dictionary $D=[\bm{d}_1, \bm{d}_2, \cdots, \bm{d}_M]$.

From the notations of the literature \cite{shi2011non, cheng2013sparse}, the framework of dictionary learning can be generally formulated as an optimization problem
\begin{equation} \label{8-1}
\arg\min_{D\in \Omega, x_i} \left \{ \frac{1}{N}\sum_{i=1}^N( \frac{1}{2} \|\bm{y}_i - D \bm{x}_i\|_2^2 + \lambda P(\bm{x}_i)) \right\}
\end{equation}
where $\Omega = \{D=[\bm{d}_1,\bm{d}_2,\cdots,\bm{d}_M]:\bm{d}_i^T\bm{d}_i=1, i=1,2,\cdots,M\}$ ($M$ here may not be equal to $N$), $N$ denotes the number of the known data set (eg. training samples in image classification), $\bm{y}_i$ is the $i$-th sample vector from a known set, $D$ is the learned dictionary and $\bm{x}_i$ is the sparsity vector. $P(\bm{x}_i)$ and $\lambda$ are the penalty or regularization term and a tuning parameter, respectively. The regularization term of problem \ref{8-1} controls the degree of sparsity. That is, different kinds of the regularization terms can immensely dominate the dictionary learning results.

One spontaneous idea of defining the penalty term $P(\bm{x}_i)$ is to introduce the $l_0$-norm regularization, which leads to the sparsest solution of problem \ref{8-1}. As a result, the theory of sparse representation can be applied to dictionary learning. The most representative dictionary learning based on the $l_0$-norm penalty is the K-SVD algorithm \cite{bruckstein2009sparse}, which is widely used in image denoising. Because the solution of $l_0$-norm regularization is usually a NP-hard problem, utilizing a convex relaxation strategy to replace $l_0$-norm regularization is an advisable choice for dictionary learning. As a convex relaxation method of $l_0$-norm regularization, the $l_1$-norm regularization based dictionary learning has been proposed in large numbers of dictionary learning schemes. In the stage of convex relaxation methods, there are three optimal forms for updating a dictionary: the one by one atom updating method, group atoms updating method, and all atoms updating method \cite{shi2011non}. Furthermore, because of over-penalization in $l_1$-norm regularization, non-convex relaxation strategies also have been employed to address dictionary learning problems. For example, Fan and Li proposed a smoothly clipped absolution deviation (SCAD) penalty \cite{fan2001variable}, which employed an iterative approximate Newton-Raphson method for penalizing least sequences and exploited the penalized likelihood approaches for variable selection in linear regression models. Zhang introduced and studied the non-convex minimax concave (MC) family \cite{zhang2010nearly} of non-convex piecewise quadratic penalties to make unbiased variable selection for the estimation of regression coefficients, which was demonstrated its effectiveness by employing an oracle inequality. Friedman proposed to use the logarithmic penalty for a model selection \cite{friedman2012fast} and used it to solve the minimization problems with non-convex regularization terms. From the viewpoint of updating strategy, most of the dictionary learning methods always iteratively update the sparse approximation or representation solution and the dictionary alternatively, and more dictionary learning theoretical results and analyses can be found in the literature \cite{rubinstein2010dictionaries, tosic2011dictionary}.

Recently, varieties of dictionary learning methods have been proposed and researchers have attempted to exploit different strategies for implementing dictionary learning tasks based on sparse representation. There are several means to categorize these dictionary learning algorithms into various groups. For example, dictionary learning methods can be divided into three groups in the context of different norms utilized in the penalty term, that is, $l_0$-norm regularization based methods, convex relaxation methods and non-convex relaxation methods \cite{bao2014l0}. Moreover, dictionary learning algorithms can also be divided into three other categories in the presence of different structures. The first category is dictionary learning under the probabilistic framework such as maximum likelihood methods \cite{olshausen1997sparse}, the method of optimal directions (MOD) \cite{engan2000multi}, and the maximum a posteriori probability method \cite{kreutz2003dictionary}. The second category is clustering based dictionary learning approaches such as KSVD \cite{aharon2006svd}, which can be viewed as a generalization of $K$-means. The third category is dictionary learning with certain structures, which are grouped into two significative aspects, i.e. directly modeling the relationship between each atom and structuring the corrections between each atom with purposive sparsity penalty functions. There are two typical models for these kinds of dictionary learning algorithms, sparse and shift-invariant representation of dictionary learning and structure sparse regularization based dictionary learning, such as hierarchical sparse dictionary learning \cite{jenatton2010proximal} and group or block sparse dictionary learning \cite{bengio2009group}. Recently, some researchers \cite{cheng2013sparse} categorized the latest methods of dictionary learning into four groups, online dictionary learning \cite{zhao2011online}, joint dictionary learning \cite{zhang2012robust}, discriminative dictionary learning \cite{zhang2010dksvd}, and supervised dictionary learning \cite{yang2011tag}.

Although there are extensive strategies to divide the available sparse representation based dictionary learning methods into different categories, the strategy used here is to categorize the current prevailing dictionary learning approaches into two main classes: supervised dictionary learning and unsupervised dictionary learning, and then specific representative algorithms are explicitly introduced.

\subsubsection{Unsupervised dictionary learning}
From the viewpoint of theoretical basis, the main difference of unsupervised and supervised dictionary learning relies on whether the class label is exploited in the process of learning for obtaining the dictionary. Unsupervised dictionary learning methods have been widely implemented to solve image processing problems, such as image compression, and feature coding of image representation \cite{bryt2008compression, wang2010locality}.

\noindent (1) KSVD for unsupervised dictionary learning

One of the most representative unsupervised dictionary learning algorithms is the KSVD method \cite{aharon2006svd}, which is a modification or an extension of method of directions (MOD) algorithm. The objective function of KSVD is
\begin{equation} \label{8-2}
\arg\min_{D, X} \{  \| Y - D X \|_F^2 \} ~~s.t.~~ \| \bm{x}_i \|_0 \leq k,~ i=1,2,\cdots,N
\end{equation}
where $Y\in \mathbb{R}^{d \times N}$ is the matrix composed of all the known examples, $D \in \mathbb{R}^{d \times N}$ is the learned dictionary, $X \in \mathbb{R}^{N \times N}$ is the matrix of coefficients, $k$ is the limit of sparsity and $\bm{x}_i$ denotes the $i$-th row vector of the matrix $X$. Problem \ref{8-2} is a joint optimization problem with respect to $D$ and $X$, and the natural method is to alternatively optimize the $D$ and $X$ iteratively.

\begin{table}[htbp]
\centering
\begin{tabular}{m{85mm}}
\toprule
\textbf{Algorithm 12.} The K-SVD algorithm for dictionary learning\\
\textbf{Task:} Learning a dictionary $D$:~~$\arg\min_{D, X} \| Y - D X \|_F^2~~s.t.~~\| \bm{x}_i \|_0 \leq k,~ i=1,2,\cdots,N$\\
\midrule
\textbf{Input:} The matrix composed of given samples $Y=[\bm{y}_1, \bm{y}_2, \cdots, \bm{y}_m]$. \\
\textbf{Initialization:}  Set the initial dictionary $D$ to the $l_2$--norm unit matrix, $i=1$.\\
While not converged do \\
~~~Step 1:~For each given example $\bm{y}_i$, employing the classical sparse\\~~~~~~~representation with $l_0$-norm regularization to solve problem \ref{8-3} \\~~~~~~~for further estimating $X^i$,set $l=1$.\\
~~~~~~~~~~~ While $l$ is not equal to $k$ do \\
~~~~~~~~~~~ ~~~~Step 2:~Compute the overall representation residual \\~~~~~~~~~~~ ~~~~~~~~~ $E_l = Y- \sum_{j \neq l} \bm{d}_j \bm{x}_j^T$. \\
~~~~~~~~~~~ ~~~~Step 3:~Extract the column items of $E_l$ which corresponds\\ ~~~~~~~~~~~~~~ ~~~~~~ to the nonzero elements of $\bm{x}_l^T$ and obtain $E_l^P$.\\
~~~~~~~~~~~ ~~~~Step 4:~SVD decomposes $E_l^P$ into $E_l^P = U \Lambda V^T$.\\
~~~~~~~~~~~ ~~~~Step 5:~Update $\bm{d}_l$ to the first column of $U$ and update\\ ~~~~~~~~~~~~~~ ~~~~~~ corresponding coefficients in $\bm{x}_l^T$ by $\Lambda (1,1)$ times the \\ ~~~~~~~~~~~~~~ ~~~~~~~first column of $V$.\\
~~~~~~~~~~~ ~~~~Step 6:~$l=l+1$.\\
~~~~~~~~~~~ End While\\
~~~Step 7:~$i=i+1$.\\
End While \\
\textbf{Output:}~dictionary $D$\\
\bottomrule
\end{tabular}
\end{table}

More specifically, when fixing dictionary $D$, problem \ref{8-2} is converted to

\begin{equation} \label{8-3}
\arg\min_{X}   \| Y - D X \|_F^2  ~~~~s.t.~~~~ \| \bm{x}_i \|_0 \leq k,~ i=1,2,\cdots,N
\end{equation}
which is called sparse coding and $k$ is the limit of sparsity. Then, its subproblem is considered as follows:
$$ \arg\min_{\bm{x}_i}  \| \bm{y}_i - D \bm{x}_i \|_2^2  ~~~~s.t.~~~~ \| \bm{x}_i \|_0 \leq k,~ i=1,2,\cdots,N $$
where we can iteratively resort to the classical sparse representation with $l_0$ -norm regularization such as MP and OMP, for estimating $\bm{x}_i$.

When fixing $X$, problem \ref{8-3} becomes a simple regression model for obtaining $D$, that is
\begin{equation} \label{8-4}
\hat D = \arg\min_{D}   \| Y - D X \|_F^2
\end{equation}
where $\hat D = YX^{\dag} = YX^T (XX^T)^{-1}$ and the method is called MOD. Considering that the computational complexity of the inverse problem in solving problem \ref{8-4} is $O(n^3)$, it is favorable, for further improvement, to update dictionary $D$ by fixing the other variables. The strategy of the KSVD algorithm rewrites the problem \ref{8-4} into
\begin{equation} \label{8-5}
\begin{split}
\hat D =& \arg\min_{D} \| Y - D X \|_F^2 = \arg\min_{D} \|Y- \sum_{j=1}^N \bm{d}_j \bm{x}_j^T \|_F^2 \\=&\arg\min_{D} \| (Y- \sum_{j \neq l} \bm{d}_j \bm{x}_j^T) - \bm{d}_l \bm{x}_l^T \|_F^2
\end{split}
\end{equation}
where $\bm{x}_j$ is the $j$-th row vector of the matrix $X$. First the overall representation residual $E_l = Y- \sum_{j \neq l} \bm{d}_j \bm{x}_j^T$ is computed, and then $\bm{d}_l$ and $\bm{x}_l$ are updated. In order to maintain the sparsity of $\bm{x}_l^T$ in this step, only the nonzero elements of $\bm{x}_l^T$ should be preserved and only the nonzero items of $E_l$ should be reserved, i.e. $E_l^P$, from $\bm{d}_l\bm{x}_l^T$. Then, SVD decomposes $E_l^P$ into $E_l^P = U \Lambda V^T$, and then updates dictionary $\bm{d}_l$. The specific KSVD algorithm for dictionary learning is summarized to Algorithm 12 and more information can be found in the literature \cite{aharon2006svd}.

\noindent (2) Locality constrained linear coding for unsupervised dictionary learning

The locality constrained linear coding (LLC) algorithm \cite{wang2010locality} is an efficient local coordinate linear coding method, which projects each descriptor into a local constraint system to obtain an effective codebook or dictionary. It has been demonstrated that the property of locality is more essential than sparsity, because the locality must lead to sparsity but not vice-versa, that is, a necessary condition of sparsity is locality, but not the reverse \cite{wang2010locality}.

Assume that $Y=[\bm{y}_1, \bm{y}_2, \cdots, \bm{y}_N] \in \mathbb{R}^{d \times N}$ is a matrix composed of local descriptors extracted from examples and the objective dictionary $D=[\bm{d}_1, \bm{d}_2, \cdots, \bm{d}_N] \in \mathbb{R}^{d \times N}$. The objective function of LLC is formulated as

\begin{equation} \label{8-6}
\begin{split}
\arg\min_{x_i, D} \sum_{i=1}^N \| \bm{y}_i - D \bm{x}_i \|_2^2 + \mu \| \bm{b} \odot \bm{x}_i \|_2^2 \\~~s.t.~~\mathbf{1}^T\bm{x}_i=1,~i=1,2,\cdots,N
\end{split}
\end{equation}
where $\mu$ is a small constant as a regularization parameter for adjusting the weighting decay speed, $\odot$ is the operator of the element-wise multiplication, $\bm{x}_i$ is the code for $\bm{y}_i$, $\mathbf{1} \in \mathbb{R}^{N\times 1}$ is defined as a vector with all elements as 1 and vector $\bm{b}$ is the locality adaptor, which is, more specifically, set as

\begin{equation} \label{8-7}
\bm{b} = \exp \left( \frac{dist(\bm{y}_i,D)}{\sigma} \right)
\end{equation}
where $dist(\bm{y}_i,D)=[dist(\bm{y}_i,\bm{d}_1), \cdots, dist(\bm{y}_i,\bm{d}_N)]$ and $dist(\bm{y}_i,\bm{d}_j)$ denotes the distance between $\bm{y}_i$ and $\bm{d}_j$ with different distance metrics, such as Euclidean distance and Chebyshev distance. Specifically, the $i$-th value of vector $\bm{b}$ is defined as $\bm{b_i} = \exp \left( \frac{dist(\bm{y}_i,d_i)}{\sigma} \right)$.

The $K$-Means clustering algorithm is applied to generate the codebook $D$, and then the solution of LLC can be deduced as:
\begin{equation} \label{8-8}
\hat {\bm{x}}_i = (C_i + \mu~diag^2(\bm{b})) \backslash \bm{1}
\end{equation}

\begin{equation} \label{8-9}
\bm{x}_i = \hat {\bm{x}}_i/ \ \mathbf{1}^T \hat {\bm{x}}_i
\end{equation}
where the operator $a \backslash b$ denotes $a^{-1}b$, and $C_i = (D^T - \mathbf{1}\bm{y}_i^T)(D^T - \mathbf{1}\bm{y}_i^T)^T$ is the covariance matrix with respect to $\bm{y}_i$. This is called the LLC algorithm. Furthermore, the incremental codebook optimization algorithm has also been proposed to obtain a more effective and optimal codebook, and the objective function is reformulated as
\begin{equation} \label{8-10}
\begin{split}
\arg\min_{x_i, D} \sum_{i=1}^N \| \bm{y}_i - D \bm{x}_i \|_2^2 + \mu \| \bm{b} \odot \bm{x}_i \|_2^2 ~~~\\ s.t.~~\mathbf{1}^T\bm{x}_i=1,~\forall i; \|\bm{d}_j\|_2^2 \leq  1, \forall j
\end{split}
\end{equation}

Actually, the problem \ref{8-10} is a process of feature extraction and the property of `locality' is achieved by constructing a local coordinate system by exploiting the local bases for each descriptor, and the local bases in the algorithm are simply obtained by using the $K$ nearest neighbors of $\bm{y}_i$. The incremental codebook optimization algorithm in problem \ref{8-10} is a joint optimization problem with respect to $D$ and $x_i$, and it can be solved by iteratively optimizing one when fixing the other alternatively. The main steps of the incremental codebook optimization algorithm are summarized in Algorithm 13 and more information can be found in the literature \cite{wang2010locality}.

\begin{table}[htbp]
\centering
\begin{tabular}{m{85mm}}
\toprule
\textbf{Algorithm 13.} The incremental codebook optimization algorithm \\
\textbf{Task:} Learning a dictionary $D$: $\arg\min_{x_i, D} \sum_{i=1}^N \| \bm{y}_i - D \bm{x}_i \|_2^2 + \mu \| \bm{b} \odot \bm{x}_i \|_2^2~~s.t.~~\mathbf{1}^T\bm{x}_i=1,~\forall i; \|\bm{d}_j\|_2^2 \leq  1, \forall j$\\
\midrule
\textbf{Input:} The matrix composed of given samples $Y=[\bm{y}_1, \bm{y}_2, \cdots, \bm{y}_N]$. \\
\textbf{Initialization:}  $i=1$, $\varepsilon = 0.01$, $D$ initialized by $K$-Means clustering algorithm.\\
While $i$ is not equal to $N$ do \\
~~~Step 1:~Initialize $\bm{b}$ with $1 \times N$ zero vector.\\
~~~Step 2:~Update locality constraint parameter $b$ with\\~~~~~~~~~~~~~~~~~~ $\bm{b}_j = \exp \left( - \frac{dist(\bm{y}_i,\bm{d}_j)}{\sigma} \right)$ for $\forall j$.\\
~~~Step 3:~Normalize $\bm{b}$ using the equation $\bm{b} = \frac{ \bm{b} - \bm{b}_{min}}{\bm{b}_{max} - \bm{b}_{min}}$.\\
~~~Step 4:~Exploit the LLC coding algorithm to obtain $\bm{x}_i$.\\
~~~Step 5:~Keep the set of $D^i$, whose corresponding entries of the code $\bm{x}_i$\\~~~~~~~~are greater than $\varepsilon$, and drop out other elements, i.e.\\~~~~~~~ $index \leftarrow \{j~|~abs\{\bm{x}_i(j)\} > \varepsilon\}~\forall j$ and $D^i \leftarrow D(:, index)$.\\
~~~Step 6:~Update $x_i$ exploiting $\arg \max \|\bm{y}_i - D^i\bm{x}_i \|_2^2~~~s.t.~~~\mathbf{1}^T\bm{x}_i=1$.\\
~~~Step 7:~Update dictionary $D$ using a classical gradient descent method\\~~~~~~~~with respect to problem \ref{8-6}.\\
~~~Step 8:~$i=i+1$.\\
End While \\
\textbf{Output:}~dictionary $D$\\
\bottomrule
\end{tabular}
\end{table}

\noindent (3) Other unsupervised dictionary learning methods

A large number of different unsupervised dictionary learning methods have been proposed. The KSVD algorithm and LLC algorithm are only two typical unsupervised dictionary learning algorithms based on sparse representation. Additionally, Jenatton et al. \cite{jenatton2010proximal} proposed a tree-structured dictionary learning problem, which exploited tree-structured sparse regularization to model the relationship between each atom and defined a proximal operator to solve the primal-dual problem. Zhou et al. \cite{zhou2012nonparametric} developed a nonparametric Bayesian dictionary learning algorithm, which utilized hierarchical Bayesian to model parameters and employed the truncated beta-Bernoulli process to learn the dictionary. Ramirez and Shapiro \cite{ramirez2012mdl} employed minimum description length to model an effective framework of sparse representation and dictionary learning, and this framework could conveniently incorporate prior information into the process of sparse representation and dictionary learning. Some other unsupervised dictionary learning algorithms also have been validated. Mairal et al. proposed an online dictionary learning \cite{mairal2010online} algorithm based on stochastic approximations, which treated the dictionary learning problem as the optimization of a smooth convex problem over a convex set and employed an iterative online algorithm at each step to solve the subproblems. Yang and Zhang proposed a sparse variation dictionary learning (SVDL) algorithm \cite{yang2013cvpr} for face recognition with a single training sample, in which a joint learning framework of adaptive projection and a sparse variation dictionary with sparse bases were simultaneously constructed from the gallery image set to the generic image set. Shi et al. proposed a minimax concave penalty based sparse dictionary learning (MCPSDL) \cite{shi2011non} algorithm, which employed a non-convex relaxation online scheme, i.e. a minimax concave penalty, instead of using regular convex relaxation approaches as approximation of $l_0$-norm penalty in sparse representation problem, and designed a coordinate descend algorithm to optimize it. Bao et al. proposed a dictionary learning by proximal algorithm (DLPM) \cite{bao2014l0}, which provided an efficient alternating proximal algorithm for solving the $l_0$-norm minimization based dictionary learning problem and demonstrated its global convergence property.

\subsubsection{Supervised dictionary learning}

Unsupervised dictionary learning just considers that the examples can be sparsely represented by the learned dictionary and leaves out the label information of the examples. Thus, unsupervised dictionary learning can perform very well in data reconstruction, such as image denoising and image compressing, but is not beneficial to perform classification. On the contrary, supervised dictionary learning embeds the class label into the process of sparse representation and dictionary learning so that this leads to the learned dictionary with discriminative information for effective classification.

\noindent (1) Discriminative KSVD for dictionary learning \label{subSec8-1}

Discriminative KSVD (DKSVD) \cite{zhang2010dksvd} was designed to solve image classification problems. Considering the priorities of supervised learning theory in classification, DKSVD incorporates the dictionary learning with discriminative information and classifier parameters into the objective function and employs the KSVD algorithm to obtain the global optimal solution for all parameters. The objective function of the DKSVD algorithm is formulated as

\begin{equation} \label{8-11}
\begin{split}
\langle D,C,X \rangle = \arg \min_{D,C,X} \|Y-DX\|_F^2 +\mu \| H-CX\|_F^2 \\ + \eta \|C\|_F^2~~~s.t.~~~\|\bm{x}_i\|_0 \leq k
\end{split}
\end{equation}
where $Y$ is the given input samples, $D$ is the learned dictionary, $X$ is the coefficient term, $H$ is the matrix composed of label information corresponding to $Y$, $C$ is the parameter term for classifier, and $\eta$ and $\mu$ are the weights. With a view to the framework of KSVD, problem \ref{8-11} can be rewritten as

\begin{equation} \label{8-12}
\begin{split}
\langle D,C,X \rangle = \arg \min_{D,C,X} \| \left( \begin{array}{c} Y\\ \sqrt{\mu} H\\ \end{array} \right) - \left( \begin{array}{c} D\\ \sqrt{\mu} C\\ \end{array} \right) X \|_F^2 \\ + \eta \|C\|_F^2~~~s.t.~~~\|\bm{x}_i\|_0 \leq k
\end{split}
\end{equation}

In consideration of the KSVD algorithm, each column of the dictionary will be normalized to $l_2$-norm unit vector and $\left(\begin{array}{c} D\\ \sqrt{\mu} C\\ \end{array} \right)$ will also be normalized, and then the penalty term $\|C\|_F^2$ will be dropped out and problem \ref{8-12} will be reformulated as

\begin{equation} \label{8-13}
\langle Z,X \rangle = \arg \min_{Z,X} \|  W - Z X \|_F^2 ~~~s.t.~~~\|\bm{x}_i\|_0 \leq k
\end{equation}
where $W=\left( \begin{array}{c} Y\\ \sqrt{\mu} H\\ \end{array} \right)$, $Z=\left( \begin{array}{c} D\\ \sqrt{\mu} C\\ \end{array} \right)$ and apparently the formulation \ref{8-13} is the same as the framework of KSVD \cite{aharon2006svd} in Eq. \ref{8-2} and it can be efficiently solved by the KSVD algorithm.

More specifically, the DKSVD algorithm contains two main phases: the training phase and classification phase. For the training phase, $Y$ is the matrix composed of the training samples and the objective is to learn a discriminative dictionary $D$ and the classifier parameter $C$. DKSVD updates $Z$ column by column and for each column vector $\bm{z}_i$, DKSVD employs the KSVD algorithm to obtain $\bm{z}_i$ and its corresponding weight. Then, the DKSVD algorithm normalizes the dictionary $D$ and classifier parameter $C$ by

\begin{equation} \label{8-14}
\begin{array}{lll}
D' &=& [\bm{d}'_1,\bm{d}'_2, \cdots ,\bm{d}'_M] = [\frac{\bm{d}_1}{\|\bm{d}_1\|}, \frac{\bm{d}_2}{\|\bm{d}_2\|}, \cdots, \frac{\bm{d}_M}{\|\bm{d}_M\|}]\\
C' &=& [\bm{c}'_1,\bm{c}'_2, \cdots ,\bm{c}'_M] = [\frac{\bm{c}_1}{\|\bm{d}_1\|}, \frac{\bm{c}_2}{\|\bm{d}_2\|}, \cdots, \frac{\bm{c}_M}{\|\bm{d}_M\|}]\\
\bm{x}'_i &=& \bm{x}_i \times\|\bm{d}_i\|\\
\end{array}
\end{equation}

For the classification phase, $Y$ is the matrix composed of the test samples. Based on the obtained learning results $D'$ and $C'$, the sparse coefficient matrix $\hat {\bm{x}}_i$ can be obtained for each test sample $\bm{y}_i$ by exploiting the OMP algorithm, which is to solve

\begin{equation} \label{8-15}
\hat {\bm{x}}_i = \arg \min \|\bm{y}_i - D' \bm{x}'_i \|_2^2~~~s.t.~~~\|\bm{x}'_i\|_0 \leq k
\end{equation}
On the basis of the corresponding sparse coefficient $\hat {\bm{x}}_i$, the final classification, for each test sample $\bm{y}_i$, can be performed by judging the label result by multiplying $\hat {\bm{x}}_i$ by classifier $C'$, that is,
\begin{equation} \label{8-16}
label = C' \times  \hat {\bm{x}}_i
\end{equation}
where the $label$ is the final predicted label vector. The class label of $\bm{y}_i$ is the determined class index of $label$.

The main highlight of DKSVD is that it employs the framework of KSVD to simultaneously learn a discriminative dictionary and classifier parameter, and then utilizes the efficient OMP algorithm to obtain a sparse representation solution and finally integrate the sparse solution and learned classifier for ultimate effective classification.

\noindent (2) Label consistent KSVD for discriminative dictionary learning

Because of the classification term, a competent dictionary can lead to effectively classification results. The original sparse representation for face recognition \cite{wright2009robust} regards the raw data as the dictionary, and then reports its promising classification results. In this section, a label consistent KSVD (LC-KSVD) \cite{jiang2011lcksvd, jiang2013label} is introduced to learn an effective discriminative dictionary for image classification. As an extension of D-KSVD, LC-KSVD exploits the supervised information to learn the dictionary and integrates the process of constructing the dictionary and optimal linear classifier into a mixed reconstructive and discriminative objective function, and then jointly obtains the learned dictionary and an effective classifier. The objective function of LC-KSVD is formulated as

\begin{equation} \label{8-17}
\begin{split}
\langle D,A,C,X \rangle = \arg \min_{D,A,C,X} \|Y-DX\|_F^2 + \mu \|L-AX\|_F^2 \\ + \eta \|H-CX\|_F^2~s.t.~\|\bm{x}_i\|_0 \leq k
\end{split}
\end{equation}
where the first term denotes the reconstruction error, the second term denotes the discriminative sparse-code error, and the final term denotes the classification error. $Y$ is the matrix composed of all the input data, $D$ is the learned dictionary, $X$ is the sparse code term, $\mu$ and $\eta$ are the weights of the corresponding contribution items, $A$ is a linear transformation matrix, $H$ is the matrix composed of label information corresponding to $Y$, $C$ is the parameter term for classifier and $L$ is a joint label matrix for labels of $Y$ and $D$. For example, providing that $Y=[y_1\dots y_4]$ and $D=[d_1\dots d_4]$ where $y_1, y_2, d_1$ and $ d_2$ are from the first class, and $y_3, y_4, d_3$ and $d_4$ are from the second class, and then the joint label matrix $L$ can be defined as $L=\left( \begin{array}{cccc} 1&1&0&0\\ 1&1&0&0\\ 0&0&1&1\\0&0&1&1 \\ \end{array} \right)$. Similar to the DKSVD algorithm, the objective function \ref{8-17} can also be reformulated as

\begin{equation} \label{8-18}
\langle Z,X \rangle = \arg \min_{Z,X} \| T - Z X \|_2^2 ~~~s.t.~~~\|\bm{x}_i\|_0 \leq k
\end{equation}
where $T=\left( \begin{array}{c} Y\\ \sqrt{\mu} L\\ \sqrt{\eta} H\\ \end{array} \right)$, $Z=\left( \begin{array}{c} D\\ \sqrt{\mu} A\\ \sqrt{\eta} C\\ \end{array} \right)$.

The learning process of the LC-KSVD algorithm, as is DKSVD, can be separated into two sections, the training term and the classification term. In the training section, since problem \ref{8-18} completely satisfies the framework of KSVD, the KSVD algorithm is applied to update $Z$ atom by atom and compute $X$. Thus $Z$ and $X$ can be obtained. Then, the LC-KSVD algorithm normalizes dictionary $D$, transform matrix $A$, and the classifier parameter $C$ by

\begin{equation} \label{8-19}
\begin{array}{lll}
D' &=& [\bm{d}'_1,\bm{d}'_2, \cdots ,\bm{d}'_M] = [\frac{\bm{d}_1}{\|\bm{d}_1\|}, \frac{\bm{d}_2}{\|\bm{d}_2\|}, \cdots, \frac{\bm{d}_M}{\|\bm{d}_M\|}]\\
A' &=& [\bm{a}'_1,\bm{a}'_2, \cdots ,\bm{a}'_M] = [\frac{\bm{a}_1}{\|\bm{d}_1\|}, \frac{\bm{a}_2}{\|\bm{d}_2\|}, \cdots, \frac{\bm{a}_M}{\|\bm{d}_M\|}]\\
C' &=& [\bm{c}'_1,\bm{c}'_2, \cdots ,\bm{c}'_M] = [\frac{\bm{c}_1}{\|\bm{d}_1\|}, \frac{\bm{c}_2}{\|\bm{d}_2\|}, \cdots, \frac{\bm{c}_M}{\|\bm{d}_M\|}]\\
\end{array}
\end{equation}

In the classification section, $Y$ is the matrix composed of the test samples. On the basis of the obtained dictionary $D'$, the sparse coefficient $\hat {\bm{x}}_i$ can be obtained for each test sample $\bm{y}_i$ by exploiting the OMP algorithm, which is to solve
\begin{equation} \label{8-20}
\hat {\bm{x}}_i = \arg \min \|\bm{y}_i - D' \bm{x}'_i \|_2^2~~~s.t.~~~\|\bm{x}'_i\|_0 \leq k
\end{equation}

The final classification is based on a simple linear predictive function
\begin{equation} \label{8-21}
l = \arg \max_f \{f=C' \times \hat {\bm{x}}_i\}
\end{equation}
where $f$ is the final predicting label vector and the test sample $\bm{y}_i$ is classified as a member of the $l$-th class.

The main contribution of LC-KSVD is to jointly incorporate the discriminative sparse coding term and classifier parameter term into the objective function for learning a discriminative dictionary and classifier parameter. The LC-KSVD demonstrates that the obtained solution, compared to other methods, can prevent learning a suboptimal or local optimal solution in the process of learning a dictionary \cite{jiang2011lcksvd}.

\noindent (3) Fisher discrimination dictionary learning for sparse representation

Fisher discrimination dictionary learning (FDDL) \cite{yang2011fisher} incorporates the supervised information (class label information) and the Fisher discrimination message into the objective function for learning a structured discriminative dictionary, which is used for pattern classification. The general model of FDDL is formulated as

\begin{equation} \label{8-22}
J(D,X) = \arg \min_{D,X} \{ f(Y,D,X) + \mu \| X\|_1 + \eta g(X) \}
\end{equation}
where $Y$ is the matrix composed of input data, $D$ is the learned dictionary, $X$ is the sparse solution, and $\mu$ and $\eta$ are two constants for tradeoff contributions. The first component is the discriminative fidelity term, the second component is the sparse regularization term, and the third component is the discriminative coefficient term, such as Fisher discrimination criterion in Eq. (\ref{8-23}).

Considering the importance of the supervised information, i.e. label information, in classification, FDDL respectively updates the dictionary and computes the sparse representation solution class by class. Assume that $Y^i$ denotes the matrix of $i$-th class of input data, vector $X^i$ denotes the sparse representation coefficient of the learned dictionary $D$ over $Y^i$ and $X_j^i$ denotes the matrix composed of the sparse representation solutions, which correspond to the $j$-th class coefficients from $X^i$. $D^i$ is denoted as the learned dictionary corresponding to the $i$-th class. Thus, the objective function of FDDL is

\begin{equation} \label{8-23}
\begin{split}
J(D,X) = \arg \min_{D,X} ( \sum_{i=1}^c f(Y^i,D,X^i) + \mu \| X\|_1 + \\\eta (tr(S_W(X)-S_B(X))+ \lambda \|X \|_F^2 ) )
\end{split}
\end{equation}
where $f(Y^i, D, X^i) = \|Y^i -D X^i\|_F^2 + \|Y^i -D^i X_i^i\|_F^2 + \sum_{j \neq i} \|D^j X_j^i\|_F^2$ and $S_W(X)$ and $S_B(X)$ are the within-class scatter of $X$ and between-class scatter of $X$, respectively. $c$ is the number of the classes. To solve problem \ref{8-23}, a natural idea of optimization is to alternatively optimize $D$ and $X$ class by class, and then the process of optimization is briefly introduced.

When fixing $D$, problem \ref{8-23} can be solved by computing $X^i$ class by class, and its sub-problem is formulated as
\begin{equation} \label{8-24}
J(X^i) = \arg \min_{X^i} \left( f(Y^i,D,X^i) + \mu \| X^i\|_1 + \eta g(X^i) \right)
\end{equation}
where $g(X^i) = \|X^i - M_i \|_F^2 -\sum_{t=1}^c \|M_t - M\|_F^2 + \lambda \|X^i\|_F^2 $ and $M_j$ and $M$ denote the mean matrices corresponding to the $j$-th class of $X^i$ and $X^i$, respectively. Problem \ref{8-23} can be solved by the iterative projection method in the literature \cite{rosasco2009iterative}.

When fixing ${\bm\alpha}$, problem \ref{8-23} can be rewritten as

\begin{equation} \label{8-25}
\begin{split}
J (D) = \arg\min_{D} (  \|Y^i - D^i X^i - \sum_{j \neq i} D^j X^j \|_F^2 +\\ \| Y^i -D^iX_i^i \|_F^2 + \sum _{j \neq i} \| D^i X_j^i \|_F^2   )
\end{split}
\end{equation}
where $X^i$ here denotes the sparse representation of $Y$ over $D^i$. In this section, each column of the learned dictionary is normalized to a unit vector with $l_2$-norm. The optimization of problem \ref{8-25} computes the dictionary class by class and it can be solved by exploiting the algorithm in the literature \cite{yang2010metaface}.

The main contribution of the FDDL algorithm lies in combining the Fisher discrimination criterion into the process of dictionary learning. The discriminative power comes from the method of constructing the discriminative dictionary using the function $f$ in problem \ref{8-22} and simultaneously formulates the discriminative sparse representation coefficients by exploiting the function $g$ in problem \ref{8-22}.

\noindent (4) Other supervised dictionary learning for sparse representation

Unlike unsupervised dictionary learning, supervised dictionary learning emphasizes the significance of the class label information and incorporates it into the learning process to enforce the discrimination of the learned dictionary. Recently, massive supervised dictionary learning algorithms have been proposed. For example, Yang et al. \cite{yang2010metaface} presented a metaface dictionary learning method, which is motivated by `metagenes' in gene expression data analysis. Rodriguez and Sapiro \cite{ramirez2010classification} produced a discriminative non-parametric dictionary learning (DNDL) framework based on the OMP algorithm for image classification. Kong et al. \cite{kong2012dictionary} introduced a learned dictionary with commonalty and particularity, called DL-COPAR, which integrated an incoherence penalty term into the objective function for obtaining the class-specific sub-dictionary. Gao et al. \cite{gao2013DL} learned a hybrid dictionary, i.e. category-specific dictionary and shared dictionary, which incorporated a cross-dictionary incoherence penalty and self-dictionary incoherence penalty into the objective function for learning a discriminative dictionary. Jafari and Plumbley \cite{jafari2011fast} presented a greedy adaptive dictionary learning method, which updated the learned dictionary with a minimum sparsity index. Some other supervised dictionary learning methods are also competent in image classification, such as supervised dictionary learning in \cite{mairal2009supervised}. Zhou et al. \cite{zhou2012learning} developed a joint dictionary learning algorithm for object categorization, which jointly learned a commonly shared dictionary and multiply category-specific dictionaries for correlated object classes and incorporated the Fisher discriminant fidelity term into the process of dictionary learning. Ramirez et al. proposed a method of dictionary learning with structured incoherence (DLSI) \cite{ramirez2010classification}, which unified the dictionary learning and sparse decomposition into a sparse dictionary learning framework for image classification and data clustering. Ma et al. presented a discriminative low-rank dictionary learning for sparse representation (DLRD\_SR) \cite{ma2012sparse}, in which the sparsity and the low-rank properties were integrated into one dictionary learning scheme where sub-dictionary with discriminative power was required to be low-rank. Lu et al. developed a simultaneous feature and dictionary learning \cite{lu2014simultaneous} method for face recognition, which jointly learned the feature projection matrix for subspace learning and the discriminative structured dictionary. Yang et al. introduced a latent dictionary learning (LDL) \cite{yang2014latent} method for sparse representation based image classification, which simultaneously learned a discriminative dictionary and a latent representation model based on the correlations between label information and dictionary atoms. Jiang et al. presented a submodular dictionary learning (SDL) \cite{jiang2012submodular} method, which integrated the entropy rate of a random walk on a graph and a discriminative term into a unified objective function and devised a greedy-based approach to optimize it. Si et al. developed a support vector guided dictionary learning (SVGDL) \cite{cai2014support} method, which constructed a discriminative term by using adaptively weighted summation of the squared distances for all pairwise of the sparse representation solutions.

\subsection{Sparse representation in image processing}
\noindent
Recently, sparse representation methods have been extensively applied to numerous real-world applications \cite{qing2014new, he2012sparse}. The techniques of sparse representation have been gradually extended and introduced to image processing, such as super-resolution image processing, image denoising and image restoration.

First, the general framework of image processing using sparse representation especially for image reconstruction should be introduced:

Step 1: Partition the degraded image into overlapped patches or blocks.

Step 2: Construct a dictionary, denoted as $D$, and assume that the following sparse representation formulation should be satisfied for each patch or block $x$ of the image:

$\hat{{\bm\alpha}} = arg\min\|{\bm\alpha}\|_p ~~~s.t.~~~\|x-HD{\bm\alpha}\|_2^2 \leq \varepsilon$

where $H$ is a degradation matrix and $0\leq p \leq 1$.

Step 3: Reconstruct each patch or block by exploiting $\hat{x}=D\hat{{\bm\alpha}}$.

Step 4: Put the reconstructed patch to the image at the corresponding location and average each overlapped patches to make the reconstructed image more consistent and natural.

Step 5: Repeat step 1 to 4 several times till a termination condition is satisfied.

The following part of this subsection is to explicitly introduce some image processing techniques using sparse representation.

The main task of super-resolution image processing is to extract the high super-resolution image from its low resolution counterpart and this challenging problem has attracted much attention. The most representative work was proposed to exploit the sparse representation theory to generate a super-resolution (SRSR) image from a single low-resolution image in literature \cite{yang2008image}.

SRSR is mainly performed on two compact learned dictionaries $D_l$ and $D_h$, which are denoted as dictionaries of low-resolution image patches and its corresponding high-resolution image patches, respectively. $D_l$ is directly employed to recover high-resolution images from dictionary $D_h$. Let $X$ and $Y$ denote the high-resolution and its corresponding low-resolution images, respectively. $\bm{x}$ and $\bm{y}$ are a high-resolution image patch and its corresponding low-resolution image patch, respectively. Thus, $\bm{x}=P\bm{y}$ and $P$ is the projection matrix. Moreover, if the low resolution image $Y$ is produced by down-sampling and blurring from the high resolution image $X$, the following reconstruction constraint should be satisfied
\begin{equation} \label{8-26}
Y=SBX
\end{equation}

where $S$ and $B$ are a downsampling operator and a blurring filter, respectively. However, the solution of problem \ref{8-26} is ill-posed because infinite solutions can be achieved for a given low-resolution input image $Y$. To this end, SRSR \cite{yang2008image} provides a prior knowledge assumption, which is formulated as
\begin{equation} \label{8-27}
\bm{x}=D_h {\bm\alpha} ~~s.t.~~ \|{\bm\alpha}\|_0 \leq k
\end{equation}
where $k$ is a small constant. This assumption gives a prior knowledge condition that any image patch $\bm{x}$ can be approximately represented by a linear combination of a few training samples from dictionary $D_h$. As presented in Subsection \ref{subsec3-2}, problem \ref{8-27} is an NP-hard problem and sparse representation with $l_1$-norm regularization is introduced. If the desired representation solution ${\bm\alpha}$ is sufficiently sparse, problem \ref{8-27} can be converted into the following problem:
\begin{equation} \label{8-28}
arg\min \|{\bm\alpha}\|_1 ~~s.t.~~ \|\bm{x}-D_h{\bm\alpha}\|_2^2 \leq \varepsilon
\end{equation}
or
\begin{equation} \label{8-29}
arg\min \|\bm{x}-D_h{\bm\alpha}\|_2^2 + \lambda \|{\bm\alpha}\|_1
\end{equation}
where $\varepsilon$ is a small constant and $\lambda$ is the Lagrange multiplier. The solution of problem \ref{8-28} can be achieved by two main phases, i.e. local model based sparse representation (LMBSR) and enhanced global reconstruction constraint. The first phase of SRSR, i.e. LMBSR, is operated on each image patch, and for each low-resolution image patch $\bm{y}$, the following equation is satisfied

\begin{equation} \label{8-30}
arg\min \|F\bm{y}-FD_l{\bm\alpha}\|_2^2 + \lambda \|{\bm\alpha}\|_1
\end{equation}
where $F$ is a feature extraction operator. One-pass algorithm similar to that of \cite{freeman2002example} is introduced to enhance the compatibility between adjacent patches. Furthermore, a modified optimization problem is proposed to guarantee that the super-resolution reconstruction coincides with the previously obtained adjacent high-resolution patches, and the problem is reformulated as
\begin{equation}\label{8-31}
arg\min \|{\bm\alpha}\|_1~s.t.~\|F\bm{y}-FD_l{\bm\alpha}\|_2^2 \leq \varepsilon_1;~\|\bm{v}-LD_h{\bm\alpha}\|_2^2 \leq \varepsilon_2
\end{equation}
where $\bm{v}$ is the previously obtained high-resolution image on the overlap region, and $L$ refers to the region of overlap between the current patch and previously obtained high-resolution image. Thus problem \ref{8-31} can be rewritten as
\begin{equation}\label{8-32}
arg\min~\|\hat{\bm{y}}-D{\bm\alpha}\|_2^2 + \lambda \|{\bm\alpha}\|_1
\end{equation}
where $\hat{\bm{y}}=\left[ \begin{array}{c} F\bm{y}\\\bm{v}\\ \end{array}\right]$ and $D=\left[ \begin{array}{c} FD_l\\LD_h\\ \end{array}\right]$. Problem \ref{8-32} can be simply solved by previously introduced solution of the sparse representation with $l_1$-norm minimization. Assume that the optimal solution of problem \ref{8-32}, i.e. ${\bm\alpha}^*$, is achieved, the high-resolution patch can be easily reconstructed by $\bm{x}=D_h{\bm\alpha}^*$.

The second phase of SRSR enforces the global reconstruction constraint to eliminate possible unconformity or noise from the first phase and make the obtained image more consistent and compatible. Suppose that the high-resolution image obtained by the first phase is denoted as matrix $X_0$, we project $X_0$ onto the solution space of the reconstruction constraint \ref{8-26} and the problem is formulated as follows
\begin{equation}\label{8-33}
X^* = arg\min \|X-X_0\|_2^2 ~~s.t.~~ Y=SBX
\end{equation}
Problem \ref{8-33} can be solved by the back-projection method in \cite{irani1993motion} and the obtained image $X^*$ is regarded as the final optimal high-resolution image. The entire super-resolution via sparse representation is summarized in algorithm 14 and more information can be found in the literature \cite{yang2008image}.

\begin{table}[htbp]
\centering
\begin{tabular}{m{85mm}}
\toprule
\textbf{Algorithm 14.} Super-resolution via sparse representation \\
\midrule
\textbf{Input:} Training image patches dictionaries $D_l$ and $D_h$, a low-resolution image $Y$. \\
For each overlapped $3\times 3$ patches $\bm{y}$ of $Y$ using one-pass algorithm, from left to right and top to bottom\\
~~~Step 1:~Compute optimal sparse representation coefficients ${\bm\alpha}^*$ in \\~~~~~~~~ problem (\ref{8-32}).\\
~~~Step 2:~Compute the high-resolution patch by $\bm{x}=D_h{\bm\alpha}^*$.\\
~~~~Step 3:~Put the patch $\bm{x}$ into a high-resolution image $X_0$ in corresponding\\~~~~~~~~ location.\\
End\\
~~~Step 4:~Compute the final super-resolution image $X^*$ in problem (\ref{8-33}).\\
\textbf{Output:}~$X^*$\\
\bottomrule
\end{tabular}
\end{table}

Furthermore, extensive other methods based on sparse representation have been proposed to solve the super-resolution image processing problem. For example, Yang et al. presented a modified version called joint dictionary learning via sparse representation (JDLSR) \cite{yang2010image}, which jointly learned two dictionaries that enforced the similarity of sparse representation for low-resolution and high-resolution images. Tang et al. \cite{tang2013greedy} first explicitly analyzed the rationales of the sparse representation theory in performing the super-resolution task, and proposed to exploit the $L_2$-Boosting strategy to learn coupled dictionaries, which were employed to construct sparse coding space. Zhang et al. \cite{zhang2012image} presented an image super-resolution reconstruction scheme by employing the dual-dictionary learning and sparse representation method for image super-resolution reconstruction and Gao et al. \cite{gao2012image} proposed a sparse neighbor embedding method, which incorporated the sparse neighbor search and HoG clustering method into the process of image super-resolution reconstruction. Fernandez-Granda and Candes \cite{fernandez2013super} designed a transform-invariant group sparse regularizer by implementing a data-driven non-parametric regularizers with learned domain transform on group sparse representation for high image super-resolution. Lu et al. \cite{lu2012geometry} proposed a geometry constrained sparse representation method for single image super-resolution by jointly obtaining an optimal sparse solution and learning a discriminative and reconstructive dictionary. Dong et al. \cite{dong2010super} proposed to harness an adaptive sparse optimization with nonlocal regularization based on adaptive principal component analysis enhanced by nonlocal similar patch grouping and nonlocal self-similarity quadratic constraint to solve the image high super-resolution problem. Dong et al. \cite{dong2011image} proposed to integrate an adaptive sparse domain selection and an adaptive regularization based on piecewise autoregressive models into the sparse representations framework for single image super-resolution reconstruction. Mallat and Yu \cite{mallat2010super} proposed a sparse mixing estimator for image super-resolution, which introduced an adaptive estimator models by combining a group of linear inverse estimators based on different prior knowledge for sparse representation.

Noise in an image is unavoidable in the process of image acquisition. The need for sparse representation may arise when noise exists in image data. In such a case, the image with noise may lead to missing information or distortion such that this results in a decrease of the precision and accuracy of image processing. Eliminating such noise is greatly beneficial to many applications. The main goal of image denoising is to distinguish the actual signal and noise signal so that we can remove the noise and reconstruct the genuine image. In the presence of image sparsity and redundancy representation \cite{Elad2010on, elad2010sparse}, sparse representation for image denoising first extracts the sparse image components, which are regarded as useful information, and then abandons the representation residual, which is treated as the image noise term, and finally reconstructs the image exploiting the pre-obtained sparse components, i.e. noise-free image. Extensive research articles for image denoising based on sparse representation have been published. For example, Donoho \cite{bruckstein2009sparse, Donoho2008Fast, donoho1995noising} first discovered the connection between the compressed sensing and image denoising. Subsequently, the most representative work of using sparse representation to make image denoising was proposed in literature \cite{elad2006image}, in which a global sparse representation model over learned dictionaries (SRMLD) was used for image denoising. The following prior assumption should be satisfied: every image block of image $x$, denoted as $z$, can be sparsely represented over a dictionary $D$, i.e. the solution of the following problem is sufficiently sparse:
\begin{equation}\label{8-34}
arg\min_{{\bm\alpha}} \|{\bm\alpha}\|_0~~s.t.~~D{\bm\alpha} = \bm{z}
\end{equation}
And an equivalent problem can be reformulated for a proper value of $\lambda$, i.e.
\begin{equation}\label{8-35}
arg\min_{\bm\alpha} \|D{\bm\alpha} - \bm{z}\|_2^2 + \lambda \|{\bm\alpha} \|_0
\end{equation}
If we take the above prior knowledge into full consideration, the objective function of SRMLD based on Bayesian treatment is formulated as
\begin{equation}\label{8-36}
arg\min_{D,{\bm\alpha}_i,\bm{x}} \delta \|\bm{x}-\bm{y}\|_2^2 + \sum_{i=1}^{M} \|D{\bm\alpha}_i - P_i\bm{x}\|_2^2 + \sum_{i=1}^{M} \lambda_i \|{\bm\alpha}_i \|_0
\end{equation}
where $\bm{x}$ is the finally denoised image, $\bm{y}$ the measured image with white and additive Gaussian white noise, $P_i$ is a projection operator that extracts the $i$-th block from image $x$, $M$ is the number of the overlapping blocks, $D$ is the learned dictionary, ${\bm\alpha}_i$ is the coefficients vector, $\delta$ is the weight of the first term and $\lambda_i$ is the Lagrange multiplier. The first term in \ref{8-36} is the log-likelihood global constraint such that the obtained noise-free image $x$ is sufficiently similar to the original image $y$. The second and third terms are the prior knowledge of the Bayesian treatment, which is presented in problem \ref{8-35}. The optimization of problem \ref{8-35} is a joint optimization problem with respect to $D$, ${\bm\alpha}_i$ and $x$. It can be solved by alternatively optimizing one variable when fixing the others. The process of optimization is briefly introduced below.

When dictionary $D$ and the solution of sparse representation ${\bm\alpha}_i$ are fixed, problem \ref{8-36} can be rewritten as

\begin{equation}\label{8-37}
arg \min_x \delta \|\bm{x}-\bm{y}\|_2^2 + \sum_{i=1}^{M} \|D{\bm\alpha}_i - \bm{z}\|_2^2
\end{equation}
where $\bm{z}=P_i \bm{x}$. Apparently, problem \ref{8-37} is a simple convex optimization problem and has a closed-form solution, which is given by
\begin{equation}\label{8-38}
\bm{x}=\left( \sum_{i=1}^M P_i^T P_i + \delta I\right)^{-1} \left( \sum_{i=1}^M P_i^T D {\bm\alpha}_i +\delta \bm{y}\right)^{-1}
\end{equation}

When $\bm{x}$ is given, problem (\ref{8-36}) can be written as
\begin{equation}\label{8-39}
arg\min_{D,{\bm\alpha}_i} \sum_{i=1}^{M} \|D{\bm\alpha}_i - P_i\bm{x}\|_2^2 + \sum_{i=1}^{M} \lambda_i \|{\bm\alpha}_i \|_0
\end{equation}
where the problem can be divided into $M$ sub-problems and the $i$-th sub-problem can be reformulated as the following dictionary learning problem:
\begin{equation}\label{8-40}
arg\min_{D,{\bm\alpha}_i} \|D{\bm\alpha}_i -\bm{z}\|_2^2~~s.t.~~\|{\bm\alpha}_i\|_0 \leq \tau
\end{equation}
where $\bm{z}=P_i\bm{x}$ and $\tau$ is small constant. One can see that the sub-problem \ref{8-39} is the same as problem \ref{8-2} and it can be solved by the KSVD algorithm previously presented in Subsection \ref{subSec8-1}. The algorithm of image denoising exploiting sparse and redundant representation over learned dictionary is summarized in Algorithm 15, and more information can be found in literature \cite{elad2006image}.

\begin{table}[htbp]
\centering
\begin{tabular}{m{85mm}}
\toprule
\textbf{Algorithm 15.} Image denoising via sparse and redundant representation over learned dictionary\\
\textbf{Task:} To denoise a measured image $y$ from white and additional Gaussian white noise:\\
 $arg\min_{D,{\bm\alpha}_i,x} \delta \|\bm{x}-\bm{y}\|_2^2 + \sum_{i=1}^{M} \|D{\bm\alpha}_i - P_i\bm{x}\|_2^2 + \sum_{i=1}^{M} \lambda_i \|{\bm\alpha}_i \|_0$\\
\midrule
\textbf{Input:} Measured image sample $\bm{y}$, the number of training iteration $T$. \\
\textbf{Initialization:}  $t=1$, set $\bm{x}=\bm{y}$, $D$ initialized by an overcomplete DCT dictionary.\\
While $t\leq T$ do \\
~~~Step 1:~For each image patch $P_i\bm{x}$, employ the KSVD algorithm to update the values of sparse representation solution ${\bm\alpha}_i$ and corresponding dictionary $D$.\\
~~~Step 2:~$t=t+1$\\
End While \\
~~~Step 3:~Compute the value of $\bm{x}$ by using Eq.(\ref{8-38}).\\
\textbf{Output:}~denoised image $\bm{x}$\\
\bottomrule
\end{tabular}
\end{table}

Moreover, extensive modified sparse representation based image denoising algorithms have been proposed. For example, Dabov et al. \cite{dabov2007image} proposed an enhanced sparse representation with a block-matching 3-D (BM3D) transform-domain filter based on self-similarities and an enhanced sparse representation by clustering similar 2-D image patches into 3-D data spaces and an iterative collaborative filtering procedure for image denoising. Mariral et al. \cite{mairal2008sparse} proposed the use of extending the KSVD-based grayscale algorithm and a generalized weighted average algorithm for color image denoising. Protter and Elad \cite{protter2009image} extended the techniques of sparse and redundant representations for image sequence denoising by exploiting spatio-temporal atoms, dictionary propagation over time and dictionary learning. Dong et al. \cite{dong2011sparsity} designed a clustering based sparse representation algorithm, which was formulated by a double-header sparse optimization problem built upon dictionary learning and structural clustering. Recently, Jiang et al. \cite{jiang2014mixed} proposed a variational encoding framework with a weighted sparse nonlocal constraint, which was constructed by integrating image sparsity prior and nonlocal self-similarity prior into a unified regularization term to overcome the mixed noise removal problem. Gu et al. \cite{gu2014weighted} studied a weighted nuclear norm minimization (WNNM) method with $F$-norm fidelity under different weighting rules optimized by non-local self-similarity for image denoising. Ji et al. \cite{ji2010robust} proposed a patch-based video denoising algorithm by stacking similar patches in both spatial and temporal domain to formulate a low-rank matrix problem with the nuclear norm. Cheng et al. \cite{cheng2013image} proposed an impressive image denoising method based on an extension of the KSVD algorithm via group sparse representation.

The primary purpose of image restoration is to recover the original image from the degraded or blurred image. The sparse representation theory has been extensively applied to image restoration. For example, Bioucas-Dias and Figueirdo \cite{bioucas2007new} introduced a two-step iterative shrinkage/thresholding (TwIST) algorithm for image restoration, which is more efficient and can be viewed as an extension of the IST method. Mairal et al. \cite{mairal2008learning} presented a multiscale sparse image representation framework based on the KSVD dictionary learning algorithm and shift-invariant sparsity prior knowledge for restoration of color images and video image sequence. Recently, Mairal et al. \cite{mairal2009non} proposed a learned simultaneous sparse coding (LSSC) model, which integrated sparse dictionary learning and nonlocal self-similarities of natural images into a unified framework for image restoration. Zoran and Weiss \cite{zoran2011learning} proposed an expected patch log likelihood (EPLL) optimization model, which restored the image from patch to the whole image based on the learned prior knowledge of any patch acquired by Maximum A-Posteriori estimation instead of using simple patch averaging. Bao et al. \cite{bao2013fast} proposed a fast orthogonal dictionary learning algorithm, in which a sparse image representation based orthogonal dictionary was learned in image restoration. Zhang et al.  \cite{zhang2014group} proposed a group-based sparse representation, which combined characteristics from local sparsity and nonlocal self-similarity of natural images to the domain of the group. Dong et al. \cite{dong2011centralized, dong2013nonlocally} proposed a centralized sparse representation (CSR) model, which combined the local and nonlocal sparsity and redundancy properties for variational problem optimization by introducing a concept of sparse coding noise term.

Here we mainly introduce a recently proposed simple but effective image restoration algorithm CSR model \cite{dong2011centralized}. For a degraded image $\bm{y}$, the problem of image restoration can be formulated as

\begin{equation}\label{8-41}
\bm{y}=H\bm{x}+\bm{v}
\end{equation}
where $H$ is a degradation operator, $\bm{x}$ is the original high-quality image and $\bm{v}$ is the Gaussian white noise. Suppose that the following two sparse optimization problems are satisfied
\begin{equation}\label{8-42}
{\bm\alpha}_x = arg \min \|{\bm\alpha}\|_1~~s.t.~~\|\bm{x}-D{\bm\alpha}\|_2^2 \leq \varepsilon
\end{equation}
\begin{equation}\label{8-43}
{\bm\alpha}_y = arg \min \|{\bm\alpha}\|_1~~s.t.~~\|\bm{x}-HD{\bm\alpha}\|_2^2 \leq \varepsilon
\end{equation}
where $y$ and $x$ respectively denote the degraded image and original high-quality image, and $\varepsilon$ is a small constant. A new concept called sparse coding noise (SCN) is defined
\begin{equation}\label{8-44}
\bm{v}_{\alpha} = {\bm\alpha}_y - {\bm\alpha}_x
\end{equation}
Given a dictionary $D$, minimizing SCN can make the image better reconstructed and improve the quality of the image restoration because $\bm{x}^* = \hat{\bm{x}} - \tilde{\bm{x}} = D{\bm\alpha}_y -D{\bm\alpha}_x =Dv_{\alpha}$. Thus, the objective function is reformulated as
\begin{equation}\label{8-45}
{\bm\alpha}_y = arg \min_{{\bm\alpha}} \|y-HD{\bm\alpha}\|_2^2 + \lambda \|{\bm\alpha}\|_1 + \mu \|{\bm\alpha}-{\bm\alpha}_x\|_1
\end{equation}
where $\lambda$ and $\mu$ are both constants. However, the value of ${\bm\alpha}_x$ is difficult to directly evaluate. Because many nonlocal similar patches are associated with the given image patch $i$, clustering these patches via block matching is advisable and the sparse code of searching similar patch $l$ to patch $i$ in cluster $\Omega_i$, denoted by ${\bm\alpha}_{il}$, can be computed. Moreover, the unbiased estimation of ${\bm\alpha}_x$, denoted by $E[{\bm\alpha}_x]$, empirically can be approximate to ${\bm\alpha}_x$ under some prior knowledge \cite{dong2011centralized}, and then SCN algorithm employs the nonlocal means estimation method \cite{buades2005non} to evaluate the unbiased estimation of ${\bm\alpha}_x$, that is, using the weighted average of all ${\bm\alpha}_{il}$ to approach $E[{\bm\alpha}_x]$, i.e.

\begin{equation}\label{8-46}
\bm{\theta}_i = \sum_{l\in \Omega_i} w_{il} {\bm\alpha}_{il}
\end{equation}
where $w_{il} = \exp \left ( -\|\bm{x}_i - \bm{x}_{il}\|_2^2 / h \right)/N$, $\bm{x}_i =D {\bm\alpha}_i$, $\bm{x}_{il} = D{\bm\alpha}_{il}$, $N$ is a normalization parameter and $h$ is a constant. Thus, the objective function \ref{8-45} can be rewritten as
\begin{equation}\label{8-47}
{\bm\alpha}_y = arg \min_{{\bm\alpha}} \|\bm{y}-HD{\bm\alpha}\|_2^2 + \lambda \|{\bm\alpha}\|_1 + \mu \sum_{i=1}^M \|{\bm\alpha}_i-\bm{\theta}_i\|_1
\end{equation}
where $M$ is the number of the separated patches. In the $j$-th iteration, the solution of problem \ref{8-47} is iteratively performed by
\begin{equation}\label{8-48}
{\bm\alpha}_y^{j+1} = arg \min_{{\bm\alpha}} \|\bm{y}-HD{\bm\alpha}\|_2^2 + \lambda \|{\bm\alpha}\|_1 + \mu \sum_{i=1}^M \|{\bm\alpha}_i-\bm{\theta}_i^j\|_1
\end{equation}
It is obvious that problem \ref{8-47} can be optimized by the augmented Lagrange multiplier method \cite{nedic2003convex} or the iterative shrinkage algorithm in \cite{daubechies2004iterative}. According to the maximum average posterior principle and the distribution of the sparse coefficients, the regularization parameter $\lambda$ and constant $\mu$ can be adaptively determined by $\lambda=\frac{2\sqrt{2}\rho^2}{\sigma_i}$ and $\mu=\frac{2\sqrt{2}\rho^2}{\eta_i}$, where $\rho$, $\sigma_i$ and $\eta_i$ are the standard deviations of the additive Gaussian noise, ${\bm\alpha}_i$ and the SCN signal, respectively. Moreover, in the process of image patches clustering for each given image patch, a local PCA dictionary is learned and employed to code each patch within its corresponding cluster. The main procedures of the CSR algorithm are summarized in Algorithm 16 and readers may refer to literature \cite{dong2011centralized} for more details.

\begin{table}[htbp]
\centering
\begin{tabular}{m{85mm}}
\toprule
\textbf{Algorithm 16.} Centralized sparse representation for image restoration\\
\midrule
\textbf{Initialization:} Set $\bm{x}=\bm{y}$,  initialize regularization parameter $\lambda$ and $\mu$, the number of training iteration $T$, $t=0$, $\bm{\theta}^0=0$.\\
Step 1: Partition the degraded image into $M$ overlapped patches.\\
While $t\leq T$ do \\
Step 2:~For each image patch, update the corresponding dictionary for each cluster via k-means and PCA.\\
Step 3:~Update the regularization parameters $\lambda$ and $\mu$ by using\\~~~~~~~~~~~~~~~~~~~~~~$\lambda=\frac{2\sqrt{2}\rho^2}{\sigma_t}$ and $\mu=\frac{2\sqrt{2}\rho^2}{\eta_t}$.\\
Step 4:~Compute the nonlocal means estimation of the unbiased estimation of ${\bm\alpha}_x$, i.e. $\bm{\theta}_i^{t+1}$, by using Eq. (\ref{8-46}) for each image patch.\\
Step 5:~For a given $\bm{\theta}_i^{t+1}$, compute the sparse representation solution, i.e. ${\bm\alpha}_y^{t+1}$, in problem (\ref{8-48}) by using the extended iterative shrinkage algorithm in literature \cite{daubechies2004iterative}.\\
Step 6:~$t=t+1$\\
End While \\
\textbf{Output:}~Restored image $\bm{x}=D{\bm\alpha}_y^{t+1}$\\
\bottomrule
\end{tabular}
\end{table}

\subsection{Sparse representation in image classification and visual tracking}
\noindent In addition to these effective applications in image processing, several other fields for sparse representation have been proposed and extensively studied in image classification and visual tracking. Since Wright et al. \cite{wright2009robust} proposed to employ sparse representation to perform robust face recognition, more and more researchers have been applying the sparse representation theory to the fields of computer vision and pattern recognition, especially in image classification and object tracking. Experimental results have suggested that the sparse representation based classification method can somewhat overcome the challenging issues from illumination changes, random pixel corruption, large block occlusion or disguise.

As face recognition is a representative component of pattern recognition and computer vision applications, the applications of sparse representation in face recognition can sufficiently reveal the potential nature of sparse representation. The most representative sparse representation for face recognition has been presented in literature [18] and the general scheme of sparse representation based classification method is summarized in Algorithm 17. Suppose that there are $n$ training samples, $X=[x_1,x_2,\cdots,x_n]$ from $c$ classes. Let $X_i$ denote the samples from the $i$-th class and the testing sample is $y$.

\begin{table}[htbp]
\centering
\begin{tabular}{m{85mm}}
\toprule
\textbf{Algorithm 17.} The scheme of sparse representation based classification method\\
\midrule
Step 1:~Normalize all the samples to have unit $l_2$-norm.\\
Step 2:~Exploit the linear combination of all the training samples to represent the test sample and the following $l_1$-norm minimization problem is satisfied\\
~~~~~~~~~~~~~~ ${\bm\alpha}^* = arg \min \|{\bm\alpha}\|_1~s.t.~\|\bm{y}-X{\bm\alpha}\|_2^2 \leq \varepsilon$.\\
Step 3:~Compute the representation residual for each class\\
~~~~~~~~~~~~~~ $\bm{r}_i = \|\bm{y}-X_i {\bm\alpha}_i^*\|_2^2$\\
where ${\bm\alpha}_i^*$ here denotes the representation coefficients vector associated with the $i$-th class.\\
Step 4:~Output the identity of the test sample $y$ by judging\\
~~~~~~~~~~~~~~~$label(\bm{y}) = arg\min_i (\bm{r}_i)$.\\
\bottomrule
\end{tabular}
\end{table}

Numerous sparse representation based classification methods have been proposed to improve the robustness, effectiveness and efficiency of face recognition. For example, Xu et al. \cite{xu2011two} proposed a two-phase sparse representation based classification method, which exploited the $l_2$-norm regularization rather than the $l_1$-norm regularization to perform a coarse to fine sparse representation based classification, which was very efficient in comparison with the conventional $l_1$-norm regularization based sparse representation. Deng et al. \cite{deng2012extended} proposed an extended sparse representation method (ESRM) for improving the robustness of SRC by eliminating the variations in face recognition, such as disguise, occlusion, expression and illumination. Deng et al. \cite{deng2013defense} also proposed a framework of superposed sparse representation based classification, which emphasized the prototype and variation components from uncontrolled images. He et al. \cite{he2011maximum} proposed utilizing the maximum correntropy criterion named CESR embedding non-negative constraint and half-quadratic optimization to present a robust face recognition algorithm. Yang et al. \cite{yang2011rsc} developed a new robust sparse coding (RSC) algorithm, which first obtained a sparsity-constrained regression model based on maximum likelihood estimation and exploited an iteratively reweighted regularized robust coding algorithm to solve the pre-proposed model. Some other sparse representation based image classification methods also have been developed. For example, Yang et al. \cite{yang2009linear} introduced an extension of the spatial pyramid matching (SPM) algorithm called ScSPM, which incorporated SIFT sparse representation into the spatial pyramid matching algorithm. Subsequently, Gao et al. \cite{gao2010kernel} developed a kernel sparse representation with the SPM algorithm called KSRSPM, and then proposed another version of an improvement of the SPM called LScSPM \cite{gao2010local}, which integrated the Laplacian matrix with local features into the objective function of the sparse representation method. Kulkarni and Li \cite{kulkarni2011discriminative} proposed a discriminative affine sparse codes method (DASC) on a learned affine-invariant feature dictionary from input images and exploited the AdaBoost-based classifier to perform image classification. Zhang et al. \cite{zhang2011image} proposed integrating the non-negative sparse coding, low-rank and sparse matrix decomposition (LR-Sc$^+$SPM) method, which exploited non-negative sparse coding and SPM for achieving local features representation and employed low-rank and sparse matrix decomposition for sparse representation, for image classification. Recently, Zhang et al. \cite{zhang2013low} presented a low-rank sparse representation (LRSR) learning method, which preserved the sparsity and spatial consistency in each procedure of feature representation and jointly exploited local features from the same spatial proximal regions for image classification. Zhang et al. \cite{zhang2013learning} developed a structured low-rank sparse representation (SLRSR) method for image classification, which constructed a discriminative dictionary in training terms and exploited low-rank matrix reconstruction for obtaining discriminative representations. Tao et al. \cite{tao2013rank} proposed a novel dimension reduction method based on the framework of rank preserving sparse learning, and then exploited the projected samples to make effective Kinect-based scene classification. Zhang et al. \cite{zhang2013discriminative} proposed a discriminative tensor sparse coding (RTSC) method for robust image classification. Recently, low-rank based sparse representation became a popular topic such as non-negative low-rank and sparse graph \cite{zhuang2014constructing}. Some sparse representation methods in face recognition can be found in a review \cite{yang2010fast} and other more image classification methods can be found in a more recent review \cite{rigamonti2014relevance}.

Mei et al. employed the idea of sparse representation to visual tracking \cite{mei2009robust} and vehicle classification \cite{mei2011robust}, which introduced nonnegative sparse constraints and dynamic template updating strategy. It, in the context of the particle filter framework, exploited the sparse technique to guarantee that each target candidate could be sparsely represented using the linear combinations of fewest targets and particle templates. It also demonstrated that sparse representation can be propagated to address object tracking problems. Extensive sparse representation methods have been proposed to address the visual tracking problem. In order to design an accelerated algorithm for $l_1$ tracker, Li et al. \cite{li2011real} proposed two real-time compressive sensing visual tracking algorithms based on sparse representation, which adopted dimension reduction and the OMP algorithm to improve the efficiency of recovery procedure in tracking, and also developed a modified version of fusing background templates into the tracking procedure for robust object tracking. Zhang et al. \cite{zhang2010robust} directly treated object tracking as a pattern recognition problem by regarding all the targets as training samples, and then employed the sparse representation classification method to do effective object tracking. Zhang et al. \cite{zhang2012multitask} employed the concept of sparse representation based on a particle filter framework to construct a multi-task sparse learning method denoted as multi-task tracking for robust visual tracking. Additionally, because of the discriminative sparse representation between the target and the background, Jia et al. \cite{jia2012visual} conceived a structural local sparse appearance model for robust object tracking by integrating the partial and spatial information from the target based on an alignment-pooling algorithm. Liu et al. \cite{liu2010robust} proposed constructing a two-stage sparse optimization based online visual tracking method, which jointly minimized the objective reconstruction error and maximized the discriminative capability by choosing distinguishable features. Liu et al. \cite{liu2011robust} introduced a local sparse appearance model (SPT) with a static sparse dictionary learned from $k$-selection and dynamic updated basis distribution to eliminate potential drifting problems in the process of visual tracking. Bao et al. \cite{bao2012real} developed a fast real time $l_1$-tracker called the APG-$l_1$tracker, which exploited the accelerated proximal gradient algorithm to improve the $l_1$-tracker solver in \cite{mei2009robust}. Zhong et al.  \cite{zhong2012robust} addressed the object tracking problem by developing a sparsity-based collaborative model, which combined a sparsity-based classifier learned from holistic templates and a sparsity-based template model generated from local representations. Zhang et al. \cite{zhang2014fast} proposed to formulate a sparse feature measurement matrix based on an appearance model by exploiting non-adaptive random projections, and employed a coarse-to-fine strategy to accelerate the computational efficiency of tracking task. Lu et al. \cite{lu2013robust} proposed to employ both non-local self-similarity and sparse representation to develop a non-local self-similarity regularized sparse representation method based on geometrical structure information of the target template data set. Wang et al. \cite{ wang2013online} proposed a sparse representation based online two-stage tracking algorithm, which learned a linear classifier based on local sparse representation on favorable image patches. More detailed visual tracking algorithms can be found in the recent reviews \cite{zhang2013sparse, smeulders2013visual}.

\begin{figure*}[htbp]
\centering
\subfloat[Parameter selection on ORL]
{
    \label{para_a}
    \includegraphics[width=3.5in]{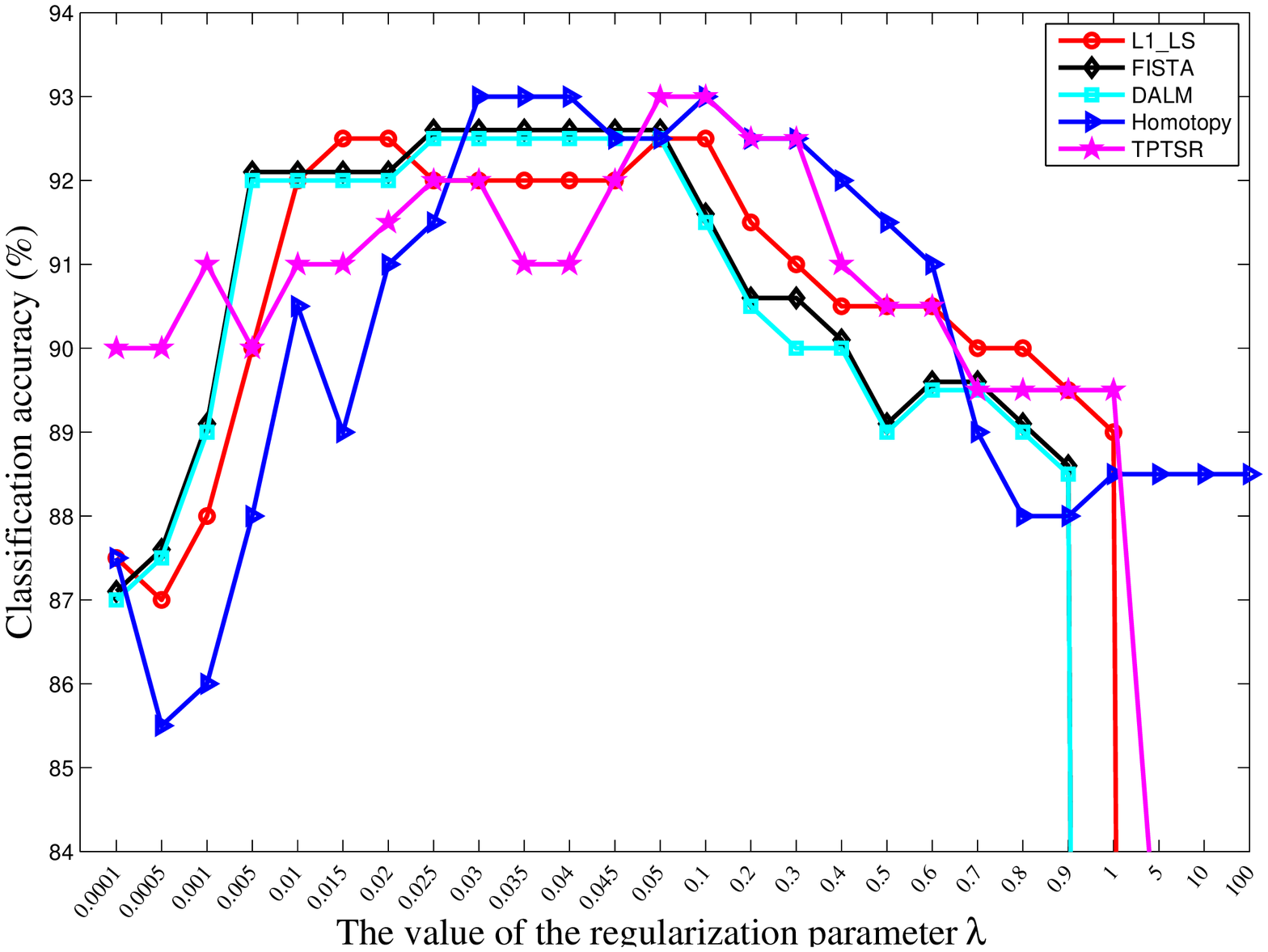}
}
\subfloat[Parameter selection on LFW]
{
    \label{para_b}
    \includegraphics[width=3.48in]{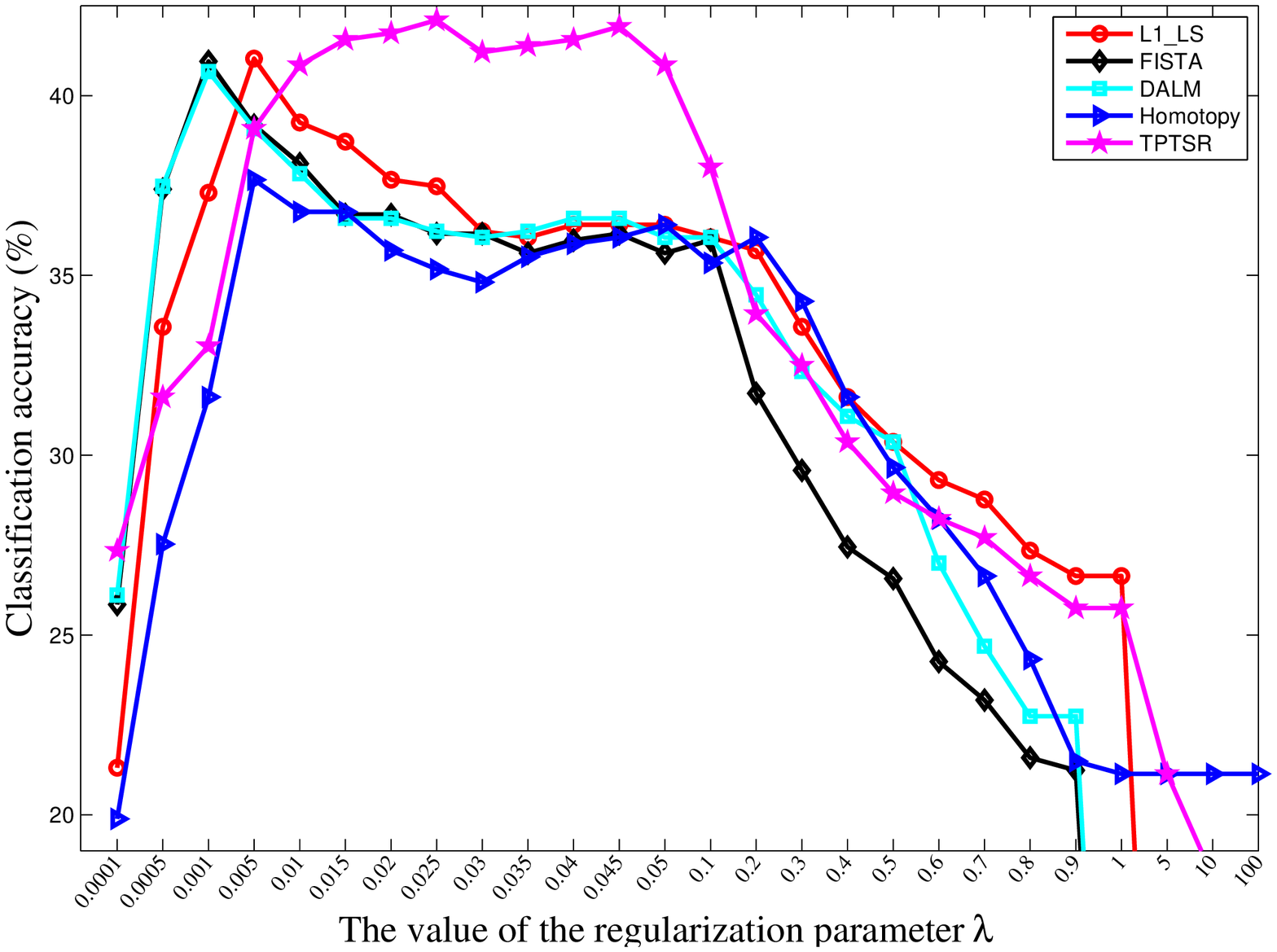}
}
\\
\subfloat[Parameter selection on Coil\_20]
{
    \label{para_c}
    \includegraphics[width=3.48in]{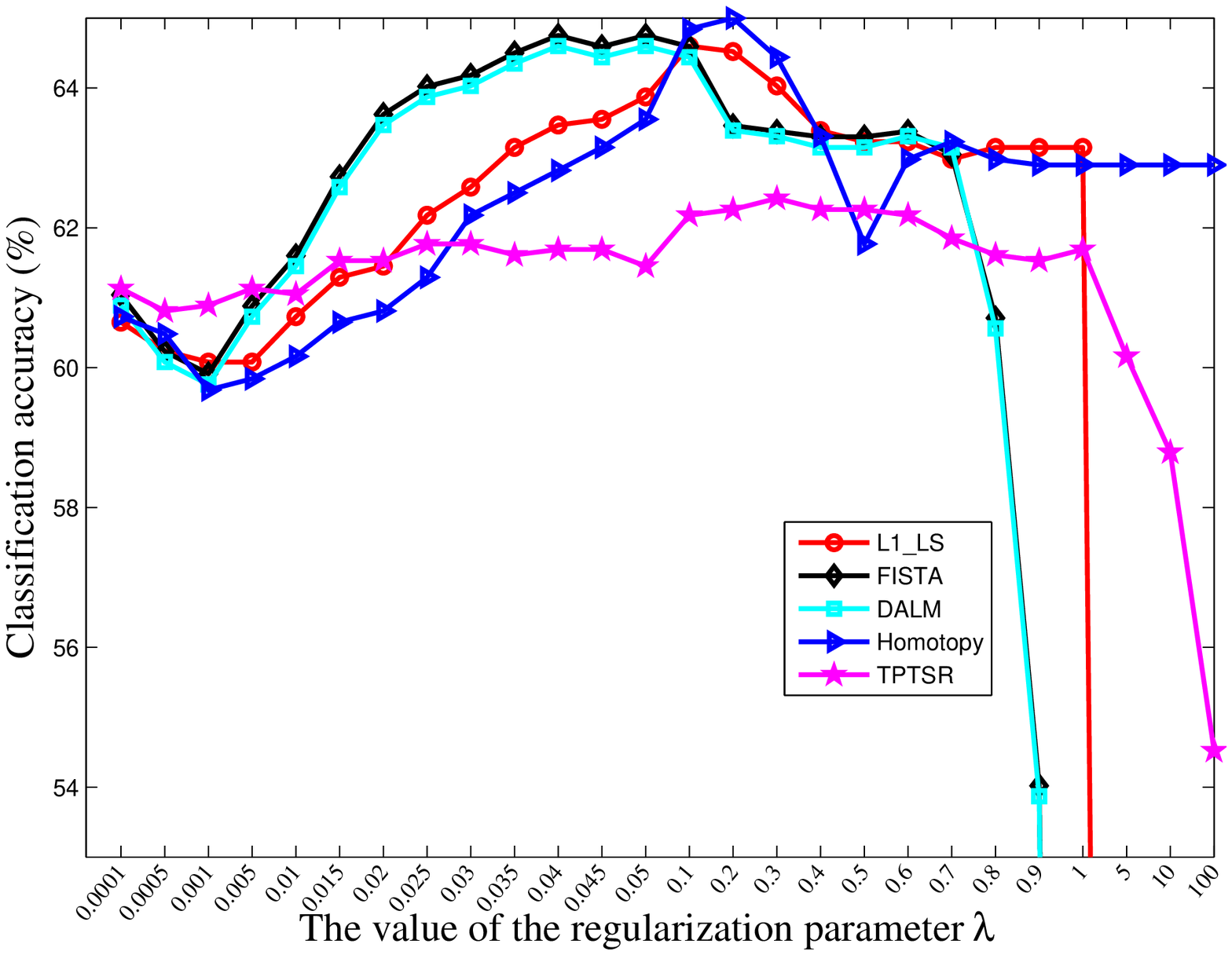}
}
\subfloat[Parameter selection on 15\_scene]
{
    \label{para_d}
    \includegraphics[width=3.48in]{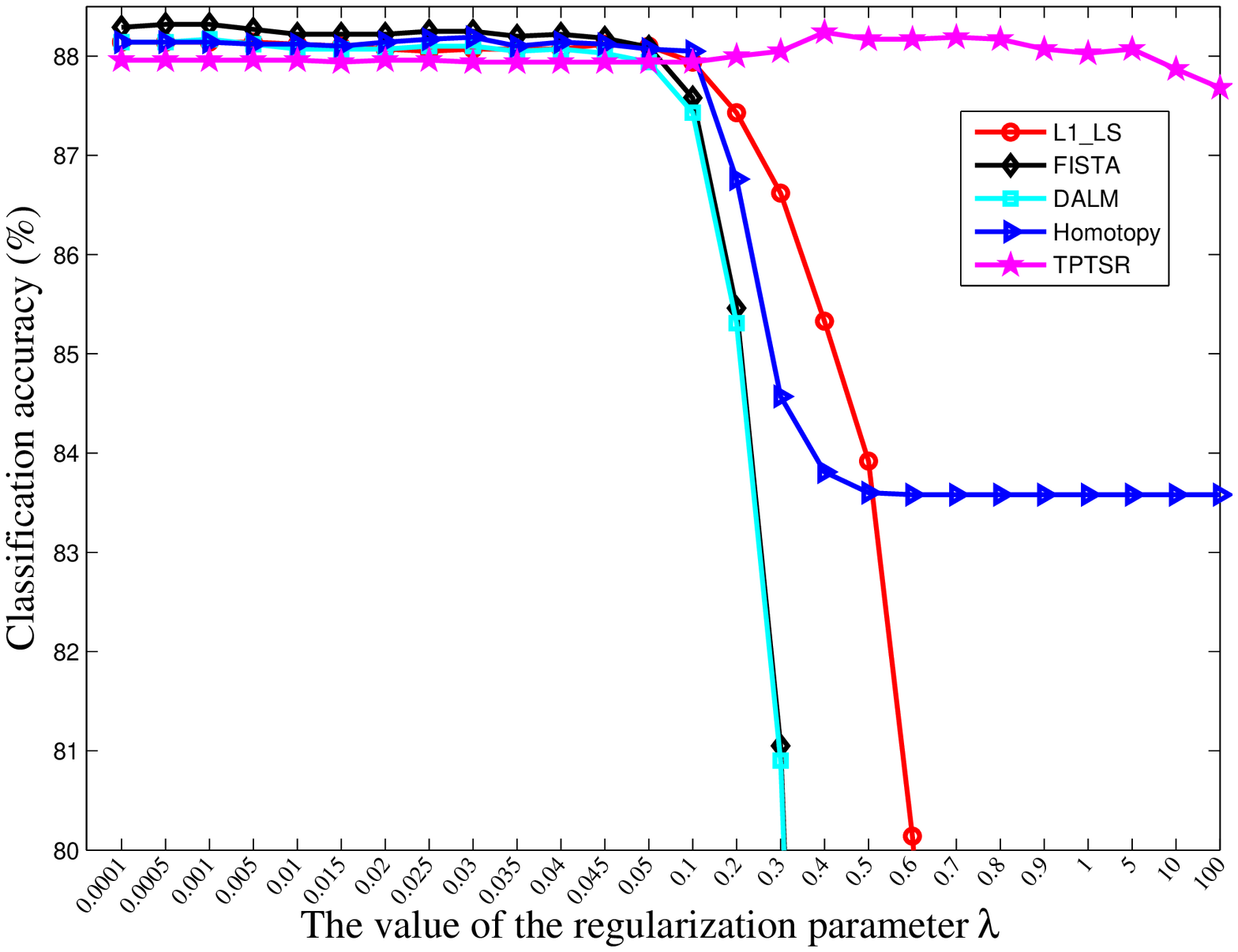}
}
\caption{Classification accuracies of using different sparse representation based classification methods versus varying values of the regularization parameter $\lambda$ on the (a) ORL (b) LFW (c) Coil20 and (d) Fifteen scene datasets. }\label{fig_para}
\end{figure*}

\section{Experimental evaluation}\label{sec_ex}
In this section, we take the object categorization problem as an example to evaluate the performance of different sparse representation based classification methods. We analyze and compare the performance of sparse representation with the most typical algorithms: OMP \cite{tropp2007signal}, $l_1\_l_s$ \cite{koh2007interior}, PALM \cite{yang2013fast}, FISTA \cite{beck2009fast}, DALM \cite{yang2013fast}, homotopy \cite{asif2008primal} and TPTSR \cite{xu2011two}.

Plenties of data sets have been collected for object categorization, especially for image classification. Several image data sets are used in our experimental evaluations.

\emph{ORL}: The ORL database includes 400 face images taken from 40 subjects each providing 10 face images \cite{samaria1994param}. For some subjects, the images were taken at different times, with varying lighting, facial expressions, and facial details. All the images were taken against a dark homogeneous background with the subjects in an upright, frontal position (with tolerance for some side movement). Each image was resized to a 56$\times$46 image matrix by using the down-sampling algorithm.

\emph{LFW face dataset}: The Labeled Faces in the Wild (LFW) face database is designed for the study of unconstrained identity verification and face recognition \cite{huang2007labeled}. It contains more than 13,000 images of faces collected from the web under the unconstrained conditions. Each face has been labeled with the name of the people pictured. 1680 of the people pictured have two or more distinct photos in the database. In our experiments, we chose 1251 images from 86 peoples and each subject has 10-20 images \cite{wang2012sparse}. Each image was manually cropped and was resized to 32$\times$32 pixels.

\emph{Extended YaleB face dataset}: The extended YaleB database contains 2432 front face images of 38 individuals and each subject having around 64 near frontal images under different illuminations \cite{georghiades2001few}. The main challenge of this database is to overcome varying illumination conditions and expressions. The facial portion of each original image was cropped to a 192$\times$168 image. All images in this data set for our experiments simply  resized these face images to 32$\times$32 pixels.

\emph{COIL20 dataset:} Columbia Object Image Library (COIL-20) database consists of 1,440 size normalized gray-scale images of 20 objects \cite{nene1996columbia}. Different object images are captured at every angle in a 360 rotation. Images of the objects were taken from varying angles at pose intervals of five degrees and each object has 72 images.

\emph{Fifteen scene dataset:} This dataset contains 4485 images under 15 natural scene categories presented in literature \cite{lazebnik2006beyond} and each category includes 210 to 410 images. The 15 scenes categories are office, kitchen, living room, bedroom, store, industrial, tall building, inside cite, street, highway, coast, open country, mountain, forest and suburb. A wide range of outdoor and indoor scenes are included in this dataset. The average image size is around 250$\times$300 pixels and the spatial pyramid matching features are used in our experiments.

\begin{table*}
\begin{center}
\begin{tabular}{lcccccccc}
\hline
Data set ($\#Tr$) &OMP&$l_1\_ls$& PALM & FISTA& DALM & Homotopy& TPTSR\\
\hline  
ORL(1) & 64.94$\pm$2.374 & 68.50$\pm$2.021  & 68.36$\pm$1.957 & 70.67$\pm$2.429 & 70.22$\pm$2.805 & 66.53$\pm$1.264 & \textbf{71.56$\pm$3.032} \\
ORL(2) & 80.59$\pm$2.256 & \textbf{84.84$\pm$2.857}  & 80.66$\pm$2.391 & 84.72$\pm$3.242 & 84.38$\pm$2.210 & 83.88$\pm$2.115 & 83.38$\pm$2.019\\
ORL(3) & 89.00$\pm$1.291 & 89.71$\pm$1.313 & 86.82$\pm$1.959 & 90.00$\pm$3.141 & 90.36$\pm$1.829 & 89.32$\pm$1.832 & \textbf{90.71$\pm$1.725}\\
ORL(4) & 91.79$\pm$1.713 & \textbf{94.83$\pm$1.024} & 88.63$\pm$2.430 & 94.13$\pm$1.310 & 94.71$\pm$1.289 & 94.38$\pm$1.115 & 94.58$\pm$1.584\\
ORL(5) & 93.75$\pm$2.125 & \textbf{95.90$\pm$1.150} & 92.05$\pm$1.039 & 95.60$\pm$1.761 & 95.50$\pm$1.269 & 95.60$\pm$1.430 & 95.75$\pm$1.439\\
ORL(6) & 95.69$\pm$1.120 & 97.25$\pm$1.222 & 92.06$\pm$1.319 & 96.69$\pm$1.319 & 96.56$\pm$1.724 & \textbf{97.31$\pm$1.143} & 95.81$\pm$1.642\\
Average Time(5) &0.0038s & 0.1363s  & 7.4448s & 0.9046s & 0.8013s& 0.0100s & \textbf{0.0017s}\\
\hline
LFW(3) & 22.22$\pm$1.369 & \textbf{28.24$\pm$0.667} & 15.16$\pm$1.202 & 26.55$\pm$0.767 & 25.46$\pm$0.705 & 26.12$\pm$0.831 & 27.32$\pm$1.095\\
LFW(5) & 27.83$\pm$1.011 & \textbf{35.58$\pm$1.489} & 12.89$\pm$1.286 & 34.13$\pm$0.459 & 33.90$\pm$1.181 & 33.95$\pm$1.680 & 35.43$\pm$1.409\\
LFW(7) & 32.76$\pm$2.318 & 40.17$\pm$2.061 & 11.63$\pm$0.937 & 39.86$\pm$1.226 & 38.40$\pm$1.890 & 38.04$\pm$1.251 & \textbf{40.92$\pm$1.201} \\
LFW(9) & 35.14$\pm$1.136 & \textbf{44.93$\pm$1.123} & 7.84$\pm$1.278 &  43.86$\pm$1.492  &  43.56$\pm$1.393  &  42.29$\pm$2.721  &  44.72$\pm$1.793 \\
Average Time(7) & \textbf{0.0140s} &  0.6825s & 33.2695s & 3.0832s  &  3.9906s & 0.2372s & 0.0424s\\
\hline
Extended YaleB(3)& 44.20$\pm$2.246 & 63.10$\pm$2.341 & 63.73$\pm$2.073 & 62.84$\pm$2.623 & \textbf{63.76$\pm$2.430} & 64.22$\pm$2.525 & 56.23$\pm$2.153 \\
Extended YaleB(6) & 72.48$\pm$2.330 & 81.97$\pm$0.850 & 81.93$\pm$0.930 & \textbf{82.25$\pm$0.734} & 81.74$\pm$1.082 & 81.64$\pm$1.159 & 78.53$\pm$1.731 \\
Extended YaleB(9) & 83.42$\pm$0.945 & 88.90$\pm$0.544 & 88.50$\pm$1.096 & \textbf{89.31$\pm$0.829} & 89.26$\pm$0.781 & 89.12$\pm$0.779 & 86.49$\pm$1.165 \\
Extended YaleB(12) & 88.23$\pm$0.961 & \textbf{92.49$\pm$0.622} & 91.07$\pm$0.725 & 92.03$\pm$1.248 & 91.85$\pm$0.710 & 92.03$\pm$0.767 & 91.30$\pm$0.741\\
Extended YaleB(15) & 91.97$\pm$0.963 & 94.22$\pm$0.719 & 93.19$\pm$0.642  &  \textbf{94.50$\pm$0.824}  &  93.07$\pm$0.538  &  93.67$\pm$0.860  &  93.38$\pm$0.785 \\
Average Time(12) &   \textbf{0.0116s} &   3.2652s & 17.4516s&  1.5739s &  1.9384s &   0.5495s &  0.0198s \\
\hline
COIL20(3)& 75.90$\pm$1.656 & 77.62$\pm$2.347 & 70.26$\pm$2.646 & 75.80$\pm$2.056 & 76.67$\pm$2.606 & \textbf{78.46$\pm$2.603} & 78.16$\pm$2.197\\
COIL20(5) & 83.00$\pm$1.892 & 82.63$\pm$1.701 & 79.55$\pm$1.153  & 84.09$\pm$2.003 & 84.38$\pm$1.319 & \textbf{84.58$\pm$1.487} & 83.69$\pm$1.804 \\
COIL20(7) & 87.26$\pm$1.289 & 88.22$\pm$1.304 & 82.88$\pm$1.445 & 88.89$\pm$1.598 & 89.00$\pm$1.000 & \textbf{89.36$\pm$1.147} & 87.75$\pm$1.451 \\
COIL20(9) & 89.56$\pm$1.763 & 90.97$\pm$1.595 & 84.94$\pm$1.563 & 90.16$\pm$1.366 & \textbf{91.82$\pm$1.555} & 91.44$\pm$1.198 & 89.41$\pm$2.167\\
COIL20(11) & 91.70$\pm$0.739 & 92.98$\pm$1.404 & 87.16$\pm$1.184 & 93.43$\pm$1.543 & \textbf{93.46$\pm$1.327} & 93.55$\pm$1.205 & 92.71$\pm$1.618 \\
COIL20(13) & 92.49$\pm$1.146 & 94.29$\pm$0.986 & 88.36$\pm$1.283 & 94.50$\pm$0.850 & 93.92$\pm$1.102 & \textbf{94.93$\pm$0.788} & 92.72$\pm$1.481\\
Average Time(13) & \textbf{0.0038s} & 0.0797s & 7.5191s&  0.7812s &  0.7762s & 0.0159s & 0.0053s \\
\hline
Fifteen scene(3)& 85.40$\pm$1.388 & \textbf{86.83$\pm$1.082} & 86.15$\pm$1.504 & 86.48$\pm$1.542 & 85.89$\pm$1.624 & 86.15$\pm$1.073 & 86.62$\pm$1.405 \\
Fifteen scene(6) & 89.14$\pm$1.033 & 90.34$\pm$0.685 & 89.97$\pm$0.601 & 90.82$\pm$0.921 & 90.12$\pm$0.998 & 89.65$\pm$0.888 & \textbf{90.83$\pm$0.737} \\
Fifteen scene(9) & 83.42$\pm$0.945 & 88.90$\pm$0.544 & 88.50$\pm$1.096 & 89.31$\pm$0.829 & 89.26$\pm$0.781 & 89.12$\pm$0.779 & \textbf{90.64$\pm$0.940} \\
Fifteen scene(12) & 91.67$\pm$0.970 & 92.06$\pm$0.536 & \textbf{92.76$\pm$0.905} & 92.22$\pm$0.720 & 92.45$\pm$0.860 & 92.35$\pm$0.706 & 92.33$\pm$0.563\\
Fifteen scene(15) & 93.32$\pm$0.609 & 93.37$\pm$0.506 & 93.63$\pm$0.510 & 93.63$\pm$0.787 & 93.53$\pm$0.829 & \textbf{93.84$\pm$0.586} & 93.80$\pm$0.461\\
Fifteen scene(18) & 93.61$\pm$0.334  & 94.31$\pm$0.551  & 94.67$\pm$0.678 & 94.28$\pm$0.396  & 94.16$\pm$0.344  & 94.16$\pm$0.642 & \textbf{94.78$\pm$0.494}  \\
Average Time(18) & \textbf{0.0037s} & 0.0759s & 0.9124s & 0.8119s & 0.8500s & 0.1811s & 0.0122s\\
\hline
\end{tabular}
\end{center}
\caption{Classification accuracies (mean classification error rates $\pm$ standard deviation \%) of different sparse representation algorithms with different numbers of training samples. The bold numbers are the lowest error rates and the least time cost of different algorithms.} \label{Table_ex}
\end{table*}

\subsection{Parameter selection}\label{subsec9-1}
\noindent Parameter selection, especially selection of the regularization parameter $\lambda$ in different minimization problems, plays an important role in sparse representation. In order to make fair comparisons with different sparse representation algorithms, performing the optimal parameter selection for different sparse representation algorithms on different datasets is advisable and indispensable. In this subsection, we perform extensive experiments for selecting the best value of the regularization parameter $\lambda$ with a wide range of options. Specifically, we implement the $l_1\_l_s$, FISTA, DALM, homotopy and TPTSR algorithms on different databases to analyze the importance of the regularization parameter. Fig. \ref{fig_para} summarizes the classification accuracies of exploiting different sparse representation based classification methods with varying values of regularization parameter $\lambda$ on the two face datasets, i.e. ORL and LFW face datasets, and two object datasets, i.e. COIL20 and Fifteen scene datasets. On the ORL and LFW face datasets, we respectively selected the first five and eight face images of each subject as training samples and the rest of image samples for testing. As for the experiments on the COIL20 and fifteen scene datasets, we respectively treated the first ten images of each subject in both datasets as training samples and used all the remaining images as test samples. Moreover, from Fig. \ref{fig_para}, one can see that the value of regularization parameter $\lambda$ can significantly dominate the classification results, and the values of $\lambda$ for achieving the best classification results on different datasets are distinctly different. An interesting scenario is that the performance of the TPTSR algorithm is almost not influenced by the variation of regularization parameter $\lambda$ in the experiments on fifteen scene dataset, as shown in Fig. \ref{fig_para}(d). However, the best classification accuracy can be always obtained within the range of 0.0001 to 1. Thus, the value of the regularization parameter is set within the range from 0.0001 to 1.

\subsection{Experimental results}\label{subsec9-2}
\noindent In order to test the performance of different kinds of sparse representation methods, an empirical study of experimental results is conducted in this subsection and seven typical sparse representation based classification methods are selected for performance evaluation followed with extensive experimental results. For all datasets, following most previous published work, we randomly choose several samples of every class as training samples and used the rest as test samples and the experiments are repeated 10 times with the optimal parameter obtained using the cross validation approach. The gray-level features of all images in these data sets are used to perform classification. For the sake of computational efficiency, principle component analysis algorithm is used as a preprocessing step to preserve 98\% energy of all the data sets. The classification results and computational time have been summarized in Table \ref{Table_ex}. From the experimental results on different databases, we can conclude that there still does not exist one extraordinary algorithm that can achieve the best classification accuracy on all databases. However, some algorithms are noteworthy to be paid much more attention. For example, the $l_1\_l_s$ algorithm in most cases can achieve better classification results than the other algorithms on the ORL database, and when the number of training samples of each class is five, the $l_1\_l_s$ algorithm can obtain the highest classification result of $95.90$\%. The TPTSR algorithm is very computationally efficient in comparison with other sparse representation with $l_1$-norm minimization algorithms and the classification accuracies obtained by the TPTSR algorithm are very similar and sometimes even better than the other sparse representation based classification algorithms.

The computational time is another indicator for measuring the performance of one specific algorithm. As shown in Table \ref{Table_ex}, the average computational time of each algorithm is shown at the bottom of the table for one specific number of training samples. Note that the computational time of OMP and TPTSR algorithms are drastically lower than that of other sparse representation with $l_1$-norm minimization algorithms. This is mainly because the sparse representation with $l_1$-norm minimization algorithms always iteratively solve the $l_1$-norm minimization problem. However, the OMP and TPTSR algorithms both exploit the fast and efficient least squares technique, which guarantees that the computational time is significantly less than other $l_1$-norm based sparse representation algorithms.

\subsection{Discussion}
\noindent Lots of sparse representation methods have been available in past decades and this paper introduces various sparse representation methods from some viewpoints, including their motivations, mathematical representations and the main algorithms. Based on the experimental results summarized in Section \ref{sec_ex}, we have the following observations.

First, a challenging task of choosing a suitable regularization parameter for sparse representation should make further extensive studies. We can see that the value of the regularization parameter can remarkably influence the performance of the sparse representation algorithms and adjusting the parameters in sparse representation algorithms requires expensive labor. Moreover, adaptive parameter selection based sparse representation methods is preferable and very few methods have been proposed to solve this critical issue.

Second, although sparse representation algorithms have achieved distinctly promising performance on some real-world databases, many efforts should be made in promoting the accuracy of sparse representation based classification, and the robustness of sparse representation should be further enhanced. In terms of the recognition accuracy, the algorithms of $l_1\_l_s$, homotopy and TPTSR achieve the best overall performance. Considering the experimental results of exploiting the seven algorithms on the five databases, the $l_1\_l_s$ algorithm has eight highest classification accuracies, followed by homotopy and TPTSR, in comparison with other algorithms. One can see that the sparse representation based classification methods still can not obtain satisfactory results on some challenge databases. For example, all these representative algorithms can achieve relatively inferior experimental results on the LFW dataset shown in Subsection \ref{subsec9-2}, because the LFW dataset is designed for studying the problem of unconstrained face recognition \cite{huang2007labeled} and most of the face images are captured under complex environments. One can see that the PALM algorithm has the worst classification accuracy on the LFW dataset and the classification accuracy even decreases mostly with the increase of the number of the training samples. Thus, devising more robust sparse representation algorithm is an urgent issue.

Third, enough attention should be paid on the computational inefficiency of sparse representation with $l_1$-norm minimization. One can see that high computational complexity is one of the most major drawbacks of the current sparse representation methods and also hampers its applications in real-time processing scenarios. In terms of speed, PALM, FISTA and DALM take much longer time to converge than the other methods. The average computational time of OMP and TPTSR is the two lowest algorithms. Moreover, compared with the $l_1$-regularized sparse representation based classification methods, the TPTSR has very competitive classification accuracy but significantly low complexity. Efficient and effective sparse representation methods are urgently needed by real-time applications. Thus, developing more efficient and effective methods is essential for future study on sparse representation.

Finally, the extensive experimental results have demonstrated that there is no absolute winner that can achieve the best performance for all datasets in terms of classification accuracy and computational efficiency. However, $l_1\_l_s$, TPTSR and homotopy algorithms as a whole outperform the other algorithms. As a compromising approach, the OMP algorithm can achieve distinct efficiency without sacrificing much recognition rate in comparison with other algorithms and it also has been extensively applied to some complex learning algorithms as a function.

\section{Conclusion}
\noindent
Sparse representation has been extensively studied in recent years. This paper summarizes and presents various available sparse representation methods and discusses their motivations, mathematical representations and extensive applications. More specifically, we have analyzed their relations in theory and empirically introduced the applications including dictionary learning based on sparse representation and real-world applications such as image processing, image classification, and visual tracking.

Sparse representation has become a fundamental tool, which has been embedded into various learning systems and also has received dramatic improvements and unprecedented achievements. Furthermore, dictionary learning is an extremely popular topic and is closely connected with sparse representation. Currently, efficient sparse representation, robust sparse representation, and dictionary learning based on sparse representation seem to be the main streams of research on sparse representation methods. The low-rank representation technique has also recently aroused intensive research interests and sparse representation has been integrated into low-rank representation for constructing more reliable representation models. However, the mathematical justification of low-rank representation seems not to be elegant as sparse representation. Because employing the ideas of sparse representation as a prior can lead to state-of-the-art results, incorporating sparse representation with low-rank representation is worth further research. Moreover, subspace learning also has been becoming one of the most prevailing techniques in pattern recognition and computer vision. It is necessary to further study the relationship between sparse representation and subspace learning, and constructing more compact models for sparse subspace learning becomes one of the popular topics in various research fields. The transfer learning technique has emerged as a new learning framework for classification, regression and clustering problems in data mining and machine learning. However, sparse representation research still has been not fully applied to the transfer learning framework and it is significant to unify the sparse representation and low-rank representation techniques into the transfer learning framework to solve domain adaption, multitask learning, sample selection bias and covariate shift problems. Furthermore, researches on deep learning seems to become an overwhelming trend in the computer vision field. However, dramatically expensive training effort is the main limitation of current deep learning technique and how to fully introduce current sparse representation methods into the framework of deep learning is valuable and unsolved.

The application scope of sparse representation has emerged and has been widely extended to machine learning and computer vision fields. Nevertheless, the effectiveness and efficiency of sparse representation methods cannot perfectly meet the need for real-world applications. Especially, the complexities of sparse representation have greatly affected the applicability, especially the applicability to large scale problems. Enhancing the robustness of sparse representation is considered as another indispensable problem when researchers design algorithms. For image classification, the robustness should be seriously considered, such as the robustness to random corruptions, varying illuminations, outliers, occlusion and complex backgrounds. Thus, developing an efficient and robust sparse representation method for sparse representation is still the main challenge and to design a more effective dictionary is being expected and is beneficial to the performance improvement.

Sparse representation still has wide potential for various possible applications, such as event detection, scene reconstruction, video tracking, object recognition, object pose estimation, medical image processing, genetic expression and natural language processing. For example, the study of sparse representation in visual tracking is an important direction and more depth studies are essential to future further improvements of visual tracking research.

In addition, most sparse representation and dictionary learning algorithms focus on employing the $l_0$-norm or $l_1$-norm regularization to obtain a sparse solution. However, there are still only a few studies on $l_{2,1}$-norm regularization based sparse representation and dictionary learning algorithms. Moreover, other extended studies of sparse representation may be fruitful. In summary, the recent prevalence of sparse representation has extensively influenced different fields. It is our hope that the review and analysis presented in this paper can help and motivate more researchers to propose perfect sparse representation methods.

\section*{Acknowledgment}
This work was supported in part by the National Natural Science Foundation of China under Grant 61370163, Grant 61233011, and Grant 61332011, and in part by the Shenzhen Municipal Science and Technology Innovation Council under Grant JCYJ20130329151843309, Grant JCYJ20130329151843309, Grant JCYJ20140417172417174, Grant CXZZ20140904154910774, the China Postdoctoral Science Foundation funded project, No. 2014M560264 and the Shaanxi Key Innovation Team of Science and Technology (Grant no.: 2012KCT-04).

We would like to thank Jian Wu for many inspiring discussions and he is ultimately responsible for many of the ideas in the algorithm and analysis. We would also like to thank Dr. Zhihui Lai, Dr. Jinxing Liu and Xiaozhao Fang for constructive suggestions. Moreover, we thank the editor, an associate editor, and referees for helpful comments and suggestions which greatly improved this paper.

\bibliographystyle{mytran}
\bibliography{IEEEfull}

\begin{IEEEbiography}[{\includegraphics[width=1in,height=1.25in,clip,keepaspectratio]{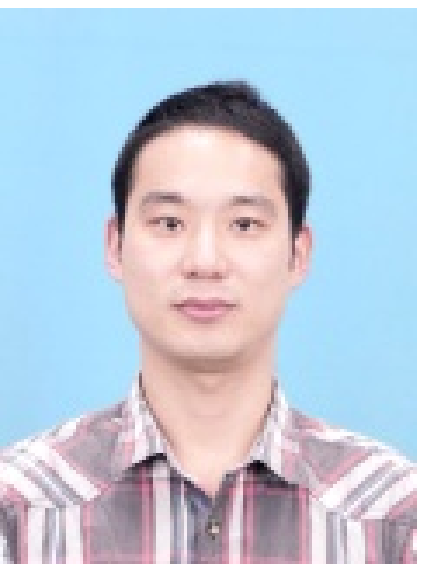}}]{Zheng~Zhang}
received the B.S degree from Henan University of Science and Technology and M.S degree from Shenzhen Graduate School, Harbin Institute of Technology (HIT) in 2012 and 2014, respectively. Currently, he is pursuing the Ph.D. degree in computer science and technology at Shenzhen Graduate School, Harbin Institute of Technology, Shenzhen, China. His current research interests include pattern recognition, machine learning and computer vision.
\end{IEEEbiography}

\begin{IEEEbiography}[{\includegraphics[width=1in,height=1.25in,clip,keepaspectratio]{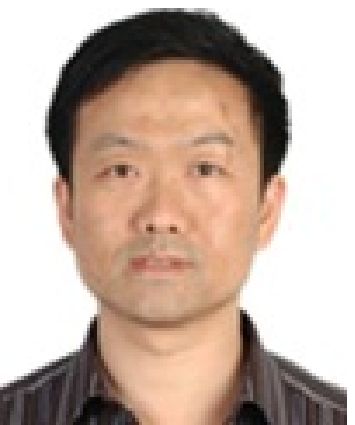}}]{Yong~Xu}
was born in Sichuan, China, in 1972. He received his B.S. degree and M.S. degree at Air Force Institute of Meteorology (China) in 1994 and 1997, respectively. He received the Ph.D. degree in Pattern recognition and Intelligence System at the Nanjing University of Science and Technology (NUST) in 2005. Now, he works at Shenzhen Graduate School, Harbin Institute of Technology. His current interests include pattern recognition, biometrics, machine learning and video analysis.
\end{IEEEbiography}


\begin{IEEEbiography}[{\includegraphics[width=1in,height=1.25in,clip,keepaspectratio]{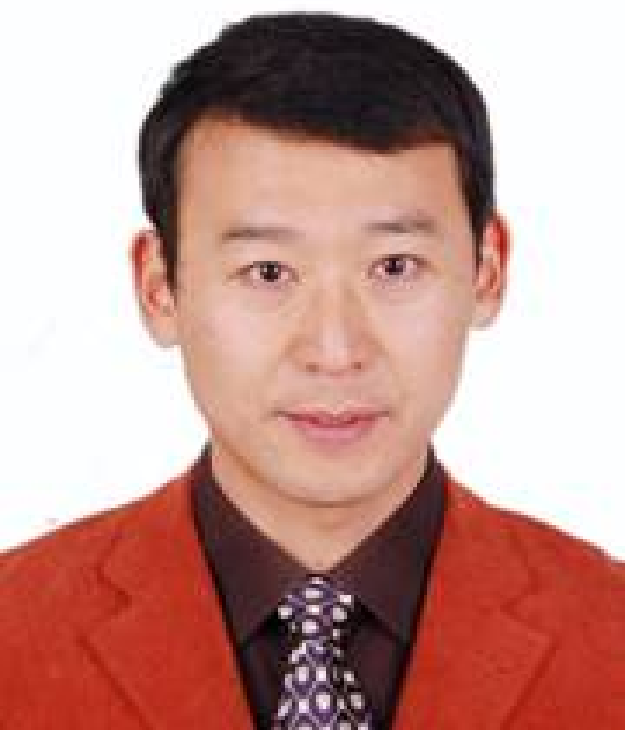}}]{Jian~Yang}
received the B.S. degree in mathematics from the Xuzhou Normal University in 1995. He received the M.S. degree in applied mathematics from the Changsha Railway University in 1998 and the Ph.D. degree from the Nanjing University of Science and Technology (NUST), on the subject of pattern recognition and intelligence systems in 2002. In 2003, he was a postdoctoral researcher at the University of Zaragoza. From 2004 to 2006, he was a Postdoctoral Fellow at Biometrics Centre of Hong Kong Polytechnic University. From 2006 to 2007, he was a Postdoctoral Fellow at Department of Computer Science of New Jersey Institute of Technology. Now, he is a professor in the School of Computer Science and Technology of NUST. He is the author of more than 80 scientific papers in pattern recognition and computer vision. His journal papers have been cited more than 1600 times in the ISI Web of Science, and 2800 times in the Web of Scholar Google. His research interests include pattern recognition, computer vision and machine learning. Currently, he is an associate editor of Pattern Recognition Letters and IEEE TRANSACTION ON NEURAL NETWORKS AND LEARNING SYSTEMS, respectively.
\end{IEEEbiography}

\begin{IEEEbiographynophoto}{Xuelong~Li}
is a Full Professor with the Center for OPTical IMagery Analysis and Learning (OPTIMAL), State Key Laboratory of Transient Optics and Photonics, Xi'an Institute of Optics and Precision Mechanics, Chinese Academy of Sciences, Xi'an 710119, Shaanxi, P.R. China.
\end{IEEEbiographynophoto}

\begin{IEEEbiography}[{\includegraphics[width=1in,height=1.25in,clip,keepaspectratio]{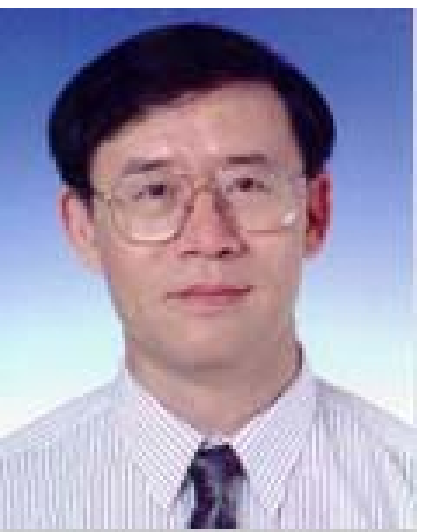}}]{David~Zhang}
graduated in Computer Science from Peking University. He received his M.Sc. in Computer Science in 1982 and his Ph.D. in 1985 from the Harbin Institute of Technology (HIT). From 1986 to 1988 he was a Postdoctoral Fellow at Tsinghua University and then an Associate Professor at the Academia Sinica, Beijing. In 1994 he received his second Ph.D. in Electrical and Computer Engineering from the University of Waterloo, Ontario, Canada. Currently, he is a Chair Professor at the Hong Kong Polytechnic University where he is the Founding Director of the Biometrics Technology Centre (UGC/CRC) supported by the Hong Kong SAR Government in 1998. He also serves as Visiting Chair Professor in Tsinghua University, and Adjunct Professor in Shanghai Jiao Tong University, Peking University, Harbin Institute of Technology, and the University of Waterloo. He is the Founder and Editor-in-Chief, International Journal of Image and Graphics (IJIG); Book Editor, Springer International Series on Biometrics (KISB); Organizer, the first International Conference on Biometrics Authentication (ICBA); Associate Editor of more than ten international journals including IEEE Transactions and Pattern Recognition; Technical Committee Chair of IEEE CIS and the author of more than 10 books and 200 journal papers. Professor Zhang is a Croucher Senior Research Fellow, Distinguished Speaker of the IEEE Computer Society, and a Fellow of both IEEE and IAPR.
\end{IEEEbiography}

\end{document}